\documentclass[sigconf]{acmart}

\AtBeginDocument{%
  }

\setcopyright{acmlicensed}
\copyrightyear{2024}
\acmYear{2024}
\acmDOI{10.1145/3664647.3680829}

\acmConference[MM '24] {Proceedings of the 32nd ACM International Conference on Multimedia}{October 28--November 1, 2024}{Melbourne, VIC, Australia.}
\acmBooktitle{Proceedings of the 32nd ACM International Conference on Multimedia (MM '24), October 28--November 1, 2024, Melbourne, VIC, Australia}

\acmISBN{979-8-4007-0686-8/24/10}

\usepackage{url}
\usepackage{hyperref}

\usepackage{multirow} 
\usepackage{makecell} 

\usepackage{colortbl}

\usepackage{subfig}
\usepackage{float}

\definecolor{unit01green}{RGB}{82,208,83}

\definecolor{unit02red}{RGB}{211,41,15}
\newcommand{\red}[1]{\textcolor{unit02red}{#1}}

\definecolor{unit02blue}{RGB}{53,49,255}
\newcommand{\blue}[1]{\textcolor{unit02blue}{#1}}

\newcommand{\dtshow}[1]{\selectfont \textbf{\blue{#1}}}
\newcommand{\plus}[1]{\selectfont {\blue{#1}}}

\usepackage{amsmath, amsfonts, bm}

\newcommand{\x}{$\bm {x}$}
\newcommand{\doc}{$\bm {d}$}

\newcommand{\mathDivLoss}{$\mathcal{L}_{div}$}
\newcommand{\mathVarLoss}{$\mathcal{L}_{var}$}

\newcommand{\mathLocalWeight}{$\lambda_{local}$}
\newcommand{\mathDivWeight}{$\lambda_{div}$}
\newcommand{\mathVarWeight}{$\lambda_{var}$}

\newcommand{\mathVisualSemanticEmbed}{${\bm B}_{V}$}
\newcommand{\mathTextSemanticEmbed}{${\bm B}_{T}$}

\newcommand{\ModelName}{EmDepart}
\newcommand{\nameViewModuleName}{SDM}

\DeclareMathOperator*{\argmax}{arg\,max}

\begin{document}

\title{Visual-Semantic Decomposition and Partial Alignment for Document-based Zero-Shot Learning}

\author{Xiangyan Qu}
\affiliation{%
  \institution{Institute of Information Engineering, Chinese Academy of Sciences}
  \institution{School of Cyber Security, University of Chinese Academy of Sciences}
  \city{Beijing, China}
  \country{}}
\email{quxiangyan@iie.ac.cn}

\author{Jing Yu}
\authornotemark[2]
\affiliation{%
  \institution{Institute of Information Engineering, Chinese Academy of Sciences}
  \institution{School of Cyber Security, University of Chinese Academy of Sciences}
  \city{Beijing, China}
  \country{}}
\email{yujing02@iie.ac.cn}

\author{Keke Gai}
\affiliation{%
  \institution{School of Cyberspace Science and Technology, Beijing Institute of Technology}
  \city{Beijing, China}
  \country{}}
\email{gaikeke@bit.edu.cn}

\author{Jiamin Zhuang}
\affiliation{%
  \institution{Institute of Information Engineering, Chinese Academy of Sciences}
  \institution{School of Cyber Security, University of Chinese Academy of Sciences}
  \city{Beijing, China}
  \country{}}
\email{zhuangjiamin@iie.ac.cn}

\author{Yuanmin Tang}
\affiliation{%
  \institution{Institute of Information Engineering, Chinese Academy of Sciences}
  \institution{School of Cyber Security, University of Chinese Academy of Sciences}
  \city{Beijing, China}
  \country{}}
\email{tangyuanmin@iie.ac.cn}

\author{Gang Xiong}
\affiliation{%
  \institution{Institute of Information Engineering, Chinese Academy of Sciences}
  \institution{School of Cyber Security, University of Chinese Academy of Sciences}
  \city{Beijing, China}
  \country{}}
\email{xionggang@iie.ac.cn}

\author{Gaopeng Gou}
\affiliation{%
  \institution{Institute of Information Engineering, Chinese Academy of Sciences}
  \institution{School of Cyber Security, University of Chinese Academy of Sciences}
  \city{Beijing, China}
  \country{}}
\email{gougaopeng@iie.ac.cn}

\author{Qi Wu}
\affiliation{%
  \institution{Australia Institute of Machine Learning, University of Adelaide}
  \city{Adelaide, Australia}
  \country{}}
\email{qi.wu01@adelaide.edu.au}

\renewcommand{\shortauthors}{Xiangyan Qu et al.}

\begin{abstract}
Recent work shows that documents from encyclopedias serve as helpful auxiliary information for zero-shot learning. Existing methods align the entire semantics of a document with corresponding images to transfer knowledge. However, they disregard that semantic information is not equivalent between them, resulting in a suboptimal alignment. In this work, we propose a novel network to extract multi-view semantic concepts from documents and images and align the matching rather than entire concepts. Specifically, we propose a semantic decomposition module to generate multi-view semantic embeddings from visual and textual sides, providing the basic concepts for partial alignment. To alleviate the issue of information redundancy among embeddings, we propose the local-to-semantic variance loss to capture distinct local details and multiple semantic diversity loss to enforce orthogonality among embeddings. Subsequently, two losses are introduced to partially align visual-semantic embedding pairs according to their semantic relevance at the view and word-to-patch levels. Consequently, we consistently outperform state-of-the-art methods under two document sources in three standard benchmarks for document-based zero-shot learning. Qualitatively, we show that our model learns the interpretable partial association. Code is available at \href{https://github.com/MorningStarOvO/EmDepart/}{here}.
\end{abstract}




\maketitle

\section{Introduction}
Image recognition tasks have achieved significant success relying on enormous manually labeled data.
However, it is impractical to collect and annotate all kinds of images.
Zero-Shot Learning (ZSL)~\citep{ZSL_invent1, ZSL_invent2} emerges as a promising paradigm to address this issue.
ZSL aims to identify unseen classes by training on a set of seen classes. 
The key challenge in ZSL is how to leverage auxiliary information to transfer knowledge from seen to unseen classes.

Common auxiliary information includes attributes~\citep{AWA_attribute_provide2,  CUB_and_attribute_provide}, word embeddings~\citep{Glove, pre_trained_LM_ACL_2012, pre_trained_LM_NIPS_2013, pre_trained_LM_EMNLP_2020_Wikipedia2vec}, and category documents~\citep{TF-IDF, BERT, LongFormer}. 
Most work ~\citep{CVPR_2023_PSVMA, CVPR_2022_En-Compactness, CVPR_2022_MSDN, MM_22_1, MM_23_3} leverages human-annotated attributes as auxiliary information.
These methods assume that seen and unseen classes can be defined with the same attributes. 
However, these attributes are labor-intensive and challenging to scale ~\citep{Attribute_drawback_CVPR_2013, Attribute_drawback_ECCV_2018}, which are impossible for many real-world scenarios. 
To address this issue, other work ~\citep{word_embed_NIPS_2013_Richard_Socher, word_embed_NIPS_2013_Devise, word_embed_CVPR_2015_Zeynep_Akata, VGSE} applies category word embeddings from the pre-trained language model to replace attributes. However, the category name offers limited discriminative information ~\citep{word_embed_CVPR_2015_Zeynep_Akata, Attribute_drawback_and_document_arxiv_2021}, imposing potential limitations on performance.
Recently, some work ~\citep{Document-based_NAACL_2021_Jihyung_Kil, I2DFormer, I2MVFormer} demonstrates that documents from the encyclopedias serve as valuable sources of auxiliary information, which contain multiple semantic concepts and knowledge from experts. 

\begin{figure}[t]
    \centering
    \includegraphics[width=0.7\linewidth]{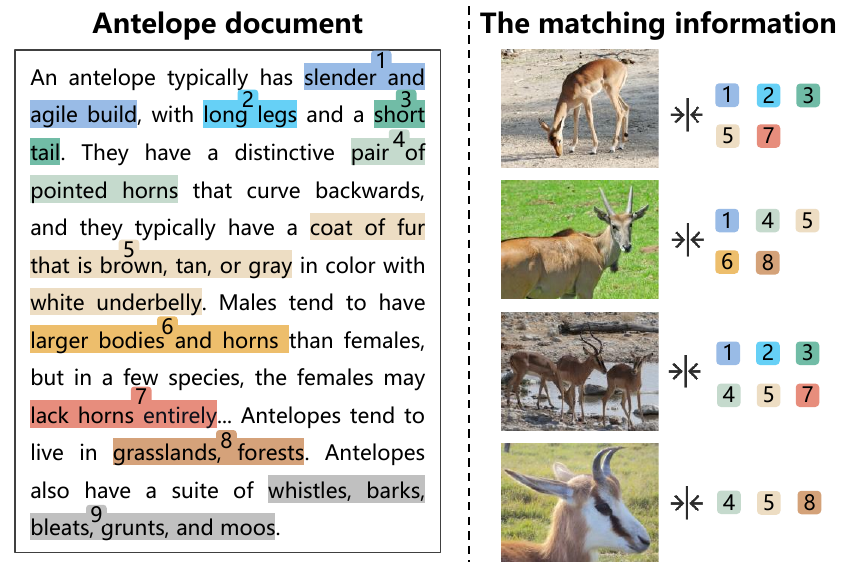}
    \vspace{-10pt}
    \caption{Partial associations between documents and images. The semantic content in the category document may partially be reflected in the image. Distinct images capture varying aspects of the semantic information within the document.}
    \Description{}
    \label{fig:Motivation}
    \vspace{-12pt}
\end{figure}

For document-based ZSL, seen and unseen classes are described by a composition of similar semantic concepts (noun phrases that visually describe a class).
Aligning basic semantic concepts with corresponding image regions accurately is the key to knowledge transfer. 
Recent methods \citep{I2DFormer, I2MVFormer} apply fine-grained interactions between words (or documents) and image patches to enhance semantic alignment.
However, they are designed without considering the \textbf{\textit{partial association}} between noisy documents and visual-diverse images:
\textbf{1) Noisy document}: Documents from encyclopedias mainly cover many views, \textit{e.g.} shape, color, habitat, sound, and diet. However, some views may not include visual information (see ``9'' in the left of Figure~\ref{fig:Motivation}), \textit{e.g.} sound and diet. These non-visual views are harmful to knowledge transfer.
\textbf{2) Exhaustive description}: Documents comprehensively describe the possible characteristics of the category. However, a single image typically captures part of them. For example, the last image on the right of Figure ~\ref{fig:Motivation} only shows the shape of the horn, color, and habitat in the antelope document.
However, these methods align the entire semantics of documents with images, obtaining a suboptimal alignment.
\textbf{3) Visual-diverse image content}: Due to variations in shooting angles, lighting, locations, and states, images of the same category convey varying semantic concepts from the document (see the right of Figure ~\ref{fig:Motivation}).
Aligning diverse images with the same document semantics makes it hard to build accurate semantic alignment.
Therefore, accurately modeling the partial association between document and image becomes an urgent problem for document-based ZSL. 


\begin{figure}[t]
    \centering
    \includegraphics[width=0.75\linewidth]{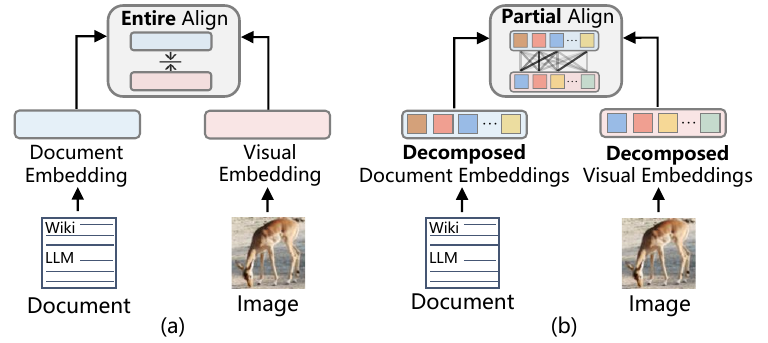}
    \vspace{-10pt}
    \caption{Illustration of different methods. (a) Existing methods align the entire semantics of documents with images. (b) Our model decomposes semantic concepts and models the partial association to align the matching concepts accurately.}
    \Description{}
    \label{fig:Motivation_New}
    \vspace{-14pt}
\end{figure} 
%

To this end, we propose an \textbf{Em}bedding \textbf{De}composition and \textbf{Part}ial Alignment (EmDepart) network, as illustrated in Figure ~\ref{fig:Motivation_New}(b), to extract multi-view semantic concepts from document and image and accurately align the matching concepts.
Specifically, the Semantic Decomposition Module (SDM) is proposed to generate multi-view semantic embeddings from visual and textual sides, providing the basic concepts for partial alignment.
However, the SDM may generate multiple embeddings with a slight variance, resulting in information redundancy, denoted as feature collapse.
To alleviate this issue, we propose the local-to-semantic variance loss to capture unique local details and multiple semantic diversity loss to make each embedding orthogonal to others.
Subsequently, we rely on the semantic similarity of every visual and textual view embedding pair to model the partial association.
Two losses are introduced to partially align these pairs according to their semantic relevance at the view and word-to-patch levels. 
Moreover, a novel score is applied to filter out unmatched information to measure semantic similarity accurately at the inference. 
Since some fine-grained categories are less described in the encyclopedia, we also design a novel prompt strategy to enrich these documents.

Our key contributions are as follows.
(1) We propose a novel network that decomposes concepts from document and image into multi-view semantic embeddings and aligns them partially according to semantic relevance. This addresses the suboptimal alignment caused by ignoring the partial association in document-based ZSL. It sheds new light on the vision-and-language partial semantic alignment.
(2) To solve the issue of information redundancy caused by feature collapse, we introduce the semantic decomposition module with the local-to-semantic variance loss to capture unique local details and multiple semantic diversity loss to enhance orthogonality among the embeddings. 
These losses also improve the performance of previous methods by 4.1\% on average.
(3) With comparable training parameters, our model consistently outperforms state-of-the-art methods for document-based ZSL and GZSL settings in three standard benchmarks.
It improves performance by 6.0\% and 5.8\% on average across all metrics under Wiki and Wiki+LLM documents. 
Moreover, we qualitatively demonstrate that our model learns the interpretable partial semantic association.


\section{Related work}

\noindent \textbf{Zero-Shot Learning}
aims to train on seen classes and generalize to recognize unseen classes~\citep{ZSL_invent1, ZSL_invent2}.
Most work leverages human-annotated attributes ~\citep{AWA_attribute_provide1, AWA_attribute_provide2, AWA1_and_attribute_provide, CUB_and_attribute_provide} as auxiliary information. 
These methods transfer knowledge by utilizing compatibility functions ~\citep{CVPR_2013_Zeynep, TPAMI_2015_Zeynep_Akata, ICML_2015_Bernardino, CVPR_2016_Soravit_Changpinyo, CVPR_2016_Yongqin_Xian} to map embeddings into a common space, incorporating generative model to generate unseen classes samples~\citep{CVPR_2018_Yongqin_Xian, CVPR_2018_Vinay_Kumar_Verma, ICCV_2019_Yizhe_zhu, CVPR_2019_f_vaegan_d2}, and enhancing semantic alignment between attributes and image regions ~\citep{CVPR_2022_MSDN, CVPR_2022_Hongzu_Su, CVPR_2022_En-Compactness, CVPR_2023_PSVMA, NIPS_2022_Wenjia_Xu, AAAI_2023_Duet, MM_22_1, MM_23_1, MM_23_2}.
However, annotating attributes needs large human resources and deep domain expertise.
In contrast, other work ~\citep{word_embed_NIPS_2013_Richard_Socher, word_embed_NIPS_2013_Devise, word_embed_CVPR_2015_Zeynep_Akata, VGSE} applies word embedding from pre-trained language models ~\citep{Glove, pre_trained_LM_ACL_2012, pre_trained_LM_NIPS_2013, pre_trained_LM_EMNLP_2020_Wikipedia2vec}, which transfers knowledge through the semantic relationship between different categories. 
Several methods~\citep{word_embed_and_graph_ICCV_2017, word_embed_and_graph_CVPR_2018_Xiaolong_Wang, word_embed_and_graph_CVPR_2019_Michael_Kampffmeyer, word_embed_and_graph_CVPR_2021_Muhammad_Ferjad_Naeem}~ enhance semantic connections by knowledge graphs.
However, they achieve poor performance because of the category name with little discriminative information and sensitivity to linguistic issues~\citep{word_embed_CVPR_2015_Zeynep_Akata, Attribute_drawback_and_document_arxiv_2021}. 
In contrast, documents are easy to collect from encyclopedias, which contain multiple semantic concepts. 
In this work, we improve knowledge transfer for document-based ZSL by accurate partial alignment.
\begin{figure*}[t]
    \centering
    \includegraphics[width=0.99\linewidth]{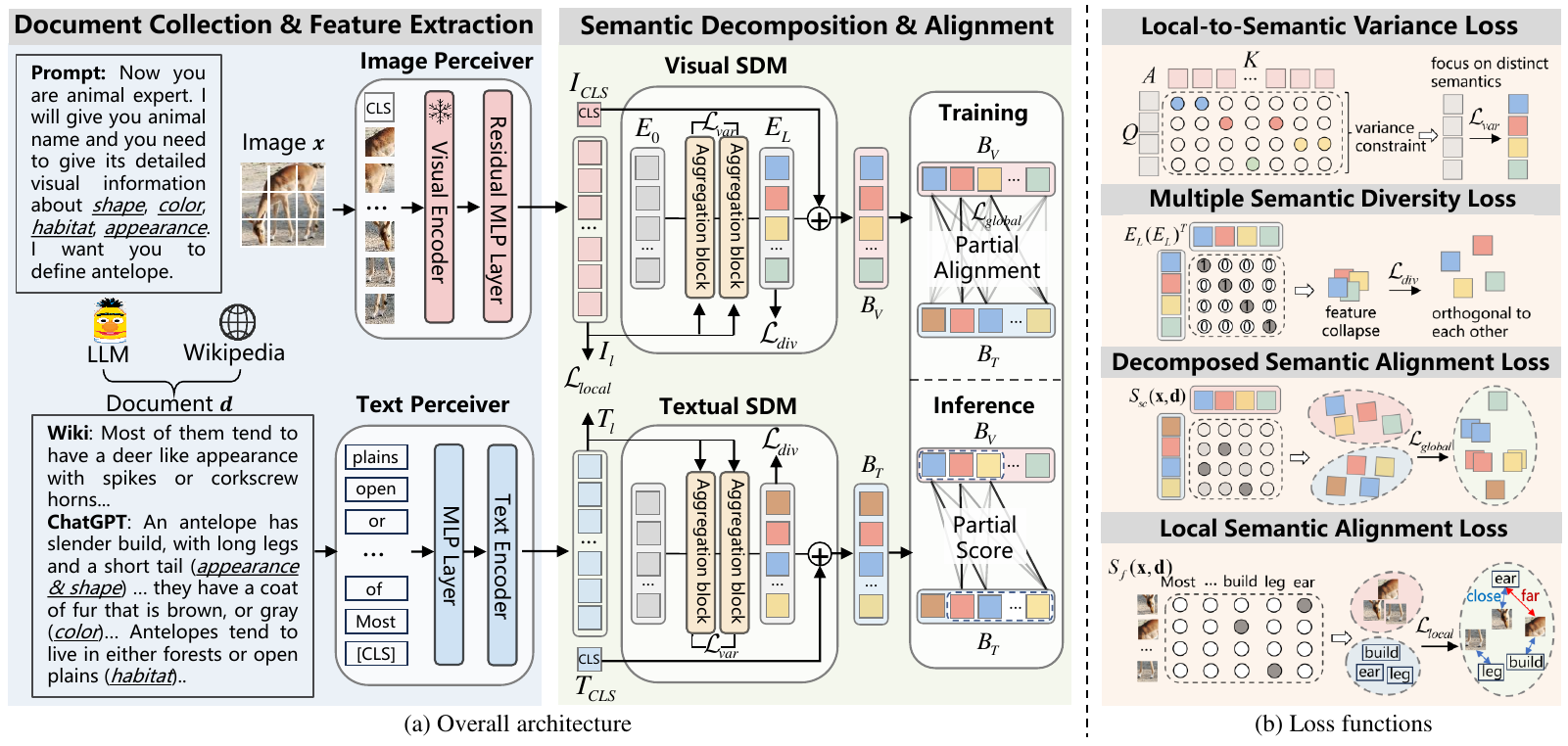}
    \vspace{-10pt}
    \caption{An overview of our model.
    (a) The EmDepart contains an image perceiver, a text perceiver, and visual and textual semantic decomposition modules.  (b) Our loss functions. The first loss encourages each view embedding to focus on distinct local details. The second loss penalizes each embedding orthogonal to others. The last two losses partially align semantics at the view and word-to-patch levels. 
    }
    \Description{}
    \label{fig:model_arch}
     \vspace{-10pt}
\end{figure*}

\noindent \textbf{Document-based Zero-Shot Learning} 
uses definition-level text corpora from encyclopedias to obtain auxiliary information.
Most work ~\citep{Document-based_ICCV_2015_Lei_Jimmy_Ba, Document-based_CVPR_2017_Ziad, Attribute_drawback_and_document_arxiv_2021, Document-based_NAACL_2021_Jihyung_Kil, Document-based_CVPR_2016_Ruizhi_Qiao} utilize document embedding by TF-IDF ~\citep{TF-IDF} or large language models ~\citep{BERT, LongFormer} as auxiliary information. 
While several methods ~\citep{Document-based_CVPR_2018_Yizhe_Zhu, Document-based_CVPR_2017_Mohamed} enhance embeddings by part detection network, annotated attributes are used to train the detection model. 
Recently, some work ~\citep{I2DFormer, I2MVFormer} learns fine-grained interactions to enrich semantic embedding. 
Specifically, I2DFromer ~\citep{I2DFormer} trains a model to align image patches with words in global and local compatibility.
I2MVFormer ~\citep{I2MVFormer} aggregates information at the document level to reduce computation cost, aligning with image regions. 
However, these methods align the entire semantics of documents with images, ignoring the partial association between them. 
This results in suboptimal semantic alignment. 
In this work, our EmDepart generates multi-view semantic embeddings and models the partial association to align the matching semantics accurately.  



\noindent \textbf{Set-based Embedding Methods}
aims to learn multiple embeddings to alleviate the semantic ambiguity in cross-modal retrieval task, \textit{i.e.}, an image is semantically matched with multiple captions. 
This is similar to the challenge for document-based ZSL, \textit{i.e.}, the partial association between the document and diverse images.
To be specific, PVSE ~\citep{set_based_embed_CVPR_2019} and TVMM ~\citep{set_based_embed_NIPS_2022} learn a set of embeddings by linear combination with local and global features. PCME ~\citep{set_based_embed_CVPR_2021} represents each sample as a probabilistic embedding. 
The state-of-the-art method ~\citep{set_based_embed_CVPR_2023} explores slot attention ~\citep{slot_attention} to enhance diversity in set-based embeddings and smooth chamfer similarity to solve sparse supervision and set collapsing problems.
However, text corpora in document-based ZSL are at definition level ($\approx$ 500 words).
It is hard to obtain multiple embeddings with little information redundancy solely through model architecture.
Therefore, we propose two losses to enhance the information difference among semantic embeddings, alleviating the problem of feature collapse. 


\section{Method}

Our \textbf{Em}bedding \textbf{De}composition and \textbf{Part}ial Alignment (EmDepart) network is illustrated in Figure ~\ref{fig:model_arch}.
We first collect documents from encyclopedias and enrich less-described categories by LLMs.
The image perceiver and text perceiver extract salient features for ZSL tasks. 
Then, the visual and textual Semantic Decomposition Modules (SDM) decompose perceived features into multi-view semantic embeddings. 
We leverage these embeddings to partially align the semantic information at the view and word-to-patch levels.


\noindent \textbf{Notations.}
Zero-Shot Learning (ZSL) aims to train a classifier on seen classes $\mathcal{Y}^s$ to recognize unseen classes $\mathcal{Y}^u$ during the test, where $\mathcal{Y}^s \cap \mathcal{Y}^u = \emptyset$. 
The training set $\{ (\bm x, \bm y, \bm d) | \bm x \in \mathcal{X}^s, \bm y \in \mathcal{Y}^s, \bm d \in \mathcal{D}^s \}$ consists of image $\bm x$, its label $\bm y$ and auxiliary information, \textit{i.e.}, document $\bm d$. 
These documents are from a collection of textual descriptions of seen classes.
At test time, another collection of images $\mathcal {X}^{test} $, their potential classes $\mathcal {Y}^{test}$, and corresponding documents $\mathcal{D}^{test} $ will be available to evaluate the model.
In the ZSL setting, test images are from unseen classes, and for generalized ZSL (GZSL), from seen and unseen classes.

\subsection{Document Collection}
Category documents are the theoretical foundation for knowledge transfer in document-based ZSL.
Each class (both seen and unseen classes) has a corresponding document that visually describes it.

\noindent \textbf{Document Collection from Encyclopedia.}
Similar to ~\citep{I2DFormer, I2MVFormer}, we leverage the A-Z animals ~\citep{azanimal} for AWA2 ~\citep{AWA2}, AllAboutBirds ~\citep{aab} for CUB ~\citep{CUB_and_attribute_provide}, and Wikipedia ~\citep{wiki} for FLO ~\citep{FLO} to collect documents.
Following previous methods, we select relevant sections in the encyclopedia to filter potential noises for the ZSL task.

\noindent \textbf{Enriching Less-Described Document.}
Some fine-grained categories are less described in the encyclopedia, such as dogs chihuahua and collie in AWA2, Nighthawk and Green Violetear in CUB, and most classes in FLO. 
Therefore, we instruct Large Language Models (LLMs) to generate category definitions to enrich these documents by the following prompt:

\textit{``Now you are a \{\textbf{type}\} expert. I will give you \{\textbf{type}\} name, and you need to give detailed visual information about its shape, color, appearance, habitat, etc. I want you to define \{\textbf{class name}\}.''}

\noindent We use category species as \{type\}, \textit{i.e.}, animal for AWA2, bird for CUB, and flower for FLO. 
We concatenate documents from encyclopedias with LLM-generated to serve as the final auxiliary information.
To save computation costs, we enrich less-described instead of all categories.
More details are shown in the supplementary.


\subsection{Feature Extractor}

\noindent \textbf{Image Perceiver.}
Given an input image \x, the image perceiver first encodes features by a fixed ViT ~\citep{vit}. 
Then, a learnable MLP layer with a residual connection maps image features to dimension $r$, extracting crucial visual information for ZSL tasks.
The image perceiver outputs [\textit{CLS}] token ${\bm I}_{CLS}  \in \mathbb{R}^r $ as the global image feature and other tokens ${\bm I}_l \in \mathbb{R}^{n \times r} $ as patch-wise local image features, where $n$ is the number of image patches.

\noindent \textbf{Text Perceiver.} 
Given a $m$-words document \doc, we use GloVe ~\citep{Glove} to initiate each word as input features. 
Similar to previous work ~\citep{I2DFormer, I2MVFormer}, the text perceiver passes these token features through a learnable MLP with dimension $r$ (r < 300) to reduce computation cost.
We add a learnable [\textit{CLS}] token to the sequence and input this sequence into the text encoder, which consists of two transformer encoder blocks.
The text encoder outputs [\textit{CLS}] token ${\bm T}_{CLS} \in \mathbb{R}^{r}$ as global text feature and ${\bm T}_l \in \mathbb{R}^{m \times r}$ as word-wise local text features.

\subsection{Semantic Decomposition Module}
\label{sec:multi_semantic_prediction_module}
The semantics in images and documents are from multiple views, such as shape, color, and habitat.
We introduce visual and textual semantic decomposition modules (SDM) to aggregate information from the perceived features of each modality and decompose them to generate multi-view semantic embeddings.
This process provides the basic semantic concepts for partial alignment.

Taking visual SDM as an example, the textual side follows a similar process.
As shown in the middle of Figure ~\ref{fig:model_arch}, SDM contains $l$-layer aggregation blocks integrating perceived features through iterations. 
In the initial iteration, we introduce a set of learnable tokens $\bm{E}_{0} \in \mathbb{R}^{k \times r}$, referred to as view embeddings later, where $k$ denotes the number of embeddings ($k \ll n$).
Subsequently, for the $t$-th iteration, we feed both $\bm{E}_{t-1}$ and local image features ${\bm I}_l$ to the $t$-th aggregation block ($t = 1, 2, ..., l$), iteratively refining the visually discriminative information. 

Each aggregation block aims to aggregate semantics and extract helpful visual information for the ZSL task.
For the $t$-th aggregation block, we map local image features $ {\bm I}_l$ to key $\bm K$ and value $\bm V $ and view embeddings $\bm{E}_{t-1}$ to query $\bm Q$ by three linear layers. 
Then, they are fed into an attention mechanism to obtain the $\hat{\bm{E}}_t \in \mathbb{R}^{k \times r}$:
\begin{equation}
    \hat {\bm{E}}_{t} = \text{softmax}  \bigg ( \frac{\bm{Q} \bm{K}^{T}}{\sqrt{r_{h}}} \bigg ) \bm{V} \bm{W}_o + \bm {E}^{t-1},
\end{equation}
where $r_h$ is the dimension of head attention and $\bm{W}_o \in \mathbb{R}^{r_h \times r}$ is a linear layer to map features to the original dimension $r$.
Subsequently, we feed $\hat{\bm{E}}_t$ to a learnable MLP followed by layer normalization, residual connection, and GELU activation ~\citep{GELU}, outputting the iterative visual-semantic information refinement of the view embeddings $\bm{E}_{t}$:
\begin{equation}
    \bm{E}_{t} = \text{MLP} (\hat {\bm{E}}_{t}) + \hat {\bm{E}}_{t}.
\end{equation}

In the last iteration, we concatenate the $\bm{E}_{L} \in \mathbb{R}^{k \times r}$, the output of the final aggregation block, with $k$ repetitions of global image feature ${\bm I}_{CLS}$ followed by a layer normalization.
This operation constraints view embeddings with small within-set variance and outputs the multi-view visual semantic embeddings \mathVisualSemanticEmbed $ \in \mathbb{R}^{k \times r}$:
\begin{equation}
    {\bm B}_{V} = \text{LayerNorm}(\bm{E}_{L} + [{\bm I}_{CLS}] ^ {\times k}).
    \label{eq: concat_global}
\end{equation}
Similarly, we obtain textual semantic embeddings \mathTextSemanticEmbed $ \in \mathbb{R}^{k \times r}$. 

\subsection{Distinct Semantic Information Learning}
Since definition-level corpora contain numerous words ($\approx 500$), view embeddings are hard to attend diverse semantics only by model architecture.
The challenge also appears on the visual side.
Specifically, the SDM may generate view embeddings with a slight variance, resulting in information redundancy, denoted as feature collapse.
To solve this issue, we introduce two losses in SDM.

\noindent \textbf{Local-to-Semantic Variance Loss}
aims to encourage each view embedding to focus on unique local information (see the first loss in Figure \ref{fig:model_arch} (b)).
Taking the visual side as an example, it enforces different view embeddings to show distinct attention to the same image patch.
We make the following variance constraints on attention maps $\bm A_V$ in visual SDM, a dot product between $\bm Q$ and $\bm K$:
\begin{equation}
    C(\bm{A}_V) = \sum^l_{{t}=1}\sum^n_{j=1}\text{max}  (0, \gamma - \sqrt{Var(\bm{a}_{tj}  ) + \epsilon} ),
\end{equation}
where $\bm a_{tj}$ denotes the attention between visual view embeddings and the $j$-th image patch in the $t$-th aggregation block. 
The $l$ and $n$ are the number of aggregation blocks and image patches, respectively.
We offer a constant value $\gamma$ for constraints, which ensures view embeddings with a certain diversity to the attention of the same patch.
The $\epsilon$ is a small scalar to maintain numerical stabilities.
Similarly, we penalize attention maps $\bm{A}_T$ between textual view embeddings and each word token.
This loss is formulated as: 
\begin{equation}
\begin{split}
     \mathcal{L}_{var} = \frac{1}{2}\bigg( C(\bm{A}_{{T}}) + C(\bm{A}_{{V}}) \bigg ).
\end{split}
\end{equation}
After $l$ iterations, each visual and textual view embedding carries distinct semantic information, establishing the foundation for decomposing information.

\noindent \textbf{Multiple Semantic Diversity Loss}
aims to enhance the information decoupling among view embeddings.
It forces minimal semantic redundancy between view embeddings by making each embedding orthogonal to others, shown in the second loss of Figure \ref{fig:model_arch}(b).
Notably, view embeddings of the final output ($\bm B_V$ and $\bm B_T$) contain the global feature, which may invalidate the orthogonality constraint. 
Therefore, we penalize redundancy among the output of the final aggregation block, \textit{i.e.}, $\bm{E}_L$. 
Specifically, we normalize each $\bm{E}_L$ and calculate the cosine similarity between them, yielding the redundant  matrix $\bm{M}_V = \bm{E}_L (\bm{E}_L)^T$. 
The loss penalizes the visual and textual redundant matrix (denoted as $\bm{M}_{{T}}$) to approximate the identity matrix $\mathbb{I} \in \mathbb{R}^{k \times k}$ via an $l_2$-norm minimization:
\begin{equation}    
    \mathcal{L}_{div} = \frac{1}{2k^2} ( \Vert \bm{M}_{{T}} - \mathbb I \Vert_2 + \Vert \bm{M}_{{V}} - \mathbb I \Vert_2 ),
\end{equation}
where $\bm{M}_{{T}}$ is computed in a similar way.
This objective ensures that different view embeddings maintain orthogonality, characterized by non-diagonal cosine similarity values converging towards zero.

To summarize, since the \mathVarLoss~ penalizes view embeddings to focus on different local information and the \mathDivLoss~ constrains each embedding to be orthogonal to others, the SDM generates decomposed view embeddings with distinct semantic information.


\subsection{Partial Semantic Alignment}
\label{sec:similarity_function}

To accurately model the partial association between documents and images, we introduce two losses to align the matching semantic concepts at the view and word-to-patch levels.

\noindent \textbf{Decomposed Semantic Alignment}
aims to align decomposed visual and textual view embeddings according to their semantic relevance.
We rely on the semantic similarity between visual and textual view embedding pairs to model the partial association by the Smooth Chamfer ~\citep{set_based_embed_CVPR_2023} function. 
We first review the Smooth Chamfer, which assigns distinct weights to every document-image embedding pair based on similarity: 
\begin{equation}
    S_{sc}(\bm x, \bm d) = \frac{1}{2k}\bigg(\sum_{{\bm b}_{{T}} \in {{\bm B}_{{T}}}} \text{LSE}({\bm b}_{{T}}, {\bm B}_{V}) 
    +  \sum_{{\bm b}_{V} \in {\bm B}_{V}} \text{LSE} ({\bm b}_{V}, {\bm B}_{T})\bigg), 
    \label{eq: smooth_chamfer}
\end{equation}
where $\text{LSE}(\bm b, \bm B)$ is smooth approximation for maximum cosine similarity between vector $\bm b$ and elements in set $\bm B$. 
Taking $\text{LSE}({\bm b}_{{T}}, {\bm B}_{V})$ as an example, it is formulated as: 
\begin{equation}
    \text{LSE}({\bm b}_{{T}}, {\bm B}_{V}) = \log \bigg(\sum_{{\bm b}_{{V}} \in {\bm B}_{V} } \text{e}^{\text{cos}({\bm b}_{{T}}, {\bm b}_{{V}})} \bigg),
\end{equation}
where $\text{cos}(\cdot)$ is the cosine similarity.
We introduce a cross-entropy loss to encourage image $\bm x$ and corresponding document $\bm d$ to be closer than other pairs: 
\begin{equation}
    \mathcal{L}_{global} = -\log \frac{\text{exp} (S_{sc}(\bm x, \bm d) / \tau)}{\sum_{{\bm d}' \in \mathcal{D}^s} \text{exp} (S_{sc}(\bm x, {\bm d}') / \tau)}, 
\end{equation}
where $\tau$ is a temperature scalar.   
The $\mathcal{L}_{global}$ is designed to smoothly align each visual embedding in \mathVisualSemanticEmbed~ with the most similar element from textual embeddings \mathTextSemanticEmbed, and vice versa (see the third loss in Figure ~\ref{fig:model_arch}(b)).
This process models the partial association between visual and textual spaces, which embodies the fact that an image is reflected as part of semantics in the document.
Consequently, we align the two spaces more accurately.

\noindent \textbf{Local Semantic Alignment} 
aims to apply interactions between image patches and word tokens for fine-grained semantic alignment (see the fourth loss in Figure \ref{fig:model_arch}(b)).
It provides the basic semantic concepts for partial semantic alignment and discriminative information for fine-grained classification.
Similar to ~\citep{I2DFormer, I2MVFormer}, we first fuse local image and text features by a cross-attention module, which leverages the semantic information in the document to enrich the visual features. 
The cross-attention module takes local image features ${\bm I}_l$ as query and local text features ${\bm T}_l$ as key and value, outputting semantic-enhanced visual features $\tilde {\bm I} \in \mathbb{R}^{n \times r}$.
Subsequently, we apply a global pooling on patch dimension to aggregate the visually fine-grained information, yielding the $\overline{\bm {I}} \in \mathbb{R}^{1 \times r}$.
Afterward, a fine-grained similarity score $S_{f} (\bm{x}, \bm{d}) = \text{D}(\overline{\bm {I}})$ is introduced through a linear layer $\text{D}(\cdot) \in \mathbb{R}^{r \times 1}$.
The objective is to encourage the image $\bm x$ to be close to the corresponding category document $\bm d$ on fine-grained score, optimizing with a cross-entropy loss:
\begin{equation}
    \mathcal{L}_{local} = - \text{log} \frac{\text{exp} \; ( S_{f} (\bm x, \bm d))}{\sum_{\bm{d}' \in \mathcal{D}^s} \text{exp} \; (S_{f} (\bm x, \bm {d}' ))}.
\end{equation}

\noindent \textbf{Training.} 
Our EmDepart is optimized with the following loss: 
\begin{equation}
    \mathcal{L} = \mathcal{L}_{global} + \lambda_{local} \mathcal{L}_{local} + \lambda_{var} \mathcal{L}_{var} + \lambda_{div}\mathcal{L}_{div}, 
    \label{eq:loss}
\end{equation}
where $\lambda_{local}$, $\lambda_{var}$, and $\lambda_{div}$ are hyper-parameters. 
The joint training enhances EmDepart to generate multi-view semantic embeddings with information decoupling and accurately align visual and textual space to a common semantic space according to the matching information, significantly improving knowledge transfer.

\setlength{\tabcolsep}{3.6pt}
\renewcommand{\arraystretch}{1.02} 

\begin{table*}[t]
    \caption{
        Comparison with SOTA methods in document-based ZSL. 
        We evaluate methods on documents sourced from Wiki and Wiki+LLM.
        The best results are in \textbf{bold}. 
        Performance gain compared to methods on the same document source is in \dtshow{blue}.
        }
    \vspace{-10pt}
    \centering
    \small
	\setlength{\aboverulesep}{0pt}\setlength{\belowrulesep}{0pt}
	\resizebox{\linewidth}{!}{ %
	\begin{tabular}{ lc lll ccl ccl ccl}
			\toprule
			& & \multicolumn{3}{c}{\textbf{Zero-Shot Learning}} & \multicolumn{9}{c}{\textbf{Generalized Zero-Shot Learning}} \\
			\cmidrule(lr){3-5} \cmidrule(lr){6-14}
			\textbf{Model}	&  \textbf{ \makecell {Auxiliary \\ Information}} & \makecell[c]{\textbf{AWA2}} & \makecell[c]{\textbf{CUB}} & \makecell[c]{\textbf{FLO}} & \multicolumn{3}{c}{\textbf{AWA2}} & \multicolumn{3}{c}{\textbf{CUB}} & \multicolumn{3}{c}{\textbf{FLO}}  \\
			\cmidrule(lr){3-3} \cmidrule(lr){4-4} \cmidrule(lr){5-5} \cmidrule(lr){6-8} \cmidrule(lr){9-11} \cmidrule(lr){12-14}
			\textbf{} & & \makecell[c]{\textbf{T1}} & \makecell[c]{\textbf{T1}} & \makecell[c]{\textbf{T1}} & \textbf{U} & \textbf{S} & \makecell[c]{\textbf{H}} & \textbf{U} & \textbf{S} & \makecell[c]{\textbf{H}} & \textbf{U} & \textbf{S} & \makecell[c]{\textbf{H}}  \\
			
            \midrule
            \texttt{GloVe}~\cite{Glove} & \texttt{CLSN} & 52.1 & 20.4 & 21.6 & 42.1 & 75.3 & 54.0 & 16.2 & 43.6 & 23.6 & 14.4 & 88.3 & 24.8 \\
            \texttt{GloVe}~\cite{Glove} & \texttt{Wiki} & 61.6 & 29.0 & 25.8 & 49.5 & 78.1 & 60.6 & 23.8 & {62.6} & 34.5 & 14.7 & 91.0 & 25.3\\
            \texttt{LongFormer}~\cite{LongFormer} & \texttt{Wiki} & 44.2 & 22.6 & 8.8 & 41.6 & 81.8 & 55.2 & 19.9 & 41.0 & 26.8 & 8.8 & 89.8 & 16.0 \\
            \texttt{MPNet}~\cite{MPNet} & \texttt{Wiki} & 61.8 & 25.8 & 26.3 & 58.0 & 76.4 & 66.0 & 20.6 & 44.3 & 28.2 & 22.2 & {96.7} & 36.1 \\
            \texttt{TF-IDF}~\cite{TF-IDF} & \texttt{Wiki} & 46.4 & 39.9 & 34.0 & 29.6 & 87.6 & 44.2 & 29.0 & 52.1 & 37.3 & 28.9 & 94.8 & 44.3 \\ 
            \texttt{VGSE}~\cite{VGSE} & \texttt{CLSN+IMG} & 69.6 & 37.1 & - & 56.9 & 82.8 & 67.4 & 27.6 & \textbf{70.6} & 39.7 & - & - & -\\ 

            \midrule
            {\texttt{I2DFormer}~\cite{I2DFormer}}
            & \texttt{Wiki} & 
            {76.4} & {45.4} & {40.0} & 
            {66.8} & {76.8} & {71.5} & 
            {35.3} & {57.6} & {43.8} & 
            {35.8} & {91.9} & {51.5} \\ 
            
            {\texttt{I2MVFormer}~\cite{I2MVFormer}}  
            & \texttt{Wiki} & 
            {73.6} & {42.1} & {41.3} & 
            {66.6} & {82.9} & {73.8} &
            {32.4} & {63.1} &  {42.8} & 
            {34.9} & {96.1} & {51.2} \\

            \textbf{\ModelName} \text{(Ours)}    
            & \texttt{Wiki}  & 
           ${81.4}^{\color{blue}\textbf{+5.0}}$  & ${50.2}^{\color{blue}\textbf{+4.8}}$ & ${47.2}^{\color{blue}\textbf{+5.9}}$ & 
            76.0 & 87.8 & ${81.5}^{\color{blue}\textbf{+7.7}}$ & 
            42.6 & 56.3 & ${48.5}^{\color{blue}\textbf{+4.7}}$ & 
            42.7 & \textbf{97.6} & ${59.5}^{\color{blue}\textbf{+8.0}}$ \\

            \midrule

            {\texttt{I2DFormer}~\cite{I2DFormer}} &  \texttt{Wiki$+$LLM}  & 
            77.3 & 47.0 & 43.0 & 
            68.6 & 77.4 & 72.7 & 
            38.5 & 59.3 & 46.7 & 
            40.4 & 80.1 & 53.8 \\  

            {\texttt{I2MVFormer}~\cite{I2MVFormer}} & \texttt{Wiki$+$LLM} & 
            79.6 & 51.1 & 46.2 & 
            75.7 & 79.6 & 77.6 & 
            42.5 & 59.9 & 49.7 & 
            41.6 & 91.0 & 57.1 \\
         
            \textbf{EmDepart} \text{(Ours)} & \texttt{Wiki$+$LLM}  & 
            $\textbf{86.1}^{\color{blue}\textbf{+6.5}}$  & $\textbf{52.8}^{\color{blue}\textbf{+1.7}}$  & $\textbf{53.3}^{\color{blue}\textbf{+7.1}}$  & 
            \textbf{81.4} & \textbf{88.5} & $\textbf{84.8}^{\color{blue}\textbf{+7.2}}$  & 
            \textbf{45.0} & 61.4 & $\textbf{51.9}^{\color{blue}\textbf{+2.2}}$  & 
            \textbf{52.3} & 94.4 & $\textbf{67.3}^{\color{blue}\textbf{+10.2}}$ \\

            \bottomrule
	\end{tabular}
}
	\label{tab:main_exp}
    \vspace{-3pt}
\end{table*}

\subsection{Inference}
\label{sec:inference}
During the inference, a partial score is proposed to filter out unmatched information to measure semantic similarity accurately.

\noindent \textbf{Partial Score Function.}
In Figure ~\ref{fig:Motivation}, it is evident that an image semantically matches a subset of the semantic concepts within the document.
Similarly, we limit the similarity computation solely to document-image semantic pairs with the highest $p$ similarity values, where $p < k$.
Specifically, we select the top $p$ similarity values between each visual view embedding and all textual view embeddings, yielding a total of $p \times k$ pairs.
Subsequently, a similar process is applied to each textual view embedding, resulting in the selection of $p \times p$ pairs. 
We denote this process as $\text{TopCos}(\bm x, \bm d, p)$ function, effectively filtering out unmatched information between documents and images.
The smooth chamfer is then used to compute the similarity among these chosen pairs:
\begin{equation}
    S_p(\bm x, \bm d) =  S_{sc}(\text{TopCos}(\bm x, \bm d, p)).
    \label{eq: partial_score}
\end{equation}
\noindent \textbf{Inference.} 
Given an input image $\bm x$, we obtain a prediction $\hat {\bm y}$ that yields the highest partial score $S_p (\bm x, {\bm d}')$ among unseen classes for ZSL, \textit{i.e.}, $\mathcal{D} = \mathcal{D}^u$, and among both seen and unseen classes for GZSL, \textit{i.e.}, $\mathcal{D} = \mathcal{D}^s \cup \mathcal{D}^u$:
\begin{equation}
    \hat {\bm y} = \argmax_{{\bm d}' \in \mathcal{D}} S_{p}(\bm x, {\bm d}').
\end{equation}


\section{Experiments}

\noindent \textbf{Datasets.} 
We evaluate on three benchmark datasets, \textit{i.e.}, 
a coarse-grained dataset Animals with Attributes2 (AWA2)  \citep{AWA2},
two fine-grained datasets Caltech-USCD Birds-200-2011 (CUB) ~\citep{CUB_and_attribute_provide} and Oxford Flowers (FLO) ~\citep{FLO}.
The seen-unseen class division is from Proposed Split ~\cite{AWA2}.
We use documents instead of human-annotated attributes in datasets as auxiliary information.

\noindent \textbf{Evaluation Metrics.} 
Following ~\citep{AWA2}, we measure the average per-class top-1 accuracy (T1) on unseen classes for ZSL.
For GZSL, we present the per-class mean accuracy on seen ($S$) and unseen classes ($U$) as well as their harmonic mean $H = 2 \times U \times S / (U + S) $.

\noindent \textbf{Implementation Details.}
Similar to ~\cite{I2DFormer, I2MVFormer}, we utilize the ViT-B/16 ~\citep{vit} pre-trained on ImageNet 1K ~\citep{ImageNet} as the visual backbone.
If not noted otherwise, we show the performance of ChatGPT ~\citep{ChatGPT} as LLMs.
Hyperparameters are optimized by grid search in the validation split.
Once the hyperparameters are confirmed, we merge the validation with the training split to obtain the performance on the test split.
We also apply calibrated stacking ~\citep{calibrated_stack} for GZSL to trade-off calibration degrees, reducing the bias towards seen classes.
More details are available in the supplementary.



{
    \setlength{\tabcolsep}{5pt}
    \renewcommand{\arraystretch}{1.02} 
    
    \begin{table}[t]
    \setlength{\aboverulesep}{0pt}\setlength{\belowrulesep}{0pt}
    \centering
    \caption{\textbf{Comparison with set-based embedding methods. Performance improvement after adding our losses is in \dtshow{blue}.}}
    \vspace{-10pt}
    
    \label{tab:Ablation_useful_loss}
    
    \resizebox{0.8\linewidth}{!}{
        \begin{tabular}{l ccc ccc}
        \toprule
        \multirow{2}{*}{\textbf{Model}} & \multicolumn{2}{c}{\textbf{AWA2}} & \multicolumn{2}{c}{\textbf{CUB}} & \multicolumn{2}{c}{\textbf{FLO}} \\
        \cmidrule(lr){2-3} \cmidrule(lr){4-5} \cmidrule(lr){6-7}
        & \textbf{T1} & \textbf{H} & \textbf{T1} & \textbf{H} & \textbf{T1} & \textbf{H} \\
        \midrule

        \texttt{TVMM}~\citep{set_based_embed_NIPS_2022} & 77.4 & 74.4 & 41.6 & 43.1 & 42.3 & 54.2 \\
        $+\mathcal{L}_{var} + \mathcal{L}_{div}$ & 81.0  & {77.5}  & 45.6  & 47.4  & 46.8  & 59.5  \\
        \textbf{Gain} & \plus{+3.6} & \plus{+3.1} & \plus{+4.0} & \plus{+4.3} & \plus{+4.5} & \plus{+5.3}\\
        \midrule
        
        \texttt{S-Chamfer}~\citep{set_based_embed_CVPR_2023} & 81.5 & 80.6 & 45.6 & 45.2 & 43.5 & 57.3 \\
        $+ \mathcal{L}_{var} + \mathcal{L}_{div}$ & {84.0} & {82.9} & 49.1 & {49.9} & {48.9} & {63.6} \\
        \textbf{Gain} & \plus{+2.5} & \plus{+2.3} & \plus{+3.5} & \plus{+4.7} & \plus{+5.4} & \plus{+6.3}\\
        \midrule
        
        \textbf{\ModelName}(Ours) & \textbf{86.1} & \textbf{84.8}& \textbf{52.8} & \textbf{51.9} & \textbf{53.3}& \textbf{67.3}  \\
        \bottomrule
        \end{tabular}}
        \vspace{-9pt}
    \end{table}
}

\subsection{Comparing with the SOTA Methods}


\noindent \textbf{Comparison with SOTA in Document-based ZSL.}
In Table ~\ref{tab:main_exp}, we compare our EmDepart with state-of-the-art (SOTA) methods in document-based ZSL. 
For a fair comparison, we evaluate methods with the same text and image perceivers.
EmDepart outperforms previous methods across all metrics (T1 and H) regarding ZSL and GZSL settings on all datasets.
It confirms that modeling the partial association is beneficial for accurate semantic alignment.
The previous SOTA methods ~\citep{I2DFormer, I2MVFormer} align complete semantics in documents with images, thus hindering knowledge transfer.

\noindent \textbf{Wiki vs Wiki+LLM Documents.}
With Wiki documents, EmDepart achieves optimal performance compared to previous methods.
Notably, EmDepart with Wiki outperforms SOTA methods with Wiki and LLM regarding T1 and H on AWA2 and FLO.
It confirms that modeling the partial association is significant for knowledge transfer.
Besides, we see a performance improvement from Wiki to Wiki+ChatGPT.
This is because visual descriptions generated by LLMs enrich semantic information in less-described classes.

\noindent \textbf{Comparison with SOTA in Set-based Embedding Methods.}
In Table ~\ref{tab:Ablation_useful_loss}, we replace SDM with TVMM ~\citep{set_based_embed_NIPS_2022} and S-Chamfer ~\citep{set_based_embed_CVPR_2023}, the SOTA set-based embedding methods.
Since documents are at the category level, it is challenging to rely solely on model architecture to produce view embeddings with little redundant information. 
SDM achieves the optimal performance by incorporating \mathDivLoss~ and \mathVarLoss ~ to enhance information difference among view embeddings.
Similarly, TVMM and S-Chamfer improve performance after adding \mathDivLoss~ and \mathVarLoss ~, facilitating differences between embeddings.

\begin{figure}[t]

\centering
\resizebox{0.9\linewidth}{!}
{
    \vspace{-10pt}
	\subfloat[w/o $\mathcal{L}_{div}$ and $\mathcal{L}_{var}$.]{\includegraphics[width=0.135\textwidth]{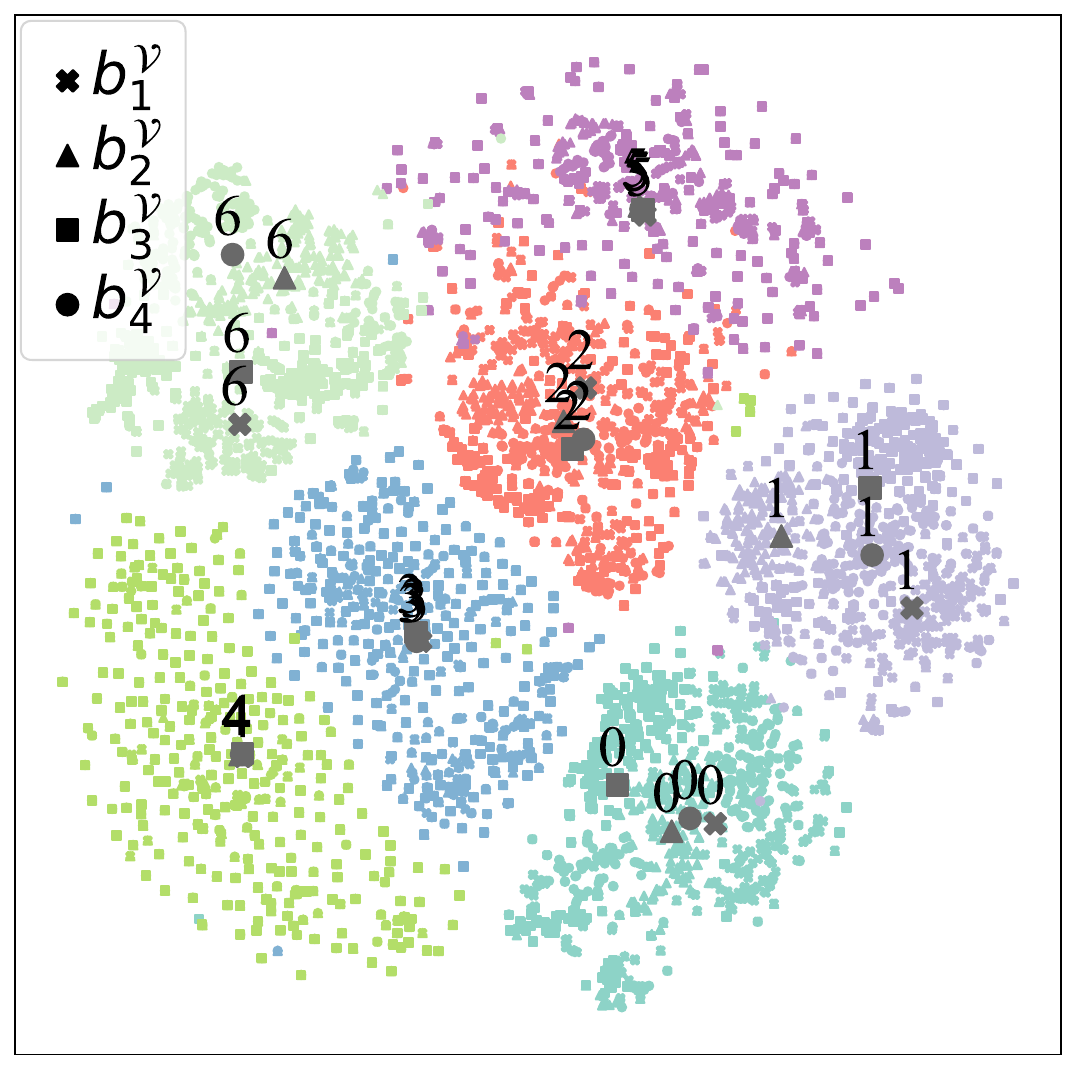}}
    \hfill
    \subfloat[w/o $\mathcal{L}_{div}$.]{\includegraphics[width=0.135\textwidth]{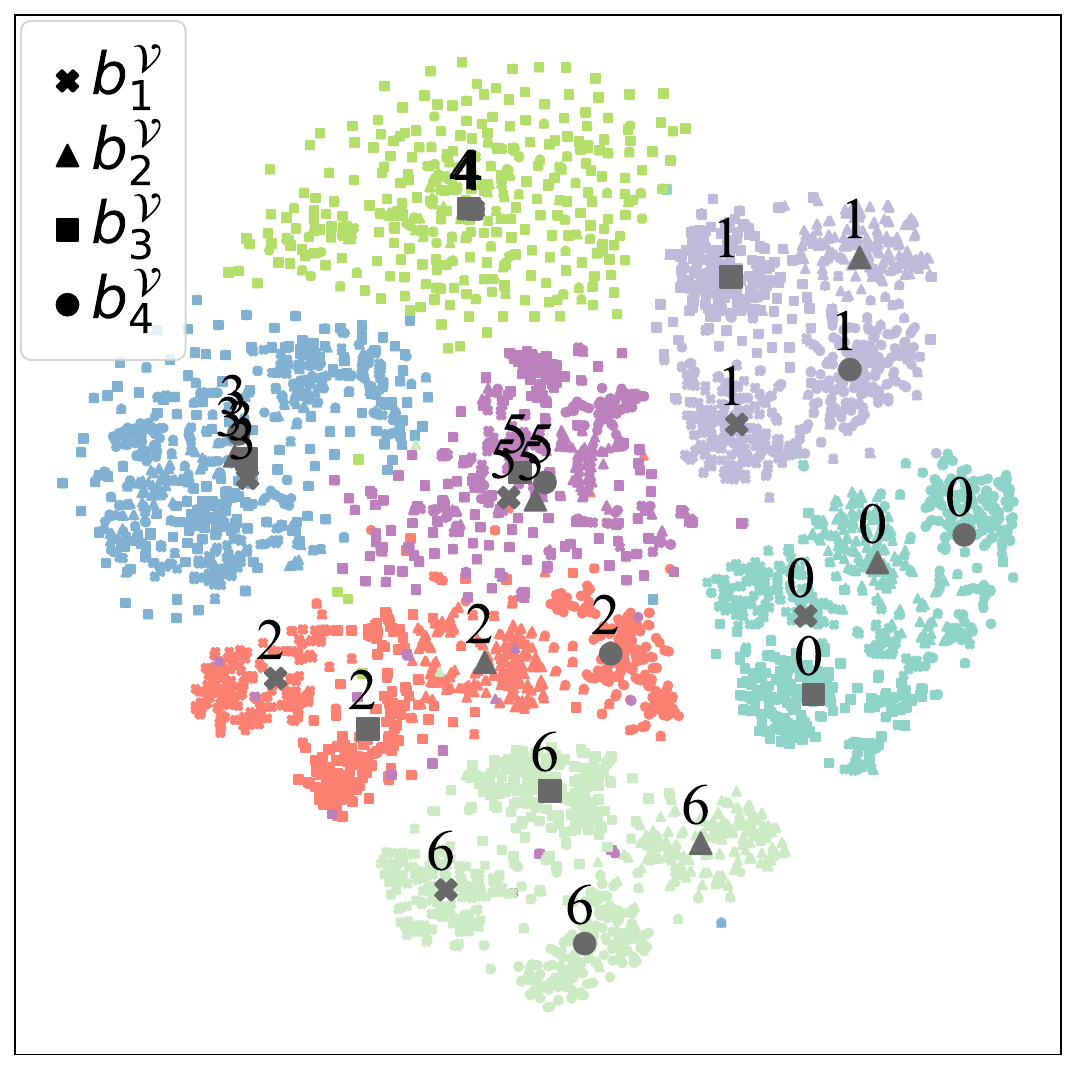}}
    \hfill
    \subfloat[EmDepart.]{\includegraphics[width=0.135\textwidth]{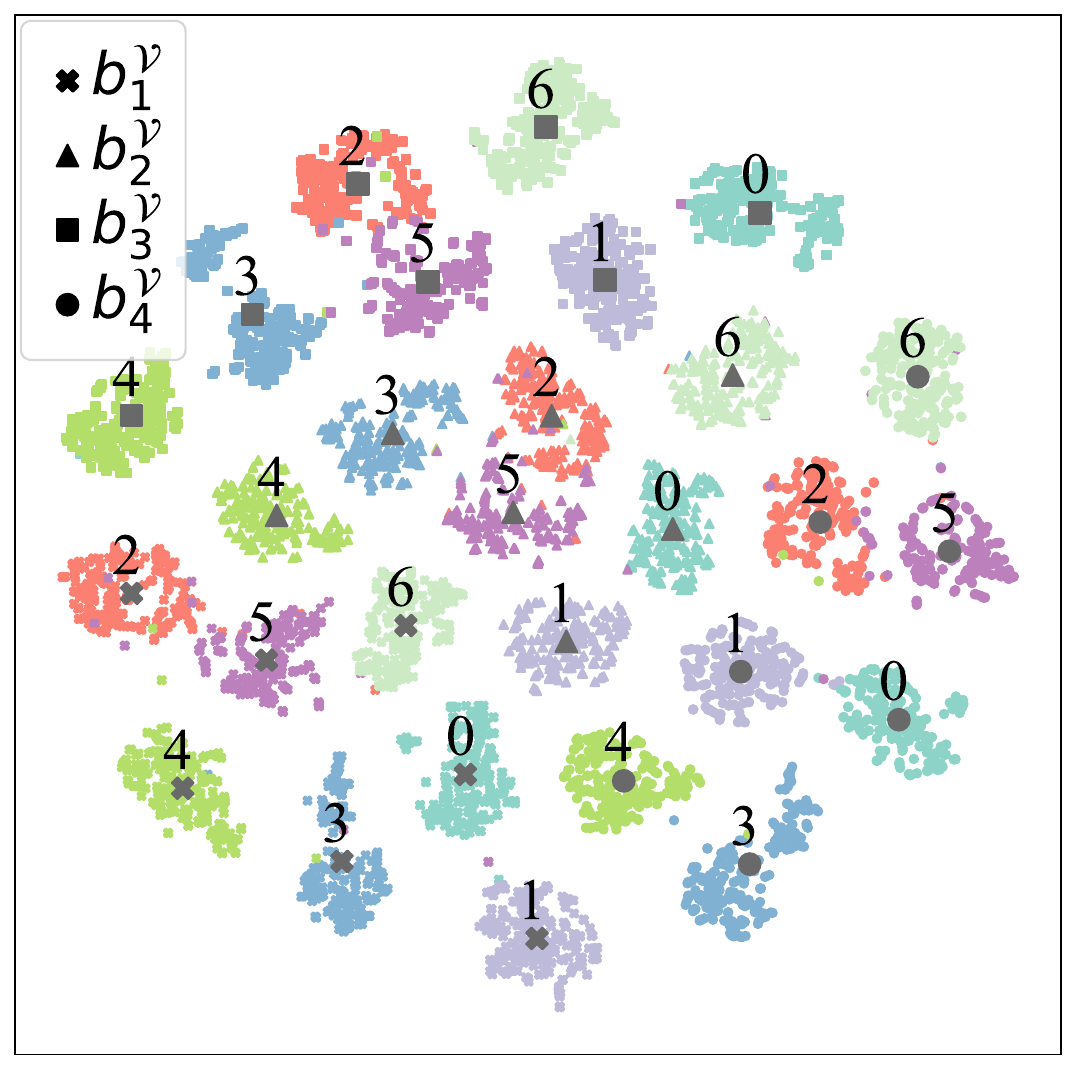}}
}
\vspace{-6pt}
\caption{Analysis of feature collapse. Each number denotes a class (same color), and each shape denotes one of the view embeddings.
With the addition of ~\mathVarLoss~ and ~\mathDivLoss~, information differences between embeddings gradually increase.}
\Description{}
\label{fig:t_sne_feature_collapse}
\vspace{-10pt}
\end{figure}

\begin{figure}[t]
\centering
    \vspace{-10pt}
\resizebox{0.9\linewidth}{!}
{
	\subfloat[Information Difference.]{\includegraphics[width=0.2\textwidth]{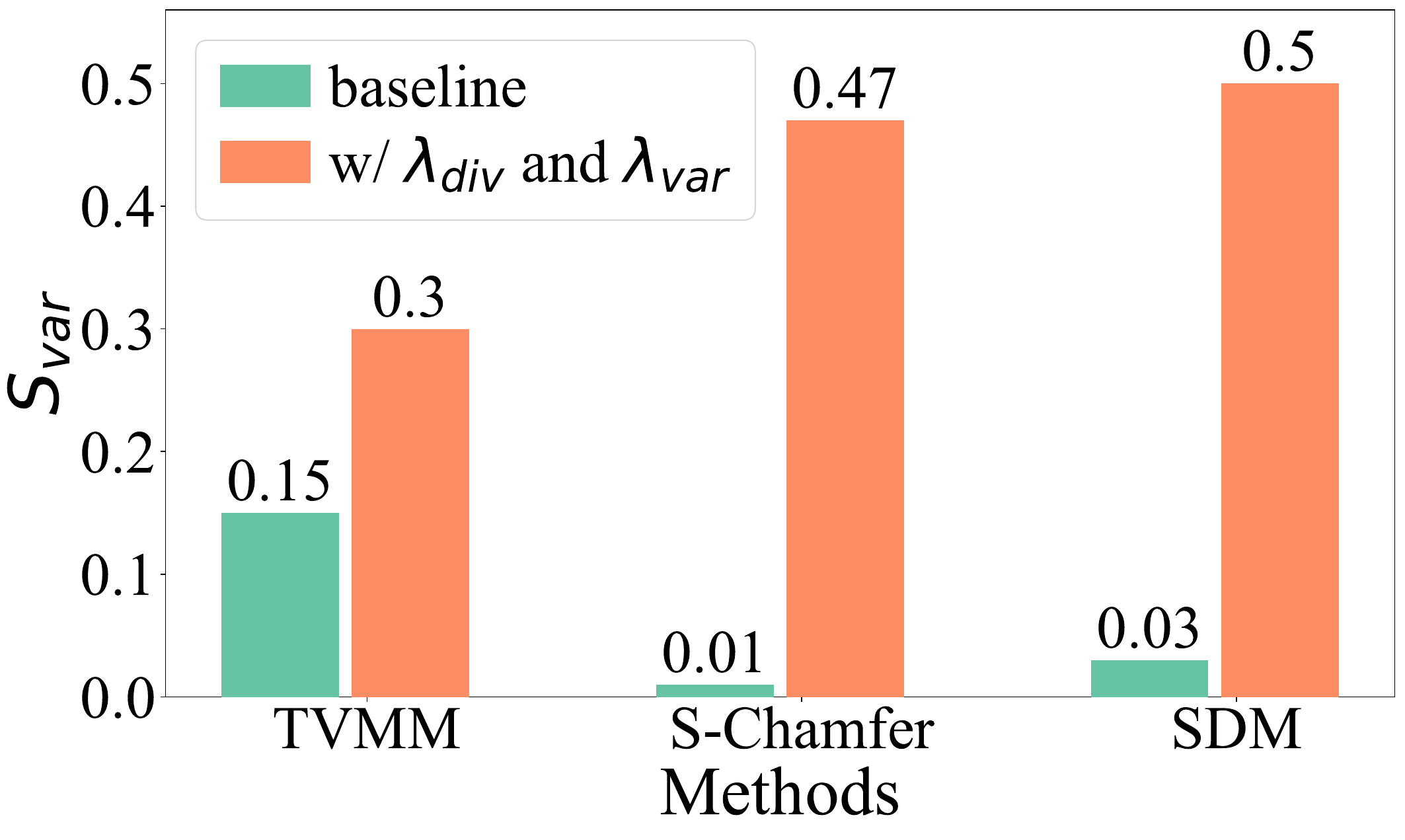}}
	\hfill
	\subfloat[Performance.]{\includegraphics[width=0.2\textwidth]{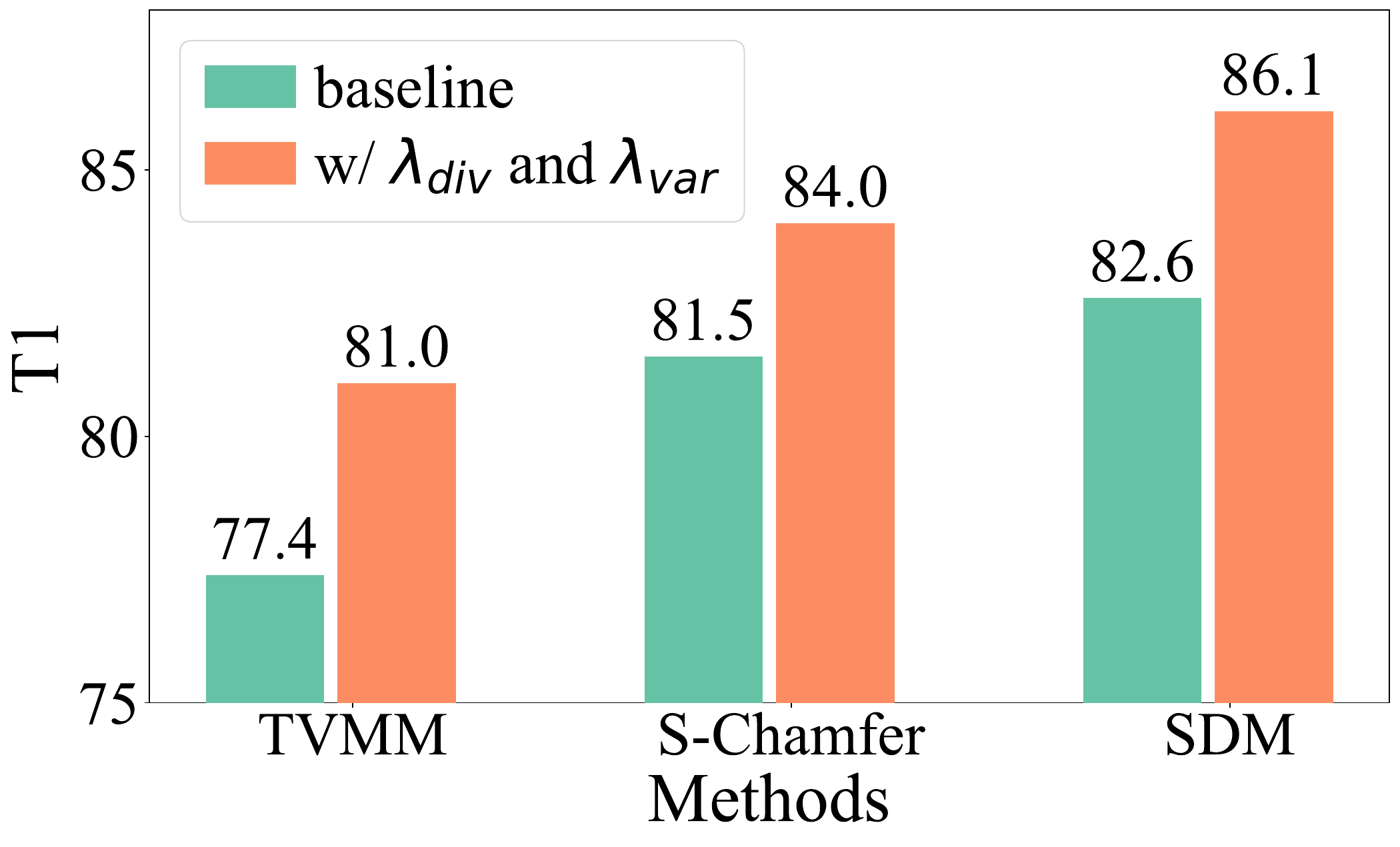}}
}
\vspace{-6pt}
\caption{Analysis of baseline and model after adding our losses. The larger $S_{var}$ denotes more distinct between embeddings, and $S_{var} = 0$ denotes embeddings are all the same.
}
\Description{}
\label{fig:feature_collapse}
\vspace{-12pt}
\end{figure}

\begin{figure*}[t]
\vspace{-10pt}
\centering
\resizebox{0.86\linewidth}{!}
{
	\subfloat[]{\includegraphics[width=0.48\textwidth]{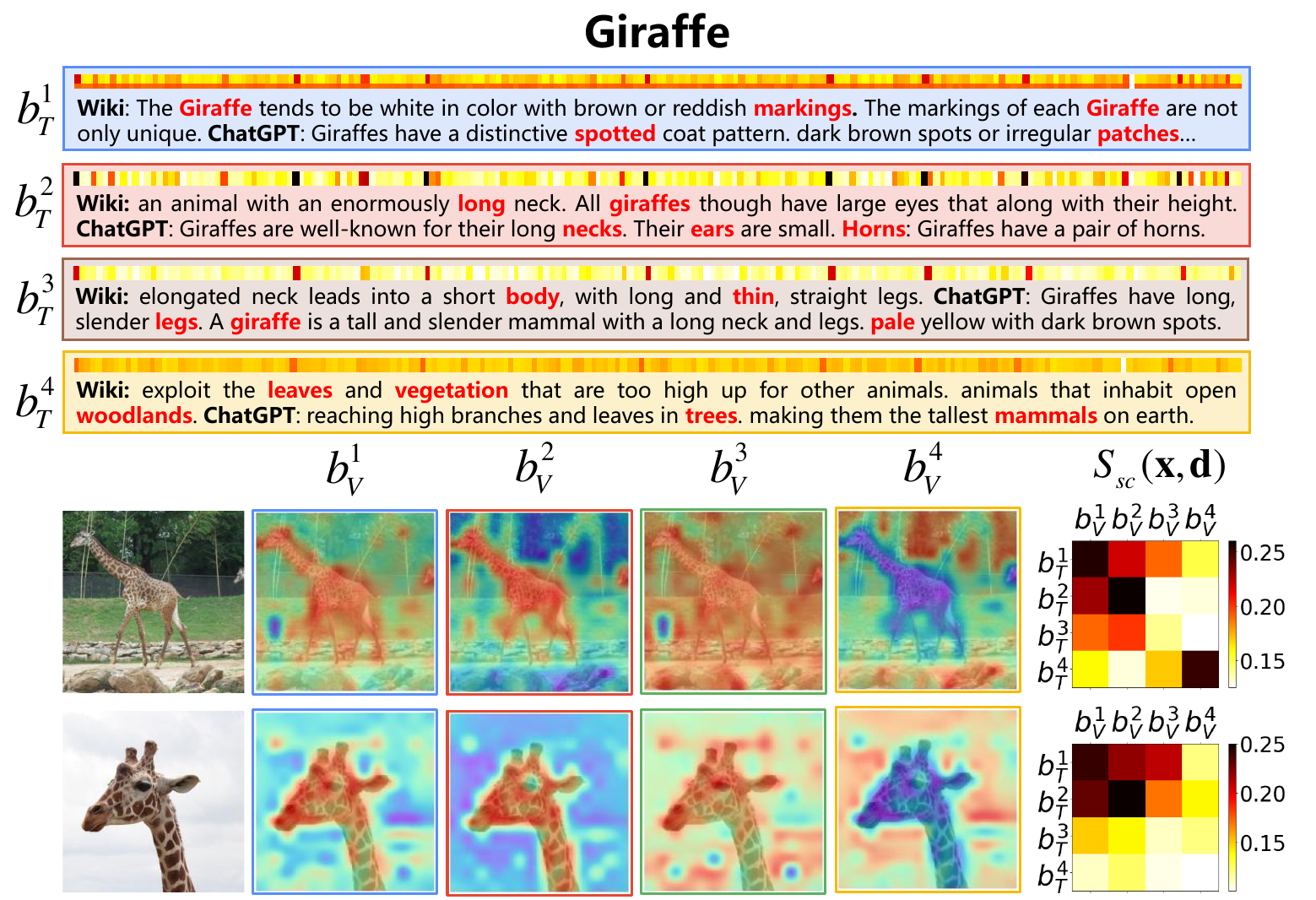}}
	\hfill
	\subfloat[]{\includegraphics[width=0.48\textwidth]{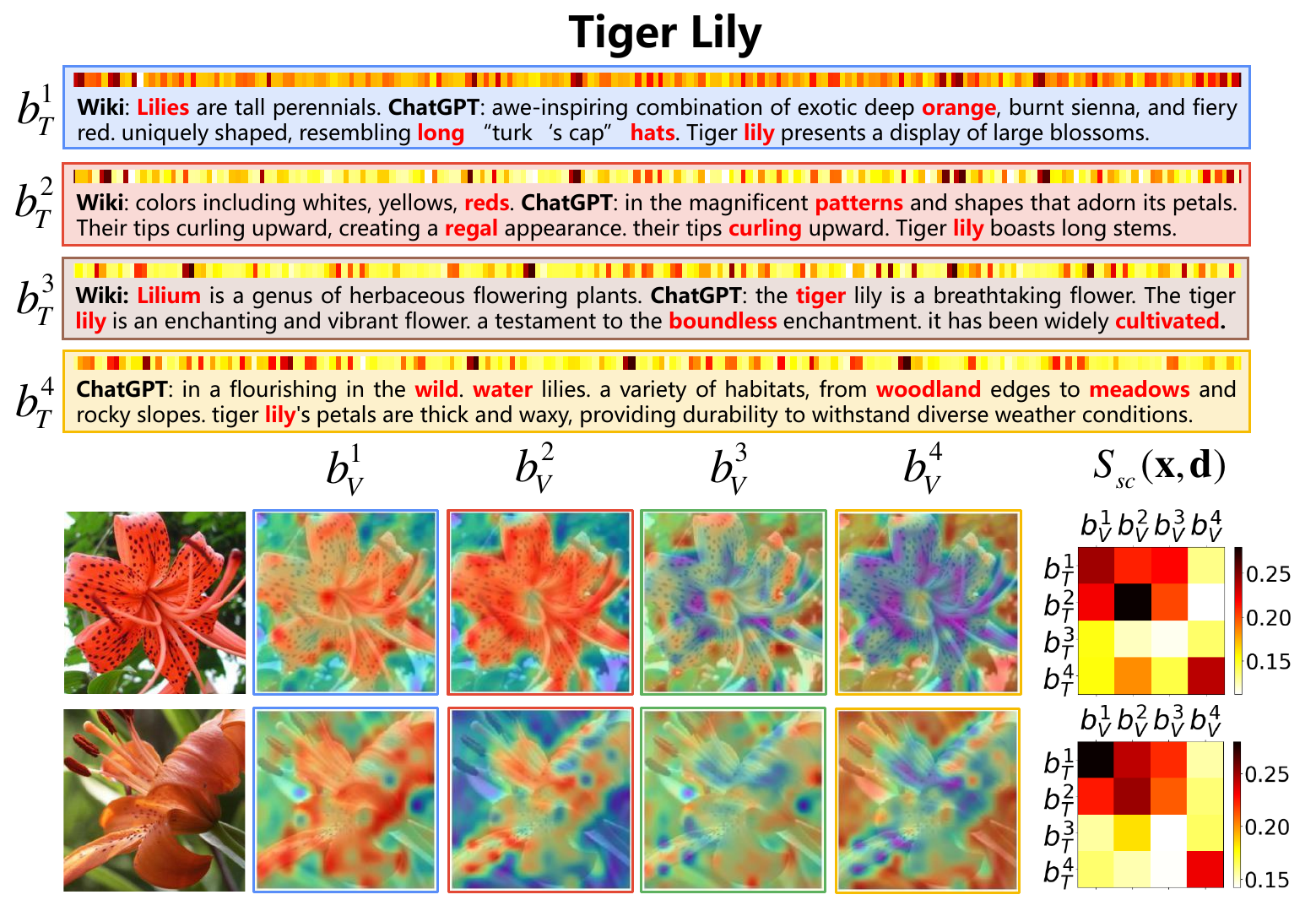}}
}
\vspace{-10pt}
\caption{Partial association analysis on AWA2 and FLO datasets. 
We present attention maps for each visual and textual view embedding, the top 5 most attended words (in \textcolor{red}{red}) with nearby words, and smooth chamfer score. In attention maps, darker colors represent larger attention values. Our EmDepart achieves accurate semantic alignment on the matching information.}
\Description{}
\label{fig:visualize_partial}
\vspace{-5pt}
\end{figure*}

\subsection{Analysis of Feature Collapse}
In Figure \ref{fig:t_sne_feature_collapse}(a), we show the feature collapse between view embeddings, \textit{i.e.}, embeddings have little variance, resulting in information redundancy.
It is harmful to model the partial association with the same embeddings.
To solve this issue, we introduce $\mathcal{L}_{var}$ to make each embedding attend to distinct local details and $\mathcal{L}_{div}$ to penalize embeddings orthogonal to others.
In Figure \ref{fig:t_sne_feature_collapse}(a-c), view embeddings carry more distinct information (at more different positions) with the $\mathcal{L}_{var}$ and $\mathcal{L}_{div}$. 
Quantitatively, we introduce the circular variance $S_{var} = 1 - \Vert \sum_{\bm b \in \bm B} \bm{b} / | \bm B | \Vert_2 $ to analyze the information difference of view embeddings. 
TVMM ~\cite{set_based_embed_NIPS_2022} and S-Chamfer ~\cite{set_based_embed_CVPR_2023} lead to feature collapse under category-level corpora in Figure \ref{fig:feature_collapse}.
After adding $\mathcal{L}_{var}$ and $\mathcal{L}_{div}$, we improve these methods performance and increase the information difference among view embeddings.

\subsection{Analysis of Partial Association}

In Figure ~\ref{fig:visualize_partial}, we qualitatively show that our model learns the interpretable partial semantic association.
It contains the visual-semantic decomposition to offer basic semantic concepts and partial semantic alignment according to the matching information.

\noindent \textbf{Visual-Semantic Decomposition.}
We see that different view embeddings focus on distinct information in each modality.
Utilizing the giraffe as an example, EmDepart focuses on appearance (in $b_{V}^1$ and $b_{V}^2$), habitat (in $b_{V}^4$), and global information (in $b_{V}^3$) for the visual side.  Similarly, there are textual descriptions on color (in $b_{T}^1$), appearance and shape (in $b_{T}^2$ and $b_{T}^3$), and habitat (in $b_{T}^4$). 
This verifies that the SDM decomposes semantics from images and documents and generates multi-view semantic embeddings.

\noindent \textbf{Partial Semantic Alignment.}
On the similarity matrix of $S_{sc}(\boldsymbol{x}, \boldsymbol{d})$, we observe the accurate semantic alignment between document and image.
In particular, since the second giraffe image does not represent the visual content about habitat and body, $(b_{V}^2, b_{T}^3)$ and $(b_{V}^4, b_{T}^4)$ has a high similarity in the first image but low in the second. 
In the first red tiger lily, $(b_{V}^2, b_{T}^2)$ has the highest score, while $(b_{V}^1, b_{T}^1)$ has the highest score in the second orange one.
This is consistent with the fact that $b_{T}^1$ pays more attention to ``orange'' and $b_{T}^2$ focuses more on ``red''. 

\subsection{Ablation Study}
We ablate key components of EmDepart in Table ~\ref{tab:Ablation_Module}.
For all models, we leverage documents from Wiki+LLM as auxiliary information. 

\noindent \textbf{Ablation on Loss Functions.} 
In row b), we see a significant drop in performance on CUB and FLO.
This is due to the lack of interaction between patches and words, which offers discriminative information for fine-grained classification.
The performance of removing \mathDivLoss~ decreases more than ~\mathVarLoss~ in rows c) and d).
This is due to \mathVarLoss~ constraining the variance in attention blocks, which means feature collapse may exist after the MLP projection in SDM.
Row e) shows a further decrease in performance, which indicates the complementary of \mathDivLoss ~ and ~\mathVarLoss.
Row f) achieves the worst performance, further verifying the effectiveness of our losses. 

{
    \setlength{\tabcolsep}{5pt}
    \renewcommand{\arraystretch}{1.02} 
    
    \begin{table}[t]
    \setlength{\aboverulesep}{0pt}\setlength{\belowrulesep}{0pt}
    \centering
    \vspace{-5pt}
    \caption{\textbf{Ablation of key components in EmDepart}}
    \vspace{-10pt}
    \label{tab:Ablation_Module}
        \resizebox{0.8\linewidth}{!}
        {
        \begin{tabular}{l ccc}
        \toprule
        \multirow{2}{*}{\textbf{Model}} & \textbf{AWA2} & \textbf{CUB} & \textbf{FLO}\\
        \cmidrule(lr){2-2} \cmidrule(lr){3-3} \cmidrule(lr){4-4} 
        & \textbf{T1} & \textbf{T1} & \textbf{T1} \\
        \midrule
         \texttt{a)\quad}\texttt{full model} &  \textbf{86.1} & \textbf{52.8} & \textbf{53.3} \\

        \midrule
        \textbf{Ablation on Loss Function} & & & \\
         \texttt{b)\quad w/o $\mathcal{L}_{local}$} &  85.8 & 45.9 & 41.7 \\
         \texttt{c)\quad w/o $\mathcal{L}_{div}$} &  83.5 & 47.7 & 41.5 \\
         \texttt{d)\quad w/o $\mathcal{L}_{var}$} &  85.5 & 50.1 & 49.9 \\
         \texttt{e)\quad w/o $\mathcal{L}_{div} + \mathcal{L}_{var}$} & 82.6 & 47.5 & 39.3 \\
         \texttt{f)\quad w/o $\mathcal{L}_{local} + \mathcal{L}_{div} + \mathcal{L}_{var}$} & 80.1 & 45.4 & 37.2 \\

        \midrule
        \textbf{Ablation on Score Function} & & & \\
         \texttt{g)\quad w/o Partial Score in Eq.\ref{eq: partial_score}} &  85.7 & 52.6 & 53.0 \\
         \texttt{h)\quad w/ average distance in Eq.\ref{eq: smooth_chamfer}} &  80.0 & 39.4 & 45.7 \\
         \texttt{i)\quad w/ maximum distance in Eq.\ref{eq: smooth_chamfer}} &  82.2 & 45.4 & 44.8 \\

        \midrule
        \textbf{Ablation on Module} & & & \\
         \texttt{j)\quad w/o global feature in Eq.\ref{eq: concat_global}} & 71.6 & 37.7 & 39.6 \\
         \texttt{k)\quad w/o \nameViewModuleName} &  79.7 & 46.0 & 45.1 \\
         \texttt{l)\quad w/o residual connection} &  81.4 & 49.7 & 48.3 \\

        \bottomrule
        \end{tabular}
        }
        \vspace{-14pt}
    \end{table}
}

\noindent \textbf{Ablation on Score Functions.} 
The performance degrades in row g) due to the partial score filtering out unmatched semantic information, measuring similarity accurately in the inference.
Rows h) and i) ablate Smooth Chamfer with the average distance in ~\citep{set_based_embed_CVPR_2021} and maximum distance in ~\citep{set_based_embed_CVPR_2019}.
Smooth Chamfer performs better because it overcomes problems posed by sparse supervision in maximum distance and feature collapse in average distance.

\begin{figure*}[t]
\vspace{-10pt}
\centering
{
	\subfloat[]{\includegraphics[width=0.21\textwidth]{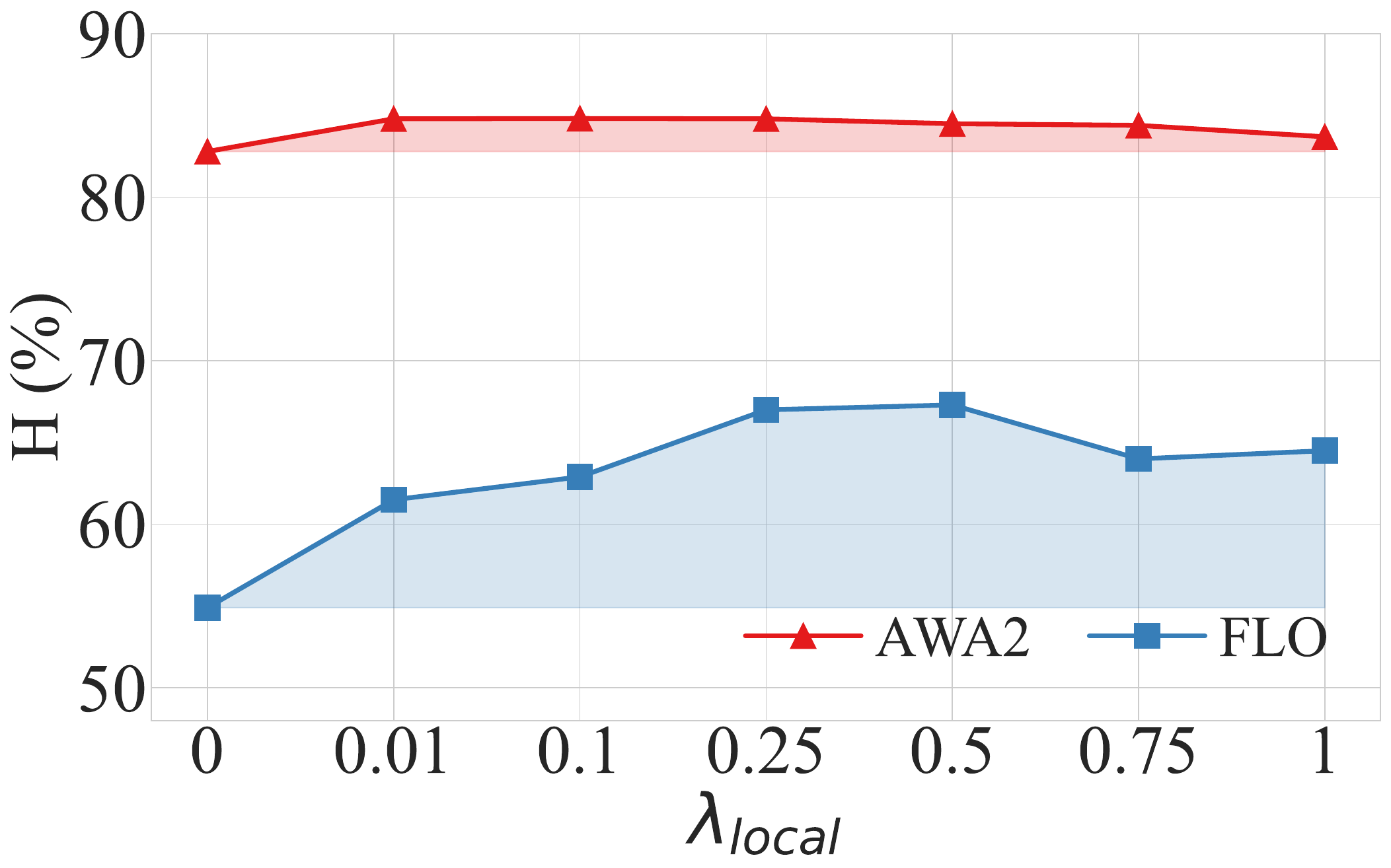}}
	\hfill
	\subfloat[]{\includegraphics[width=0.21\textwidth]{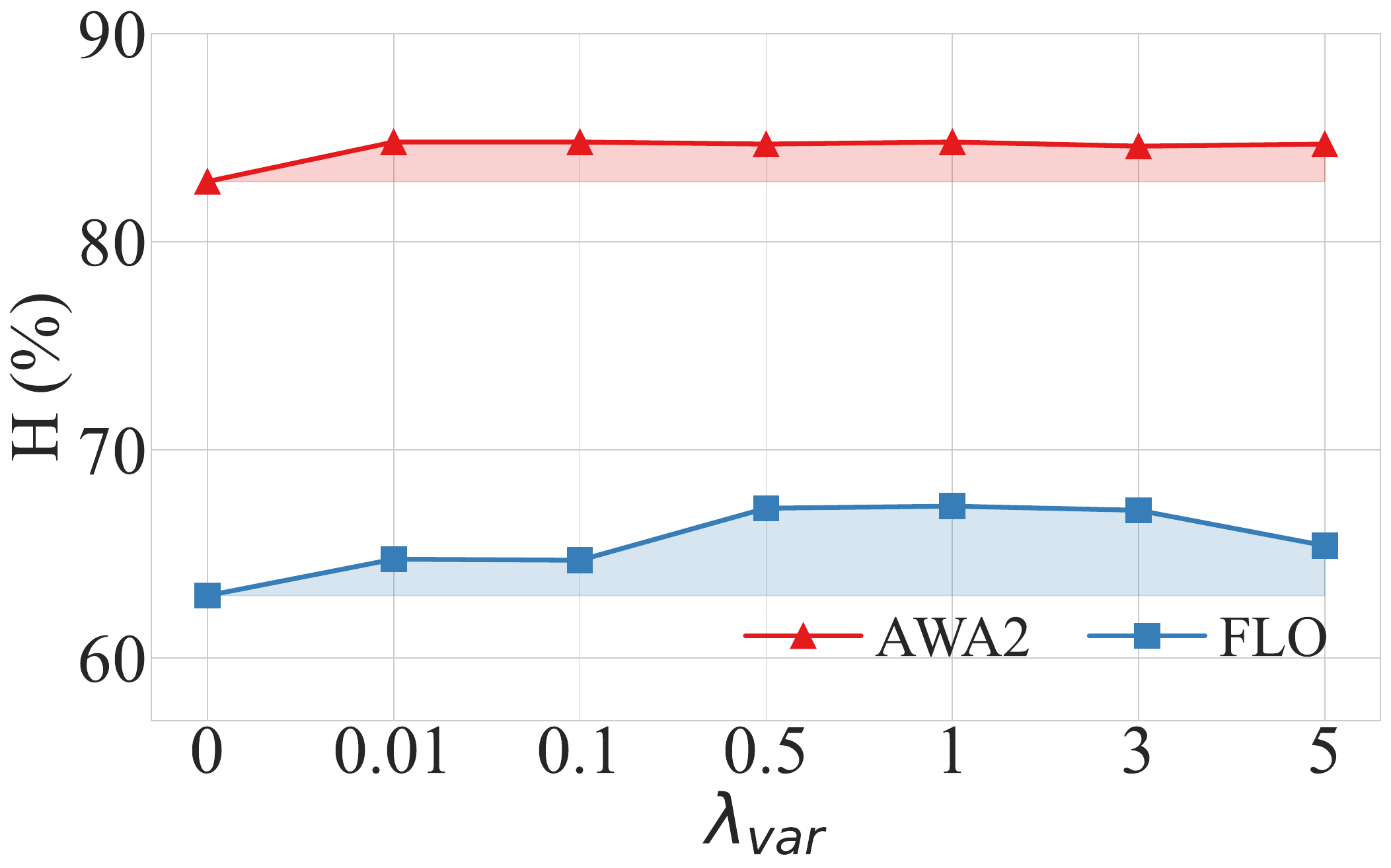}}
	\hfill
	\subfloat[]{\includegraphics[width=0.21\textwidth]{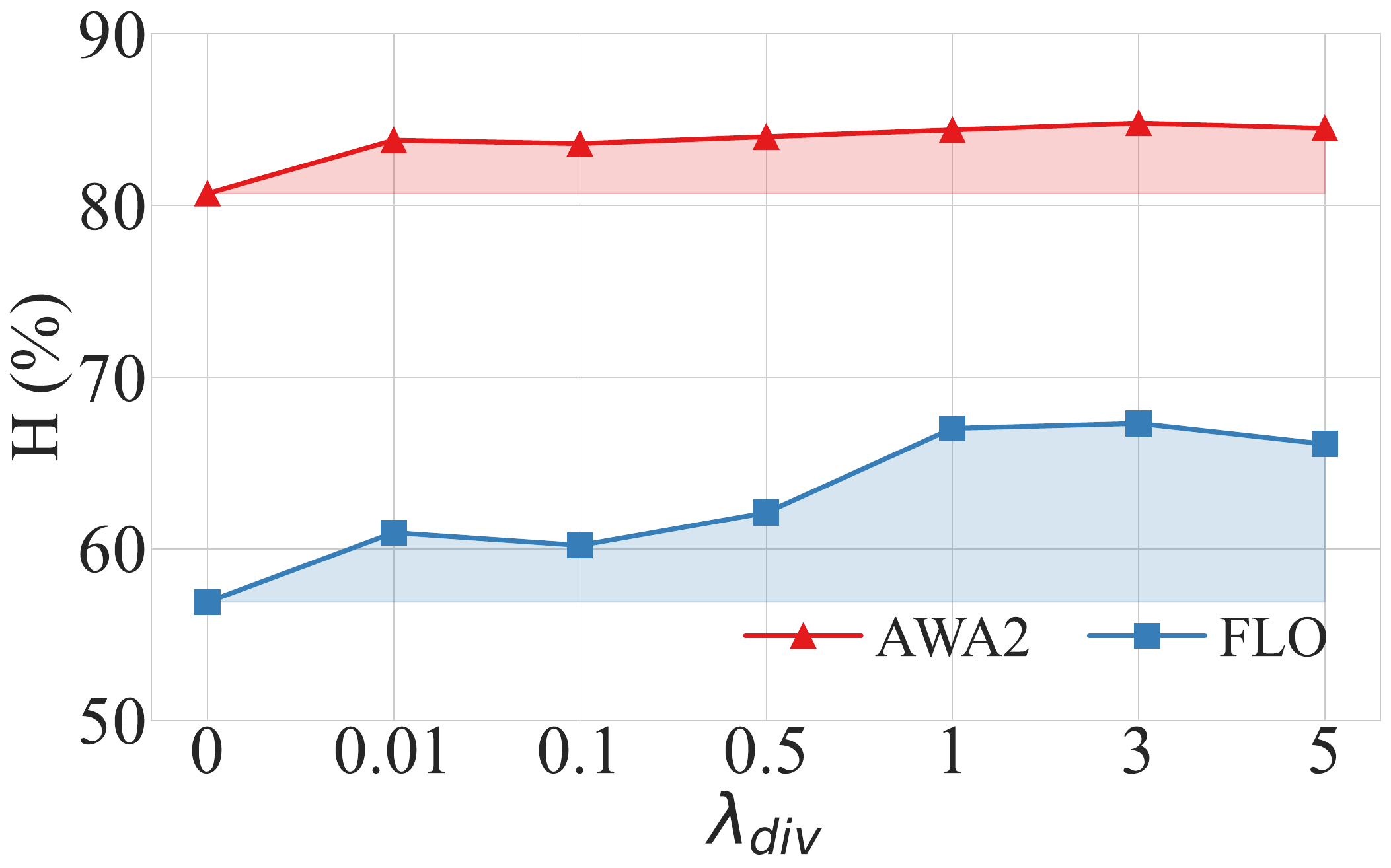}} 
    \hfill
	\subfloat[]{\includegraphics[width=0.21\textwidth]{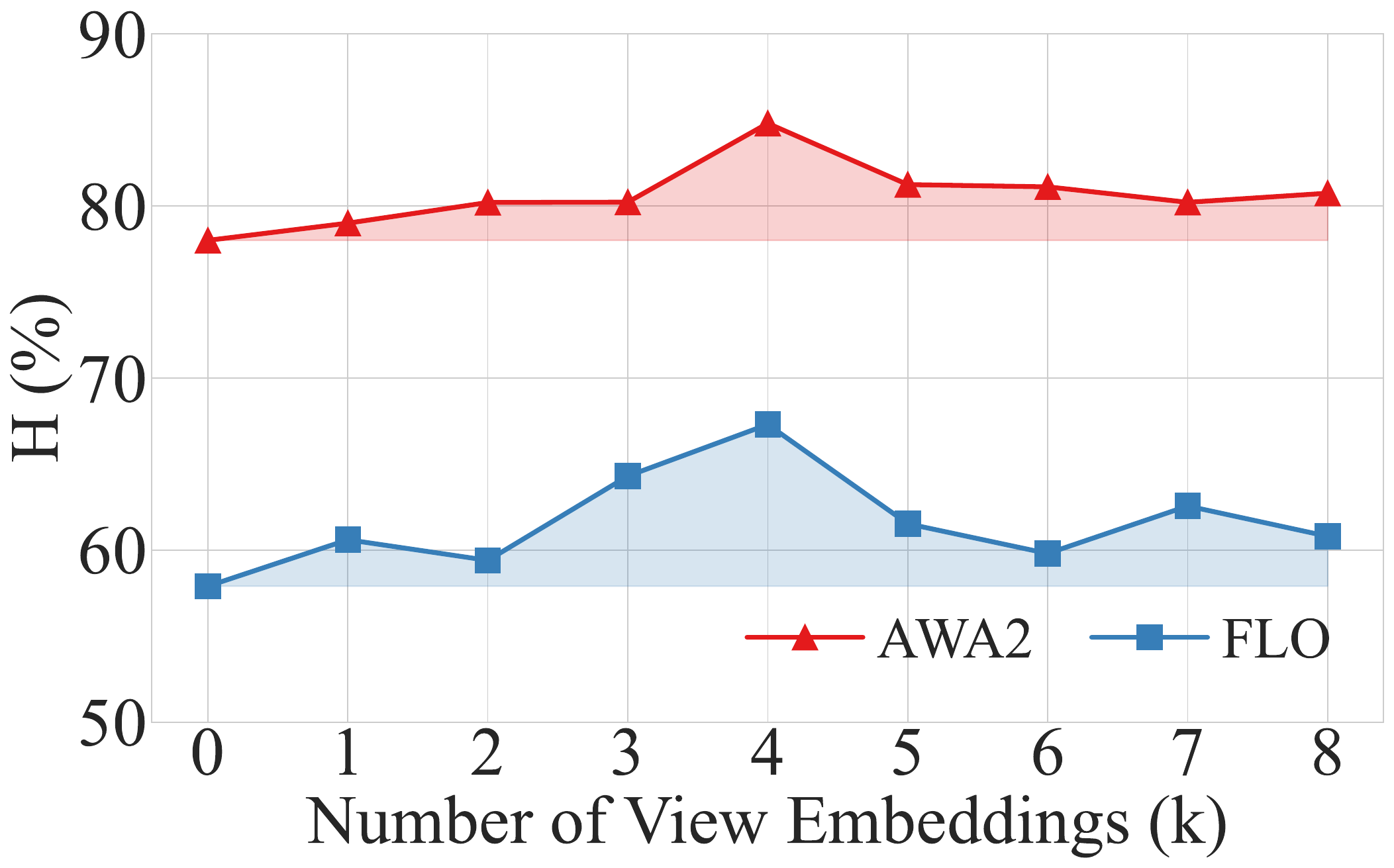}}
}
\vspace{-10pt}
\caption{Effect of loss weights (a-c) and number of view embeddings (d) on coarse-grained AWA2 and fine-grained FLO datasets. The shaded area indicates the performance improvement compared to hyperparameters set as 0.}
\Description{}
\label{fig:loss_influence}
\vspace{-10pt}
\end{figure*}

\noindent \textbf{Ablation on Proposed Modules.} 
In row j), we remove the global feature in Eq.\ref{eq: concat_global}. 
It performs worse as view embeddings using only local features may introduce a large variance, leading to overfitting on seen classes. 
In contrast, global features in Eq.\ref{eq: concat_global} ensure that the variance remains within a controlled range.
In row k), we show the result of removing the SDM and leveraging the global feature to align entire semantics of documents to images like  ~\cite{I2DFormer, I2MVFormer}.
The performance decreases due to the suboptimal semantic alignment, ignoring the partial association.
Besides, performance drops in row l) when the visual perceiver lacks a residual connection, which preserves the original visual knowledge inherent to ViT ~\citep{vit}, a crucial factor for knowledge transfer.

\noindent \textbf{Ablation of Different LLMs.}
In Table \ref{tab:ablation_LLMs}, we show the effect of different LLMs, consistently improving the performance compared to Wiki documents. 
It verifies the effectiveness of enriching less-described documents.
The performance in FLO improves significantly due to the lack of detailed descriptions for most classes in Wiki.
The ChatGPT \cite{ChatGPT} achieves the best result, which generates more detailed descriptions with rich semantics. 

{
    \setlength{\tabcolsep}{3.5pt}
    \renewcommand{\arraystretch}{1.05} 
    
    \begin{table}[t]
    \setlength{\aboverulesep}{0pt}\setlength{\belowrulesep}{0pt}
    \centering
    \caption{Ablation of LLMs. 
    The error bars are obtained from three different documents generated by LLMs.
    }
    \vspace{-10pt}
    \label{tab:ablation_LLMs}
    \resizebox{0.9\linewidth}{!}
        {
        \begin{tabular}{l ll ll}
        \toprule
        \textbf{Auxiliary} &  \multicolumn{2}{c}{\textbf{AWA2}} & \multicolumn{2}{c}{\textbf{FLO}} \\
        \cmidrule(lr){2-3} \cmidrule(lr){4-5} 
        \textbf{Information} & \textbf{T1} & \textbf{H} & \textbf{T1} & \textbf{H} \\
        \midrule
        \texttt{Wiki} & 81.4 & 81.5 & 47.2 & 59.5  \\
        \texttt{Wiki$+$GPT3}~\cite{prompt_1} & $82.3_{\pm0.45}$ & $82.2_{\pm0.61}$ & $53.2_{\pm0.78}$ & $65.5_{\pm0.87}$    \\
        \texttt{Wiki$+$LLaMa2}~\cite{llama2} & $82.1_{\pm0.37}$ & $82.8_{\pm0.27}$ & $49.5_{\pm0.65}$ & $62.8_{\pm0.61}$   \\
        \texttt{Wiki$+$ChatGPT}~\cite{ChatGPT} & $\textbf{86.1}_{\pm0.16}$ & $\textbf{84.8}_{\pm0.29}$ & $\textbf{53.3}_{\pm0.41}$ & $\textbf{67.3}_{\pm0.82}$  \\
        \bottomrule
        \end{tabular}}
        \vspace{-11pt}
    \end{table}
}

\subsection{Impact of Hyperparameters}
\noindent \textbf{Effect of Loss Weights.} 
In Figure ~\ref{fig:loss_influence}(a-c), as the ~\mathLocalWeight, \mathVarWeight, and \mathDivWeight~ increase, the model consistently improves performance compared to the value set as 0. 
It confirms the effectiveness of our losses.
In Figure ~\ref{fig:loss_influence}(a), the performance rises consistently on two datasets, which verifies that fine-grained interactions are essential for knowledge transfer.
In Figure \ref{fig:loss_influence}(b-c), we see a uniform performance improvement in AWA2 with the increase of \mathVarWeight~ and \mathDivWeight.
It verifies that \mathVarLoss~ and \mathDivLoss~ aid the model to generate multi-view semantic embeddings with information decoupling, facilitating the partial semantic alignment. 
The results in FLO demonstrate the same conclusion when $\lambda_{var} \geq 0.5$ and $\lambda_{div} \geq 1.0$.

\noindent \textbf{Effect of Number of View Embeddings $k$ in SDM.} 
In Figure ~\ref{fig:loss_influence}(d), we report the H when varying $k$ from 1 to 8.
Besides, we compare with $k = 0$, the baseline without SDM, \textit{i.e.}, leveraging global feature as the view embedding.
As the $k$ increases, we see progressive performance improvement across both datasets. 
This is due to multi-view embeddings capturing distinct semantics. 
However, a high $k$ may be biased to seen classes, thus harming the performance.

\subsection{Computation Cost Analysis}
In Table \ref{tab:compute_cost}, we compare the trainable parameters, the time for training one epoch and inference single image, and the performance with previous methods.
With comparable computation cost, our method outperforms previous methods ~\citep{I2DFormer, I2MVFormer}.
It verifies that the performance improvement is due to more accurate semantic alignment instead of increased parameters.
Moreover, after adding the SDM, the performance improves significantly with a slight increase in model parameters and training time.
This indicates the importance of decomposing semantics for modeling the partial association. 
{
    \setlength{\tabcolsep}{3.4pt}
    \renewcommand{\arraystretch}{1.02} 
    
    \begin{table}[t]
    \setlength{\aboverulesep}{0pt}\setlength{\belowrulesep}{0pt}
    \centering
    \caption{\textbf{Computation cost analysis on the FLO dataset.}}
    \vspace{-10pt}
    \label{tab:compute_cost}
    \resizebox{0.75\linewidth}{!}
        {
        \begin{tabular}{lcccc}
        \toprule
        \multirow{2}{*}{\textbf{Model}} &  \textbf{Params} &\textbf{Train} & \textbf{Inference}  &   \textbf{FLO} \\
          & ($\times 10^6$) & (min) & (ms) & (H) \\
        \midrule
        \texttt{I2DFormer}~\citep{I2DFormer}  & 2.18 & 0.72 & 4.7 &  53.8 \\
        \texttt{I2MVFormer}~\citep{I2MVFormer}   & \textbf{3.86} & 0.80 & \textbf{5.3} & 57.1 \\
        \midrule
        \textbf{\ModelName} \texttt{w/o \nameViewModuleName} & 1.52 & 0.67 &  4.6 &  57.9  \\
        \textbf{\ModelName}  & 3.10 & \textbf{0.98} & 5.2 & \textbf{67.3} \\
        \bottomrule
        \end{tabular}}
        \vspace{-12pt}
    \end{table}
}

\section{Conclusion}
Our EmDepart models the partial association between documents and corresponding images, accurately aligning visual and textual space based on the matching information. 
By introducing local-to-semantic variance loss and multiple semantic diversity loss, SDM generates multi-view semantic embeddings. 
These losses also help the previous methods solve the feature collapse problem.
Moreover, we introduce two losses to partially align the semantic concepts between documents and images at the view and word-to-patch levels. 
In addition, we propose a partial score to filter out unmatched information and evaluate semantic similarity accurately. 
With comparable training parameters, EmDepart outperforms SOTA methods on three benchmarks for document-based ZSL.
Qualitatively, our model learns the interpretable partial semantic association.



\begin{acks}
This work was supported by the Central Guidance for Local Special Project (Grant No. Z231100005923044) and the Climbing Plan Project (Grant No. E3Z0261).
\end{acks}

\bibliographystyle{ACM-Reference-Format}
\bibliography{references}


\begin{thebibliography}{79}


\ifx \showCODEN    \undefined \def \showCODEN     #1{\unskip}     \fi
\ifx \showDOI      \undefined \def \showDOI       #1{#1}\fi
\ifx \showISBNx    \undefined \def \showISBNx     #1{\unskip}     \fi
\ifx \showISBNxiii \undefined \def \showISBNxiii  #1{\unskip}     \fi
\ifx \showISSN     \undefined \def \showISSN      #1{\unskip}     \fi
\ifx \showLCCN     \undefined \def \showLCCN      #1{\unskip}     \fi
\ifx \shownote     \undefined \def \shownote      #1{#1}          \fi
\ifx \showarticletitle \undefined \def \showarticletitle #1{#1}   \fi
\ifx \showURL      \undefined \def \showURL       {\relax}        \fi
\providecommand\bibfield[2]{#2}
\providecommand\bibinfo[2]{#2}
\providecommand\natexlab[1]{#1}
\providecommand\showeprint[2][]{arXiv:#2}

\bibitem[Akata et~al\mbox{.}(2013)]%
        {CVPR_2013_Zeynep}
\bibfield{author}{\bibinfo{person}{Zeynep Akata}, \bibinfo{person}{Florent Perronnin}, \bibinfo{person}{Za{\"{\i}}d Harchaoui}, {and} \bibinfo{person}{Cordelia Schmid}.} \bibinfo{year}{2013}\natexlab{}.
\newblock \showarticletitle{Label-Embedding for Attribute-Based Classification}. In \bibinfo{booktitle}{\emph{2013 {IEEE} Conference on Computer Vision and Pattern Recognition, {CVPR} 2013.}} \bibinfo{pages}{819--826}.
\newblock


\bibitem[Akata et~al\mbox{.}(2016)]%
        {TPAMI_2015_Zeynep_Akata}
\bibfield{author}{\bibinfo{person}{Zeynep Akata}, \bibinfo{person}{Florent Perronnin}, \bibinfo{person}{Za{\"{\i}}d Harchaoui}, {and} \bibinfo{person}{Cordelia Schmid}.} \bibinfo{year}{2016}\natexlab{}.
\newblock \showarticletitle{Label-Embedding for Image Classification}.
\newblock \bibinfo{journal}{\emph{{IEEE} Trans. Pattern Anal. Mach. Intell.}} \bibinfo{volume}{38}, \bibinfo{number}{7} (\bibinfo{year}{2016}), \bibinfo{pages}{1425--1438}.
\newblock


\bibitem[Akata et~al\mbox{.}(2015)]%
        {word_embed_CVPR_2015_Zeynep_Akata}
\bibfield{author}{\bibinfo{person}{Zeynep Akata}, \bibinfo{person}{Scott~E. Reed}, \bibinfo{person}{Daniel Walter}, \bibinfo{person}{Honglak Lee}, {and} \bibinfo{person}{Bernt Schiele}.} \bibinfo{year}{2015}\natexlab{}.
\newblock \showarticletitle{Evaluation of output embeddings for fine-grained image classification}. In \bibinfo{booktitle}{\emph{2015 {IEEE} Conference on Computer Vision and Pattern Recognition, {CVPR} 2015.}} \bibinfo{pages}{2927--2936}.
\newblock


\bibitem[Al{-}Halah and Stiefelhagen(2017)]%
        {Document-based_CVPR_2017_Ziad}
\bibfield{author}{\bibinfo{person}{Ziad Al{-}Halah} {and} \bibinfo{person}{Rainer Stiefelhagen}.} \bibinfo{year}{2017}\natexlab{}.
\newblock \showarticletitle{Automatic Discovery, Association Estimation and Learning of Semantic Attributes for a Thousand Categories}. In \bibinfo{booktitle}{\emph{2017 {IEEE} Conference on Computer Vision and Pattern Recognition, {CVPR} 2017.}} \bibinfo{pages}{5112--5121}.
\newblock


\bibitem[andEhsan Elhamifar(2020)]%
        {DAZLE}
\bibfield{author}{\bibinfo{person}{Dat~Huynh andEhsan Elhamifar}.} \bibinfo{year}{2020}\natexlab{}.
\newblock \showarticletitle{Fine-Grained Generalized Zero-Shot Learning via Dense Attribute-Based Attention}. In \bibinfo{booktitle}{\emph{2020 {IEEE} Conference on Computer Vision and Pattern Recognition, {CVPR} 2020.}} \bibinfo{pages}{4482--4492}.
\newblock


\bibitem[Ba et~al\mbox{.}(2015)]%
        {Document-based_ICCV_2015_Lei_Jimmy_Ba}
\bibfield{author}{\bibinfo{person}{Lei~Jimmy Ba}, \bibinfo{person}{Kevin Swersky}, \bibinfo{person}{Sanja Fidler}, {and} \bibinfo{person}{Ruslan Salakhutdinov}.} \bibinfo{year}{2015}\natexlab{}.
\newblock \showarticletitle{Predicting Deep Zero-Shot Convolutional Neural Networks Using Textual Descriptions}. In \bibinfo{booktitle}{\emph{2015 {IEEE/CVF} International Conference on Computer Vision, {ICCV} 2015.}} \bibinfo{pages}{4247--4255}.
\newblock


\bibitem[Beltagy et~al\mbox{.}(2020)]%
        {LongFormer}
\bibfield{author}{\bibinfo{person}{Iz Beltagy}, \bibinfo{person}{Matthew~E. Peters}, {and} \bibinfo{person}{Arman Cohan}.} \bibinfo{year}{2020}\natexlab{}.
\newblock \showarticletitle{Longformer: The Long-Document Transformer}.
\newblock \bibinfo{journal}{\emph{CoRR}}  \bibinfo{volume}{abs/2004.05150} (\bibinfo{year}{2020}).
\newblock
\showeprint[arXiv]{2004.05150}


\bibitem[Brown et~al\mbox{.}(2020)]%
        {prompt_1}
\bibfield{author}{\bibinfo{person}{Tom~B. Brown}, \bibinfo{person}{Benjamin Mann}, \bibinfo{person}{Nick Ryder}, \bibinfo{person}{Melanie Subbiah}, \bibinfo{person}{Jared Kaplan}, \bibinfo{person}{Prafulla Dhariwal}, \bibinfo{person}{Arvind Neelakantan}, \bibinfo{person}{Pranav Shyam}, \bibinfo{person}{Girish Sastry}, \bibinfo{person}{Amanda Askell}, \bibinfo{person}{Sandhini Agarwal}, \bibinfo{person}{Ariel Herbert{-}Voss}, \bibinfo{person}{Gretchen Krueger}, \bibinfo{person}{Tom Henighan}, \bibinfo{person}{Rewon Child}, \bibinfo{person}{Aditya Ramesh}, \bibinfo{person}{Daniel~M. Ziegler}, \bibinfo{person}{Jeffrey Wu}, \bibinfo{person}{Clemens Winter}, \bibinfo{person}{Christopher Hesse}, \bibinfo{person}{Mark Chen}, \bibinfo{person}{Eric Sigler}, \bibinfo{person}{Mateusz Litwin}, \bibinfo{person}{Scott Gray}, \bibinfo{person}{Benjamin Chess}, \bibinfo{person}{Jack Clark}, \bibinfo{person}{Christopher Berner}, \bibinfo{person}{Sam McCandlish}, \bibinfo{person}{Alec Radford}, \bibinfo{person}{Ilya Sutskever},
  {and} \bibinfo{person}{Dario Amodei}.} \bibinfo{year}{2020}\natexlab{}.
\newblock \showarticletitle{Language Models are Few-Shot Learners}. In \bibinfo{booktitle}{\emph{Advances in Neural Information Processing Systems 33: Annual Conference on Neural Information Processing Systems 2020, NeurIPS 2020.}}
\newblock


\bibitem[Bujwid and Sullivan(2021)]%
        {Attribute_drawback_and_document_arxiv_2021}
\bibfield{author}{\bibinfo{person}{Sebastian Bujwid} {and} \bibinfo{person}{Josephine Sullivan}.} \bibinfo{year}{2021}\natexlab{}.
\newblock \showarticletitle{Large-Scale Zero-Shot Image Classification from Rich and Diverse Textual Descriptions}.
\newblock \bibinfo{journal}{\emph{CoRR}}  \bibinfo{volume}{abs/2103.09669} (\bibinfo{year}{2021}).
\newblock
\showeprint[arXiv]{2103.09669}


\bibitem[Changpinyo et~al\mbox{.}(2016)]%
        {CVPR_2016_Soravit_Changpinyo}
\bibfield{author}{\bibinfo{person}{Soravit Changpinyo}, \bibinfo{person}{Wei{-}Lun Chao}, \bibinfo{person}{Boqing Gong}, {and} \bibinfo{person}{Fei Sha}.} \bibinfo{year}{2016}\natexlab{}.
\newblock \showarticletitle{Synthesized Classifiers for Zero-Shot Learning}. In \bibinfo{booktitle}{\emph{2016 {IEEE} Conference on Computer Vision and Pattern Recognition, {CVPR} 2016.}} \bibinfo{pages}{5327--5336}.
\newblock


\bibitem[Chao et~al\mbox{.}(2016)]%
        {calibrated_stack}
\bibfield{author}{\bibinfo{person}{Wei{-}Lun Chao}, \bibinfo{person}{Soravit Changpinyo}, \bibinfo{person}{Boqing Gong}, {and} \bibinfo{person}{Fei Sha}.} \bibinfo{year}{2016}\natexlab{}.
\newblock \showarticletitle{An Empirical Study and Analysis of Generalized Zero-Shot Learning for Object Recognition in the Wild}. In \bibinfo{booktitle}{\emph{Computer Vision - {ECCV} 2016.}}, Vol.~\bibinfo{volume}{9906}. \bibinfo{pages}{52--68}.
\newblock


\bibitem[Chen et~al\mbox{.}(2022a)]%
        {TransZero}
\bibfield{author}{\bibinfo{person}{Shiming Chen}, \bibinfo{person}{Ziming Hong}, \bibinfo{person}{Yang Liu}, \bibinfo{person}{Guo{-}Sen Xie}, \bibinfo{person}{Baigui Sun}, \bibinfo{person}{Hao Li}, \bibinfo{person}{Qinmu Peng}, \bibinfo{person}{Ke Lu}, {and} \bibinfo{person}{Xinge You}.} \bibinfo{year}{2022}\natexlab{a}.
\newblock \showarticletitle{TransZero: Attribute-Guided Transformer for Zero-Shot Learning}. In \bibinfo{booktitle}{\emph{Thirty-Sixth {AAAI} Conference on Artificial Intelligence, {AAAI} 2022}}. \bibinfo{pages}{330--338}.
\newblock


\bibitem[Chen et~al\mbox{.}(2024)]%
        {RGAT}
\bibfield{author}{\bibinfo{person}{Shiming Chen}, \bibinfo{person}{Ziming Hong}, \bibinfo{person}{Guo{-}Sen Xie}, \bibinfo{person}{Qinmu Peng}, \bibinfo{person}{Xinge You}, \bibinfo{person}{Weiping Ding}, {and} \bibinfo{person}{Ling Shao}.} \bibinfo{year}{2024}\natexlab{}.
\newblock \showarticletitle{{GNDAN:} Graph Navigated Dual Attention Network for Zero-Shot Learning}.
\newblock \bibinfo{journal}{\emph{{IEEE} Trans. Neural Networks Learn. Syst.}} \bibinfo{volume}{35}, \bibinfo{number}{4} (\bibinfo{year}{2024}), \bibinfo{pages}{4516--4529}.
\newblock


\bibitem[Chen et~al\mbox{.}(2022b)]%
        {CVPR_2022_MSDN}
\bibfield{author}{\bibinfo{person}{Shiming Chen}, \bibinfo{person}{Ziming Hong}, \bibinfo{person}{Guo{-}Sen Xie}, \bibinfo{person}{Wenhan Yang}, \bibinfo{person}{Qinmu Peng}, \bibinfo{person}{Kai Wang}, \bibinfo{person}{Jian Zhao}, {and} \bibinfo{person}{Xinge You}.} \bibinfo{year}{2022}\natexlab{b}.
\newblock \showarticletitle{{MSDN:} Mutually Semantic Distillation Network for Zero-Shot Learning}. In \bibinfo{booktitle}{\emph{2022 {IEEE} Conference on Computer Vision and Pattern Recognition, {CVPR} 2022.}} \bibinfo{pages}{7602--7611}.
\newblock


\bibitem[Chen et~al\mbox{.}(2023a)]%
        {AAAI_2023_Duet}
\bibfield{author}{\bibinfo{person}{Zhuo Chen}, \bibinfo{person}{Yufeng Huang}, \bibinfo{person}{Jiaoyan Chen}, \bibinfo{person}{Yuxia Geng}, \bibinfo{person}{Wen Zhang}, \bibinfo{person}{Yin Fang}, \bibinfo{person}{Jeff~Z. Pan}, {and} \bibinfo{person}{Huajun Chen}.} \bibinfo{year}{2023}\natexlab{a}.
\newblock \showarticletitle{{DUET:} Cross-Modal Semantic Grounding for Contrastive Zero-Shot Learning}. In \bibinfo{booktitle}{\emph{Thirty-Seventh {AAAI} Conference on Artificial Intelligence, {AAAI} 2023, Thirty-Fifth Conference on Innovative Applications of Artificial Intelligence, {IAAI} 2023.}} \bibinfo{pages}{405--413}.
\newblock


\bibitem[Chen et~al\mbox{.}(2023b)]%
        {MM_23_3}
\bibfield{author}{\bibinfo{person}{Zhi Chen}, \bibinfo{person}{Peng{-}Fei Zhang}, \bibinfo{person}{Jingjing Li}, \bibinfo{person}{Sen Wang}, {and} \bibinfo{person}{Zi Huang}.} \bibinfo{year}{2023}\natexlab{b}.
\newblock \showarticletitle{Zero-Shot Learning by Harnessing Adversarial Samples}. In \bibinfo{booktitle}{\emph{Proceedings of the 31st {ACM} International Conference on Multimedia, {MM} 2023.}} \bibinfo{pages}{4138--4146}.
\newblock


\bibitem[Chowdhery et~al\mbox{.}(2022)]%
        {PaLM}
\bibfield{author}{\bibinfo{person}{Aakanksha Chowdhery}, \bibinfo{person}{Sharan Narang}, \bibinfo{person}{Jacob Devlin}, \bibinfo{person}{Maarten Bosma}, \bibinfo{person}{Gaurav Mishra}, \bibinfo{person}{Adam Roberts}, \bibinfo{person}{Paul Barham}, \bibinfo{person}{Hyung~Won Chung}, \bibinfo{person}{Charles Sutton}, \bibinfo{person}{Sebastian Gehrmann}, \bibinfo{person}{Parker Schuh}, \bibinfo{person}{Kensen Shi}, \bibinfo{person}{Sasha Tsvyashchenko}, \bibinfo{person}{Joshua Maynez}, \bibinfo{person}{Abhishek Rao}, \bibinfo{person}{Parker Barnes}, \bibinfo{person}{Yi Tay}, \bibinfo{person}{Noam Shazeer}, \bibinfo{person}{Vinodkumar Prabhakaran}, \bibinfo{person}{Emily Reif}, \bibinfo{person}{Nan Du}, \bibinfo{person}{Ben Hutchinson}, \bibinfo{person}{Reiner Pope}, \bibinfo{person}{James Bradbury}, \bibinfo{person}{Jacob Austin}, \bibinfo{person}{Michael Isard}, \bibinfo{person}{Guy Gur{-}Ari}, \bibinfo{person}{Pengcheng Yin}, \bibinfo{person}{Toju Duke}, \bibinfo{person}{Anselm Levskaya},
  \bibinfo{person}{Sanjay Ghemawat}, \bibinfo{person}{Sunipa Dev}, \bibinfo{person}{Henryk Michalewski}, \bibinfo{person}{Xavier Garcia}, \bibinfo{person}{Vedant Misra}, \bibinfo{person}{Kevin Robinson}, \bibinfo{person}{Liam Fedus}, \bibinfo{person}{Denny Zhou}, \bibinfo{person}{Daphne Ippolito}, \bibinfo{person}{David Luan}, \bibinfo{person}{Hyeontaek Lim}, \bibinfo{person}{Barret Zoph}, \bibinfo{person}{Alexander Spiridonov}, \bibinfo{person}{Ryan Sepassi}, \bibinfo{person}{David Dohan}, \bibinfo{person}{Shivani Agrawal}, \bibinfo{person}{Mark Omernick}, \bibinfo{person}{Andrew~M. Dai}, \bibinfo{person}{Thanumalayan~Sankaranarayana Pillai}, \bibinfo{person}{Marie Pellat}, \bibinfo{person}{Aitor Lewkowycz}, \bibinfo{person}{Erica Moreira}, \bibinfo{person}{Rewon Child}, \bibinfo{person}{Oleksandr Polozov}, \bibinfo{person}{Katherine Lee}, \bibinfo{person}{Zongwei Zhou}, \bibinfo{person}{Xuezhi Wang}, \bibinfo{person}{Brennan Saeta}, \bibinfo{person}{Mark Diaz}, \bibinfo{person}{Orhan Firat},
  \bibinfo{person}{Michele Catasta}, \bibinfo{person}{Jason Wei}, \bibinfo{person}{Kathy Meier{-}Hellstern}, \bibinfo{person}{Douglas Eck}, \bibinfo{person}{Jeff Dean}, \bibinfo{person}{Slav Petrov}, {and} \bibinfo{person}{Noah Fiedel}.} \bibinfo{year}{2022}\natexlab{}.
\newblock \showarticletitle{PaLM: Scaling Language Modeling with Pathways}.
\newblock \bibinfo{journal}{\emph{CoRR}}  \bibinfo{volume}{abs/2204.02311} (\bibinfo{year}{2022}).
\newblock
\showeprint[arXiv]{2204.02311}


\bibitem[Chun et~al\mbox{.}(2021)]%
        {set_based_embed_CVPR_2021}
\bibfield{author}{\bibinfo{person}{Sanghyuk Chun}, \bibinfo{person}{Seong~Joon Oh}, \bibinfo{person}{Rafael~Sampaio de Rezende}, \bibinfo{person}{Yannis Kalantidis}, {and} \bibinfo{person}{Diane Larlus}.} \bibinfo{year}{2021}\natexlab{}.
\newblock \showarticletitle{Probabilistic Embeddings for Cross-Modal Retrieval}. In \bibinfo{booktitle}{\emph{2021 {IEEE} Conference on Computer Vision and Pattern Recognition, {CVPR} 2021.}} \bibinfo{pages}{8415--8424}.
\newblock


\bibitem[Daniel et~al\mbox{.}(1991)]%
        {AWA_attribute_provide1}
\bibfield{author}{\bibinfo{person}{Daniel}, \bibinfo{person}{N.}, \bibinfo{person}{Osherson}, \bibinfo{person}{Joshua}, \bibinfo{person}{Stern}, \bibinfo{person}{Ormond}, \bibinfo{person}{Wilkie}, \bibinfo{person}{Michael}, \bibinfo{person}{Stob}, {and} \bibinfo{person}{Edward}.} \bibinfo{year}{1991}\natexlab{}.
\newblock \showarticletitle{Default Probability}.
\newblock \bibinfo{journal}{\emph{Cognitive Science}} (\bibinfo{year}{1991}).
\newblock


\bibitem[Deng et~al\mbox{.}(2009)]%
        {ImageNet}
\bibfield{author}{\bibinfo{person}{Jia Deng}, \bibinfo{person}{Wei Dong}, \bibinfo{person}{Richard Socher}, \bibinfo{person}{Li{-}Jia Li}, \bibinfo{person}{Kai Li}, {and} \bibinfo{person}{Li Fei{-}Fei}.} \bibinfo{year}{2009}\natexlab{}.
\newblock \showarticletitle{ImageNet: {A} large-scale hierarchical image database}. In \bibinfo{booktitle}{\emph{2009 {IEEE} Conference on Computer Vision and Pattern Recognition, {CVPR} 2009.}} \bibinfo{pages}{248--255}.
\newblock


\bibitem[Devlin et~al\mbox{.}(2019)]%
        {BERT}
\bibfield{author}{\bibinfo{person}{Jacob Devlin}, \bibinfo{person}{Ming{-}Wei Chang}, \bibinfo{person}{Kenton Lee}, {and} \bibinfo{person}{Kristina Toutanova}.} \bibinfo{year}{2019}\natexlab{}.
\newblock \showarticletitle{{BERT:} Pre-training of Deep Bidirectional Transformers for Language Understanding}. In \bibinfo{booktitle}{\emph{Proceedings of the 2019 Conference of the North American Chapter of the Association for Computational Linguistics: Human Language Technologies, {NAACL-HLT} 2019.}} \bibinfo{pages}{4171--4186}.
\newblock


\bibitem[Dosovitskiy et~al\mbox{.}(2021)]%
        {vit}
\bibfield{author}{\bibinfo{person}{Alexey Dosovitskiy}, \bibinfo{person}{Lucas Beyer}, \bibinfo{person}{Alexander Kolesnikov}, \bibinfo{person}{Dirk Weissenborn}, \bibinfo{person}{Xiaohua Zhai}, \bibinfo{person}{Thomas Unterthiner}, \bibinfo{person}{Mostafa Dehghani}, \bibinfo{person}{Matthias Minderer}, \bibinfo{person}{Georg Heigold}, \bibinfo{person}{Sylvain Gelly}, \bibinfo{person}{Jakob Uszkoreit}, {and} \bibinfo{person}{Neil Houlsby}.} \bibinfo{year}{2021}\natexlab{}.
\newblock \showarticletitle{An Image is Worth 16x16 Words: Transformers for Image Recognition at Scale}. In \bibinfo{booktitle}{\emph{9th International Conference on Learning Representations, {ICLR} 2021.}}
\newblock


\bibitem[Elhoseiny et~al\mbox{.}(2017)]%
        {Document-based_CVPR_2017_Mohamed}
\bibfield{author}{\bibinfo{person}{Mohamed Elhoseiny}, \bibinfo{person}{Yizhe Zhu}, \bibinfo{person}{Han Zhang}, {and} \bibinfo{person}{Ahmed~M. Elgammal}.} \bibinfo{year}{2017}\natexlab{}.
\newblock \showarticletitle{Link the Head to the "Beak": Zero Shot Learning from Noisy Text Description at Part Precision}. In \bibinfo{booktitle}{\emph{2017 {IEEE} Conference on Computer Vision and Pattern Recognition, {CVPR} 2017.}} \bibinfo{pages}{6288--6297}.
\newblock


\bibitem[Eloundou et~al\mbox{.}(2023)]%
        {ChatGPT}
\bibfield{author}{\bibinfo{person}{Tyna Eloundou}, \bibinfo{person}{Sam Manning}, \bibinfo{person}{Pamela Mishkin}, {and} \bibinfo{person}{Daniel Rock}.} \bibinfo{year}{2023}\natexlab{}.
\newblock \showarticletitle{GPTs are GPTs: An Early Look at the Labor Market Impact Potential of Large Language Models}.
\newblock \bibinfo{journal}{\emph{CoRR}}  \bibinfo{volume}{abs/2303.10130} (\bibinfo{year}{2023}).
\newblock
\showeprint[arXiv]{2303.10130}


\bibitem[Frome et~al\mbox{.}(2013)]%
        {word_embed_NIPS_2013_Devise}
\bibfield{author}{\bibinfo{person}{Andrea Frome}, \bibinfo{person}{Gregory~S. Corrado}, \bibinfo{person}{Jonathon Shlens}, \bibinfo{person}{Samy Bengio}, \bibinfo{person}{Jeffrey Dean}, \bibinfo{person}{Marc'Aurelio Ranzato}, {and} \bibinfo{person}{Tom{\'{a}}s Mikolov}.} \bibinfo{year}{2013}\natexlab{}.
\newblock \showarticletitle{DeViSE: {A} Deep Visual-Semantic Embedding Model}. In \bibinfo{booktitle}{\emph{Advances in Neural Information Processing Systems 26: Annual Conference on Neural Information Processing Systems 2013, NeurIPS 2013.}} \bibinfo{pages}{2121--2129}.
\newblock


\bibitem[Ge et~al\mbox{.}(2022)]%
        {MM_22_1}
\bibfield{author}{\bibinfo{person}{Jiannan Ge}, \bibinfo{person}{Hongtao Xie}, \bibinfo{person}{Shaobo Min}, \bibinfo{person}{Pandeng Li}, {and} \bibinfo{person}{Yongdong Zhang}.} \bibinfo{year}{2022}\natexlab{}.
\newblock \showarticletitle{Dual Part Discovery Network for Zero-Shot Learning}. In \bibinfo{booktitle}{\emph{Proceedings of the 30st {ACM} International Conference on Multimedia, {MM} 2022.}} \bibinfo{pages}{3244--3252}.
\newblock


\bibitem[Hendrycks and Gimpel(2016)]%
        {GELU}
\bibfield{author}{\bibinfo{person}{Dan Hendrycks} {and} \bibinfo{person}{Kevin Gimpel}.} \bibinfo{year}{2016}\natexlab{}.
\newblock \showarticletitle{Bridging Nonlinearities and Stochastic Regularizers with Gaussian Error Linear Units}.
\newblock \bibinfo{journal}{\emph{CoRR}}  \bibinfo{volume}{abs/1606.08415} (\bibinfo{year}{2016}).
\newblock
\showeprint[arXiv]{1606.08415}


\bibitem[Huang et~al\mbox{.}(2012)]%
        {pre_trained_LM_ACL_2012}
\bibfield{author}{\bibinfo{person}{Eric~H. Huang}, \bibinfo{person}{Richard Socher}, \bibinfo{person}{Christopher~D. Manning}, {and} \bibinfo{person}{Andrew~Y. Ng}.} \bibinfo{year}{2012}\natexlab{}.
\newblock \showarticletitle{Improving Word Representations via Global Context and Multiple Word Prototypes}. In \bibinfo{booktitle}{\emph{The 50th Annual Meeting of the Association for Computational Linguistics, Proceedings of the Conference.}} \bibinfo{pages}{873--882}.
\newblock


\bibitem[Jurie et~al\mbox{.}(2017)]%
        {word_embed_and_graph_ICCV_2017}
\bibfield{author}{\bibinfo{person}{Fr{\'{e}}d{\'{e}}ric Jurie}, \bibinfo{person}{Maxime Bucher}, {and} \bibinfo{person}{St{\'{e}}phane Herbin}.} \bibinfo{year}{2017}\natexlab{}.
\newblock \showarticletitle{Generating Visual Representations for Zero-Shot Classification}. In \bibinfo{booktitle}{\emph{2017 {IEEE/CVF} International Conference on Computer Vision, {ICCV} 2017 - Workshops.}} \bibinfo{pages}{2666--2673}.
\newblock


\bibitem[Kampffmeyer et~al\mbox{.}(2019)]%
        {word_embed_and_graph_CVPR_2019_Michael_Kampffmeyer}
\bibfield{author}{\bibinfo{person}{Michael Kampffmeyer}, \bibinfo{person}{Yinbo Chen}, \bibinfo{person}{Xiaodan Liang}, \bibinfo{person}{Hao Wang}, \bibinfo{person}{Yujia Zhang}, {and} \bibinfo{person}{Eric~P. Xing}.} \bibinfo{year}{2019}\natexlab{}.
\newblock \showarticletitle{Rethinking Knowledge Graph Propagation for Zero-Shot Learning}. In \bibinfo{booktitle}{\emph{2019 {IEEE} Conference on Computer Vision and Pattern Recognition, {CVPR} 2019.}} \bibinfo{pages}{11487--11496}.
\newblock


\bibitem[Kemp et~al\mbox{.}(2006)]%
        {AWA_attribute_provide2}
\bibfield{author}{\bibinfo{person}{Charles Kemp}, \bibinfo{person}{Joshua~B. Tenenbaum}, \bibinfo{person}{Thomas~L. Griffiths}, \bibinfo{person}{Takeshi Yamada}, {and} \bibinfo{person}{Naonori Ueda}.} \bibinfo{year}{2006}\natexlab{}.
\newblock \showarticletitle{Learning Systems of Concepts with an Infinite Relational Model}. In \bibinfo{booktitle}{\emph{Proceedings, The Twenty-First National Conference on Artificial Intelligence and the Eighteenth Innovative Applications of Artificial Intelligence Conference.}} \bibinfo{pages}{381--388}.
\newblock


\bibitem[Kil and Chao(2021)]%
        {Document-based_NAACL_2021_Jihyung_Kil}
\bibfield{author}{\bibinfo{person}{Jihyung Kil} {and} \bibinfo{person}{Wei{-}Lun Chao}.} \bibinfo{year}{2021}\natexlab{}.
\newblock \showarticletitle{Revisiting Document Representations for Large-Scale Zero-Shot Learning}. In \bibinfo{booktitle}{\emph{Proceedings of the 2021 Conference of the North American Chapter of the Association for Computational Linguistics: Human Language Technologies, {NAACL-HLT} 2021.}} \bibinfo{pages}{3117--3128}.
\newblock


\bibitem[Kim et~al\mbox{.}(2023)]%
        {set_based_embed_CVPR_2023}
\bibfield{author}{\bibinfo{person}{Dongwon Kim}, \bibinfo{person}{Namyup Kim}, {and} \bibinfo{person}{Suha Kwak}.} \bibinfo{year}{2023}\natexlab{}.
\newblock \showarticletitle{Improving Cross-Modal Retrieval with Set of Diverse Embeddings}. In \bibinfo{booktitle}{\emph{2023 {IEEE} Conference on Computer Vision and Pattern Recognition, {CVPR} 2023.}} \bibinfo{pages}{23422--23431}.
\newblock


\bibitem[Kong et~al\mbox{.}(2022)]%
        {CVPR_2022_En-Compactness}
\bibfield{author}{\bibinfo{person}{Xia Kong}, \bibinfo{person}{Zuodong Gao}, \bibinfo{person}{Xiaofan Li}, \bibinfo{person}{Ming Hong}, \bibinfo{person}{Jun Liu}, \bibinfo{person}{Chengjie Wang}, \bibinfo{person}{Yuan Xie}, {and} \bibinfo{person}{Yanyun Qu}.} \bibinfo{year}{2022}\natexlab{}.
\newblock \showarticletitle{En-Compactness: Self-Distillation Embedding {\&} Contrastive Generation for Generalized Zero-Shot Learning}. In \bibinfo{booktitle}{\emph{2022 {IEEE} Conference on Computer Vision and Pattern Recognition, {CVPR} 2022.}} \bibinfo{pages}{9296--9305}.
\newblock


\bibitem[Lampert et~al\mbox{.}(2009)]%
        {ZSL_invent1}
\bibfield{author}{\bibinfo{person}{Christoph~H. Lampert}, \bibinfo{person}{Hannes Nickisch}, {and} \bibinfo{person}{Stefan Harmeling}.} \bibinfo{year}{2009}\natexlab{}.
\newblock \showarticletitle{Learning to detect unseen object classes by between-class attribute transfer}. In \bibinfo{booktitle}{\emph{2009 {IEEE} Conference on Computer Vision and Pattern Recognition, {CVPR} 2009.}} \bibinfo{pages}{951--958}.
\newblock


\bibitem[Lampert et~al\mbox{.}(2014)]%
        {AWA1_and_attribute_provide}
\bibfield{author}{\bibinfo{person}{Christoph~H. Lampert}, \bibinfo{person}{Hannes Nickisch}, {and} \bibinfo{person}{Stefan Harmeling}.} \bibinfo{year}{2014}\natexlab{}.
\newblock \showarticletitle{Attribute-Based Classification for Zero-Shot Visual Object Categorization}.
\newblock \bibinfo{journal}{\emph{{IEEE} Trans. Pattern Anal. Mach. Intell.}} \bibinfo{volume}{36}, \bibinfo{number}{3} (\bibinfo{year}{2014}), \bibinfo{pages}{453--465}.
\newblock


\bibitem[Lin et~al\mbox{.}(2022)]%
        {set_based_embed_NIPS_2022}
\bibfield{author}{\bibinfo{person}{Chengzhi Lin}, \bibinfo{person}{Ancong Wu}, \bibinfo{person}{Junwei Liang}, \bibinfo{person}{Jun Zhang}, \bibinfo{person}{Wenhang Ge}, \bibinfo{person}{Wei{-}Shi Zheng}, {and} \bibinfo{person}{Chunhua Shen}.} \bibinfo{year}{2022}\natexlab{}.
\newblock \showarticletitle{Text-Adaptive Multiple Visual Prototype Matching for Video-Text Retrieval}. In \bibinfo{booktitle}{\emph{Advances in Neural Information Processing Systems 35: Annual Conference on Neural Information Processing Systems 2022, NeurIPS 2022.}}
\newblock


\bibitem[Liu et~al\mbox{.}(2023)]%
        {CVPR_2023_PSVMA}
\bibfield{author}{\bibinfo{person}{Man Liu}, \bibinfo{person}{Feng Li}, \bibinfo{person}{Chunjie Zhang}, \bibinfo{person}{Yunchao Wei}, \bibinfo{person}{Huihui Bai}, {and} \bibinfo{person}{Yao Zhao}.} \bibinfo{year}{2023}\natexlab{}.
\newblock \showarticletitle{Progressive Semantic-Visual Mutual Adaption for Generalized Zero-Shot Learning}. In \bibinfo{booktitle}{\emph{2023 {IEEE} Conference on Computer Vision and Pattern Recognition, {CVPR} 2023.}} \bibinfo{pages}{15337--15346}.
\newblock


\bibitem[Locatello et~al\mbox{.}(2020)]%
        {slot_attention}
\bibfield{author}{\bibinfo{person}{Francesco Locatello}, \bibinfo{person}{Dirk Weissenborn}, \bibinfo{person}{Thomas Unterthiner}, \bibinfo{person}{Aravindh Mahendran}, \bibinfo{person}{Georg Heigold}, \bibinfo{person}{Jakob Uszkoreit}, \bibinfo{person}{Alexey Dosovitskiy}, {and} \bibinfo{person}{Thomas Kipf}.} \bibinfo{year}{2020}\natexlab{}.
\newblock \showarticletitle{Object-Centric Learning with Slot Attention}. In \bibinfo{booktitle}{\emph{Advances in Neural Information Processing Systems 33: Annual Conference on Neural Information Processing Systems 2020, NeurIPS 2020.}}
\newblock


\bibitem[Maniparambil et~al\mbox{.}(2023)]%
        {VDT_2023_ICCV_gpt4}
\bibfield{author}{\bibinfo{person}{Mayug Maniparambil}, \bibinfo{person}{Chris Vorster}, \bibinfo{person}{Derek Molloy}, \bibinfo{person}{Noel Murphy}, \bibinfo{person}{Kevin McGuinness}, {and} \bibinfo{person}{Noel~E. O'Connor}.} \bibinfo{year}{2023}\natexlab{}.
\newblock \showarticletitle{Enhancing {CLIP} with {GPT-4:} Harnessing Visual Descriptions as Prompts}. In \bibinfo{booktitle}{\emph{2023 {IEEE/CVF} International Conference on Computer Vision, {ICCV} 2023 - Workshops.}} \bibinfo{pages}{262--271}.
\newblock


\bibitem[Menon and Vondrick(2023)]%
        {VDT_2023_ICLR}
\bibfield{author}{\bibinfo{person}{Sachit Menon} {and} \bibinfo{person}{Carl Vondrick}.} \bibinfo{year}{2023}\natexlab{}.
\newblock \showarticletitle{Visual Classification via Description from Large Language Models}. In \bibinfo{booktitle}{\emph{11th International Conference on Learning Representations, {ICLR} 2023.}}
\newblock


\bibitem[Mikolov et~al\mbox{.}(2013)]%
        {pre_trained_LM_NIPS_2013}
\bibfield{author}{\bibinfo{person}{Tom{\'{a}}s Mikolov}, \bibinfo{person}{Ilya Sutskever}, \bibinfo{person}{Kai Chen}, \bibinfo{person}{Gregory~S. Corrado}, {and} \bibinfo{person}{Jeffrey Dean}.} \bibinfo{year}{2013}\natexlab{}.
\newblock \showarticletitle{Distributed Representations of Words and Phrases and their Compositionality}. In \bibinfo{booktitle}{\emph{Advances in Neural Information Processing Systems 26: Annual Conference on Neural Information Processing Systems 2013, NeurIPS 2013.}} \bibinfo{pages}{3111--3119}.
\newblock


\bibitem[Naeem et~al\mbox{.}(2023)]%
        {I2MVFormer}
\bibfield{author}{\bibinfo{person}{M. Naeem}, \bibinfo{person}{M.~Ali Khan}, \bibinfo{person}{Y. Xian}, \bibinfo{person}{M. Afzal}, \bibinfo{person}{D. Stricker}, \bibinfo{person}{L.~Van Gool}, {and} \bibinfo{person}{F. Tombari}.} \bibinfo{year}{2023}\natexlab{}.
\newblock \showarticletitle{I2MVFormer: Large Language Model Generated Multi-View Document Supervision for Zero-Shot Image Classification}. In \bibinfo{booktitle}{\emph{2023 {IEEE} Conference on Computer Vision and Pattern Recognition, {CVPR} 2023.}} \bibinfo{pages}{15169--15179}.
\newblock


\bibitem[Naeem et~al\mbox{.}(2022)]%
        {I2DFormer}
\bibfield{author}{\bibinfo{person}{Muhammad~Ferjad Naeem}, \bibinfo{person}{Yongqin Xian}, \bibinfo{person}{Luc~Van Gool}, {and} \bibinfo{person}{Federico Tombari}.} \bibinfo{year}{2022}\natexlab{}.
\newblock \showarticletitle{I2DFormer: Learning Image to Document Attention for Zero-Shot Image Classification}. In \bibinfo{booktitle}{\emph{Advances in Neural Information Processing Systems 35: Annual Conference on Neural Information Processing Systems 2022, NeurIPS 2022.}}
\newblock


\bibitem[Naeem et~al\mbox{.}(2021)]%
        {word_embed_and_graph_CVPR_2021_Muhammad_Ferjad_Naeem}
\bibfield{author}{\bibinfo{person}{Muhammad~Ferjad Naeem}, \bibinfo{person}{Yongqin Xian}, \bibinfo{person}{Federico Tombari}, {and} \bibinfo{person}{Zeynep Akata}.} \bibinfo{year}{2021}\natexlab{}.
\newblock \showarticletitle{Learning Graph Embeddings for Compositional Zero-Shot Learning}. In \bibinfo{booktitle}{\emph{2021 {IEEE} Conference on Computer Vision and Pattern Recognition, {CVPR} 2021.}} \bibinfo{pages}{953--962}.
\newblock


\bibitem[Nilsback and Zisserman(2008)]%
        {FLO}
\bibfield{author}{\bibinfo{person}{Maria{-}Elena Nilsback} {and} \bibinfo{person}{Andrew Zisserman}.} \bibinfo{year}{2008}\natexlab{}.
\newblock \showarticletitle{Automated Flower Classification over a Large Number of Classes}. In \bibinfo{booktitle}{\emph{Sixth Indian Conference on Computer Vision, Graphics {\&} Image Processing, {ICVGIP} 2008.}} \bibinfo{pages}{722--729}.
\newblock


\bibitem[Palatucci et~al\mbox{.}(2009)]%
        {ZSL_invent2}
\bibfield{author}{\bibinfo{person}{Mark Palatucci}, \bibinfo{person}{Dean Pomerleau}, \bibinfo{person}{Geoffrey~E. Hinton}, {and} \bibinfo{person}{Tom~M. Mitchell}.} \bibinfo{year}{2009}\natexlab{}.
\newblock \showarticletitle{Zero-shot Learning with Semantic Output Codes}. In \bibinfo{booktitle}{\emph{Advances in Neural Information Processing Systems 22: Annual Conference on Neural Information Processing Systems 2009, NeurIPS 2009.}} \bibinfo{pages}{1410--1418}.
\newblock


\bibitem[Pennington et~al\mbox{.}(2014)]%
        {Glove}
\bibfield{author}{\bibinfo{person}{Jeffrey Pennington}, \bibinfo{person}{Richard Socher}, {and} \bibinfo{person}{Christopher~D. Manning}.} \bibinfo{year}{2014}\natexlab{}.
\newblock \showarticletitle{Glove: Global Vectors for Word Representation}. In \bibinfo{booktitle}{\emph{Proceedings of the 2014 Conference on Empirical Methods in Natural Language Processing, {EMNLP} 2014.}} \bibinfo{pages}{1532--1543}.
\newblock


\bibitem[Pratt et~al\mbox{.}(2023)]%
        {VDT_2023_ICCV_what_does}
\bibfield{author}{\bibinfo{person}{Sarah~M. Pratt}, \bibinfo{person}{Ian Covert}, \bibinfo{person}{Rosanne Liu}, {and} \bibinfo{person}{Ali Farhadi}.} \bibinfo{year}{2023}\natexlab{}.
\newblock \showarticletitle{What does a platypus look like? Generating customized prompts for zero-shot image classification}. In \bibinfo{booktitle}{\emph{{IEEE/CVF} International Conference on Computer Vision, {ICCV} 2023.}} \bibinfo{pages}{15645--15655}.
\newblock


\bibitem[Qiao et~al\mbox{.}(2016)]%
        {Document-based_CVPR_2016_Ruizhi_Qiao}
\bibfield{author}{\bibinfo{person}{Ruizhi Qiao}, \bibinfo{person}{Lingqiao Liu}, \bibinfo{person}{Chunhua Shen}, {and} \bibinfo{person}{Anton van~den Hengel}.} \bibinfo{year}{2016}\natexlab{}.
\newblock \showarticletitle{Less is More: Zero-Shot Learning from Online Textual Documents with Noise Suppression}. In \bibinfo{booktitle}{\emph{2016 {IEEE} Conference on Computer Vision and Pattern Recognition, {CVPR} 2016.}} \bibinfo{pages}{2249--2257}.
\newblock


\bibitem[Radford et~al\mbox{.}(2021)]%
        {CLIP}
\bibfield{author}{\bibinfo{person}{Alec Radford}, \bibinfo{person}{Jong~Wook Kim}, \bibinfo{person}{Chris Hallacy}, \bibinfo{person}{Aditya Ramesh}, \bibinfo{person}{Gabriel Goh}, \bibinfo{person}{Sandhini Agarwal}, \bibinfo{person}{Girish Sastry}, \bibinfo{person}{Amanda Askell}, \bibinfo{person}{Pamela Mishkin}, \bibinfo{person}{Jack Clark}, \bibinfo{person}{Gretchen Krueger}, {and} \bibinfo{person}{Ilya Sutskever}.} \bibinfo{year}{2021}\natexlab{}.
\newblock \showarticletitle{Learning Transferable Visual Models From Natural Language Supervision}. In \bibinfo{booktitle}{\emph{Proceedings of the 38th International Conference on Machine Learning, {ICML} 2021.}}, Vol.~\bibinfo{volume}{139}. \bibinfo{pages}{8748--8763}.
\newblock


\bibitem[Ren et~al\mbox{.}(2023)]%
        {VDT_2023_NIPS_Hierarchical}
\bibfield{author}{\bibinfo{person}{Zhiyuan Ren}, \bibinfo{person}{Yiyang Su}, {and} \bibinfo{person}{Xiaoming Liu}.} \bibinfo{year}{2023}\natexlab{}.
\newblock \showarticletitle{ChatGPT-Powered Hierarchical Comparisons for Image Classification}. In \bibinfo{booktitle}{\emph{Advances in Neural Information Processing Systems 36: Annual Conference on Neural Information Processing Systems 2023, NeurIPS 2023.}}
\newblock


\bibitem[Romera{-}Paredes and Torr(2015)]%
        {ICML_2015_Bernardino}
\bibfield{author}{\bibinfo{person}{Bernardino Romera{-}Paredes} {and} \bibinfo{person}{Philip H.~S. Torr}.} \bibinfo{year}{2015}\natexlab{}.
\newblock \showarticletitle{An embarrassingly simple approach to zero-shot learning}. In \bibinfo{booktitle}{\emph{Proceedings of the 32nd International Conference on Machine Learning, {ICML} 2015.}}, Vol.~\bibinfo{volume}{37}. \bibinfo{pages}{2152--2161}.
\newblock


\bibitem[Roth et~al\mbox{.}(2023)]%
        {VDT_2023_ICCV_Waffle}
\bibfield{author}{\bibinfo{person}{Karsten Roth}, \bibinfo{person}{Jae{-}Myung Kim}, \bibinfo{person}{A.~Sophia Koepke}, \bibinfo{person}{Oriol Vinyals}, \bibinfo{person}{Cordelia Schmid}, {and} \bibinfo{person}{Zeynep Akata}.} \bibinfo{year}{2023}\natexlab{}.
\newblock \showarticletitle{Waffling around for Performance: Visual Classification with Random Words and Broad Concepts}. In \bibinfo{booktitle}{\emph{2023 {IEEE/CVF} International Conference on Computer Vision, {ICCV} 2023.}} \bibinfo{pages}{15700--15711}.
\newblock


\bibitem[Salton and Buckley(1988)]%
        {TF-IDF}
\bibfield{author}{\bibinfo{person}{Gerard Salton} {and} \bibinfo{person}{Chris Buckley}.} \bibinfo{year}{1988}\natexlab{}.
\newblock \showarticletitle{Term-Weighting Approaches in Automatic Text Retrieval}.
\newblock \bibinfo{journal}{\emph{Inf. Process. Manag.}} \bibinfo{volume}{24}, \bibinfo{number}{5} (\bibinfo{year}{1988}), \bibinfo{pages}{513--523}.
\newblock


\bibitem[Socher et~al\mbox{.}(2013)]%
        {word_embed_NIPS_2013_Richard_Socher}
\bibfield{author}{\bibinfo{person}{Richard Socher}, \bibinfo{person}{Milind Ganjoo}, \bibinfo{person}{Christopher~D. Manning}, {and} \bibinfo{person}{Andrew~Y. Ng}.} \bibinfo{year}{2013}\natexlab{}.
\newblock \showarticletitle{Zero-Shot Learning Through Cross-Modal Transfer}. In \bibinfo{booktitle}{\emph{Advances in Neural Information Processing Systems 26: Annual Conference on Neural Information Processing Systems 2013, NeurIPS 2013.}} \bibinfo{pages}{935--943}.
\newblock


\bibitem[Song et~al\mbox{.}(2018)]%
        {Attribute_drawback_ECCV_2018}
\bibfield{author}{\bibinfo{person}{Jie Song}, \bibinfo{person}{Chengchao Shen}, \bibinfo{person}{Jie Lei}, \bibinfo{person}{Anxiang Zeng}, \bibinfo{person}{Kairi Ou}, \bibinfo{person}{Dacheng Tao}, {and} \bibinfo{person}{Mingli Song}.} \bibinfo{year}{2018}\natexlab{}.
\newblock \showarticletitle{Selective Zero-Shot Classification with Augmented Attributes}. In \bibinfo{booktitle}{\emph{Computer Vision - {ECCV} 2018.}}, Vol.~\bibinfo{volume}{11213}. \bibinfo{pages}{474--490}.
\newblock


\bibitem[Song et~al\mbox{.}(2020)]%
        {MPNet}
\bibfield{author}{\bibinfo{person}{Kaitao Song}, \bibinfo{person}{Xu Tan}, \bibinfo{person}{Tao Qin}, \bibinfo{person}{Jianfeng Lu}, {and} \bibinfo{person}{Tie{-}Yan Liu}.} \bibinfo{year}{2020}\natexlab{}.
\newblock \showarticletitle{MPNet: Masked and Permuted Pre-training for Language Understanding}. In \bibinfo{booktitle}{\emph{Advances in Neural Information Processing Systems 33: Annual Conference on Neural Information Processing Systems 2020, NeurIPS 2020.}}
\newblock


\bibitem[Song and Soleymani(2019)]%
        {set_based_embed_CVPR_2019}
\bibfield{author}{\bibinfo{person}{Yale Song} {and} \bibinfo{person}{Mohammad Soleymani}.} \bibinfo{year}{2019}\natexlab{}.
\newblock \showarticletitle{Polysemous Visual-Semantic Embedding for Cross-Modal Retrieval}. In \bibinfo{booktitle}{\emph{2019 {IEEE} Conference on Computer Vision and Pattern Recognition, {CVPR} 2019.}} \bibinfo{pages}{1979--1988}.
\newblock


\bibitem[Su et~al\mbox{.}(2022)]%
        {CVPR_2022_Hongzu_Su}
\bibfield{author}{\bibinfo{person}{Hongzu Su}, \bibinfo{person}{Jingjing Li}, \bibinfo{person}{Zhi Chen}, \bibinfo{person}{Lei Zhu}, {and} \bibinfo{person}{Ke Lu}.} \bibinfo{year}{2022}\natexlab{}.
\newblock \showarticletitle{Distinguishing Unseen from Seen for Generalized Zero-shot Learning}. In \bibinfo{booktitle}{\emph{2022 {IEEE} Conference on Computer Vision and Pattern Recognition, {CVPR} 2022.}} \bibinfo{pages}{7875--7884}.
\newblock


\bibitem[Touvron et~al\mbox{.}(2023)]%
        {llama2}
\bibfield{author}{\bibinfo{person}{Hugo Touvron}, \bibinfo{person}{Louis Martin}, \bibinfo{person}{Kevin Stone}, \bibinfo{person}{Peter Albert}, \bibinfo{person}{Amjad Almahairi}, \bibinfo{person}{Yasmine Babaei}, \bibinfo{person}{Nikolay Bashlykov}, \bibinfo{person}{Soumya Batra}, \bibinfo{person}{Prajjwal Bhargava}, \bibinfo{person}{Shruti Bhosale}, \bibinfo{person}{Dan Bikel}, \bibinfo{person}{Lukas Blecher}, \bibinfo{person}{Cristian Canton{-}Ferrer}, \bibinfo{person}{Moya Chen}, \bibinfo{person}{Guillem Cucurull}, \bibinfo{person}{David Esiobu}, \bibinfo{person}{Jude Fernandes}, \bibinfo{person}{Jeremy Fu}, \bibinfo{person}{Wenyin Fu}, \bibinfo{person}{Brian Fuller}, \bibinfo{person}{Cynthia Gao}, \bibinfo{person}{Vedanuj Goswami}, \bibinfo{person}{Naman Goyal}, \bibinfo{person}{Anthony Hartshorn}, \bibinfo{person}{Saghar Hosseini}, \bibinfo{person}{Rui Hou}, \bibinfo{person}{Hakan Inan}, \bibinfo{person}{Marcin Kardas}, \bibinfo{person}{Viktor Kerkez}, \bibinfo{person}{Madian Khabsa},
  \bibinfo{person}{Isabel Kloumann}, \bibinfo{person}{Artem Korenev}, \bibinfo{person}{Punit~Singh Koura}, \bibinfo{person}{Marie{-}Anne Lachaux}, \bibinfo{person}{Thibaut Lavril}, \bibinfo{person}{Jenya Lee}, \bibinfo{person}{Diana Liskovich}, \bibinfo{person}{Yinghai Lu}, \bibinfo{person}{Yuning Mao}, \bibinfo{person}{Xavier Martinet}, \bibinfo{person}{Todor Mihaylov}, \bibinfo{person}{Pushkar Mishra}, \bibinfo{person}{Igor Molybog}, \bibinfo{person}{Yixin Nie}, \bibinfo{person}{Andrew Poulton}, \bibinfo{person}{Jeremy Reizenstein}, \bibinfo{person}{Rashi Rungta}, \bibinfo{person}{Kalyan Saladi}, \bibinfo{person}{Alan Schelten}, \bibinfo{person}{Ruan Silva}, \bibinfo{person}{Eric~Michael Smith}, \bibinfo{person}{Ranjan Subramanian}, \bibinfo{person}{Xiaoqing~Ellen Tan}, \bibinfo{person}{Binh Tang}, \bibinfo{person}{Ross Taylor}, \bibinfo{person}{Adina Williams}, \bibinfo{person}{Jian~Xiang Kuan}, \bibinfo{person}{Puxin Xu}, \bibinfo{person}{Zheng Yan}, \bibinfo{person}{Iliyan Zarov}, \bibinfo{person}{Yuchen
  Zhang}, \bibinfo{person}{Angela Fan}, \bibinfo{person}{Melanie Kambadur}, \bibinfo{person}{Sharan Narang}, \bibinfo{person}{Aur{\'{e}}lien Rodriguez}, \bibinfo{person}{Robert Stojnic}, \bibinfo{person}{Sergey Edunov}, {and} \bibinfo{person}{Thomas Scialom}.} \bibinfo{year}{2023}\natexlab{}.
\newblock \showarticletitle{Llama 2: Open Foundation and Fine-Tuned Chat Models}.
\newblock \bibinfo{journal}{\emph{CoRR}}  \bibinfo{volume}{abs/2307.09288} (\bibinfo{year}{2023}).
\newblock


\bibitem[Verma et~al\mbox{.}(2018)]%
        {CVPR_2018_Vinay_Kumar_Verma}
\bibfield{author}{\bibinfo{person}{Vinay~Kumar Verma}, \bibinfo{person}{Gundeep Arora}, \bibinfo{person}{Ashish Mishra}, {and} \bibinfo{person}{Piyush Rai}.} \bibinfo{year}{2018}\natexlab{}.
\newblock \showarticletitle{Generalized Zero-Shot Learning via Synthesized Examples}. In \bibinfo{booktitle}{\emph{2018 {IEEE} Conference on Computer Vision and Pattern Recognition, {CVPR} 2018.}} \bibinfo{pages}{4281--4289}.
\newblock


\bibitem[Wah et~al\mbox{.}(2011)]%
        {CUB_and_attribute_provide}
\bibfield{author}{\bibinfo{person}{Catherine Wah}, \bibinfo{person}{Steve Branson}, \bibinfo{person}{Peter Welinder}, \bibinfo{person}{Pietro Perona}, {and} \bibinfo{person}{Serge Belongie}.} \bibinfo{year}{2011}\natexlab{}.
\newblock \showarticletitle{The Caltech-UCSD Birds-200-2011 Dataset}.
\newblock \bibinfo{journal}{\emph{california institute of technology}} (\bibinfo{year}{2011}).
\newblock


\bibitem[Wang et~al\mbox{.}(2018)]%
        {word_embed_and_graph_CVPR_2018_Xiaolong_Wang}
\bibfield{author}{\bibinfo{person}{Xiaolong Wang}, \bibinfo{person}{Yufei Ye}, {and} \bibinfo{person}{Abhinav Gupta}.} \bibinfo{year}{2018}\natexlab{}.
\newblock \showarticletitle{Zero-Shot Recognition via Semantic Embeddings and Knowledge Graphs}. In \bibinfo{booktitle}{\emph{2018 {IEEE} Conference on Computer Vision and Pattern Recognition, {CVPR} 2018.}} \bibinfo{pages}{6857--6866}.
\newblock


\bibitem[Website(2001)]%
        {wiki}
\bibfield{author}{\bibinfo{person}{Website}.} \bibinfo{year}{2001}\natexlab{}.
\newblock \bibinfo{title}{Wikipedia}.
\newblock \bibinfo{howpublished}{\url{https://en.wikipedia.org/}}.
\newblock


\bibitem[Website(2020)]%
        {azanimal}
\bibfield{author}{\bibinfo{person}{Website}.} \bibinfo{year}{2020}\natexlab{}.
\newblock \bibinfo{title}{A-Z Animals}.
\newblock \bibinfo{howpublished}{\url{https://a-z-animals.com/}}.
\newblock


\bibitem[Website(2022)]%
        {aab}
\bibfield{author}{\bibinfo{person}{Website}.} \bibinfo{year}{2022}\natexlab{}.
\newblock \bibinfo{title}{All About Birds}.
\newblock \bibinfo{howpublished}{\url{https://www.allaboutbirds.org/}}.
\newblock


\bibitem[Xian et~al\mbox{.}(2016)]%
        {CVPR_2016_Yongqin_Xian}
\bibfield{author}{\bibinfo{person}{Yongqin Xian}, \bibinfo{person}{Zeynep Akata}, \bibinfo{person}{Gaurav Sharma}, \bibinfo{person}{Quynh Nguyen}, \bibinfo{person}{Matthias Hein}, {and} \bibinfo{person}{Bernt Schiele}.} \bibinfo{year}{2016}\natexlab{}.
\newblock \showarticletitle{Latent Embeddings for Zero-Shot Classification}. In \bibinfo{booktitle}{\emph{2016 {IEEE} Conference on Computer Vision and Pattern Recognition, {CVPR} 2016.}} \bibinfo{pages}{69--77}.
\newblock


\bibitem[Xian et~al\mbox{.}(2019a)]%
        {AWA2}
\bibfield{author}{\bibinfo{person}{Yongqin Xian}, \bibinfo{person}{Christoph~H. Lampert}, \bibinfo{person}{Bernt Schiele}, {and} \bibinfo{person}{Zeynep Akata}.} \bibinfo{year}{2019}\natexlab{a}.
\newblock \showarticletitle{Zero-Shot Learning - {A} Comprehensive Evaluation of the Good, the Bad and the Ugly}.
\newblock \bibinfo{journal}{\emph{{IEEE} Trans. Pattern Anal. Mach. Intell.}} \bibinfo{volume}{41}, \bibinfo{number}{9} (\bibinfo{year}{2019}), \bibinfo{pages}{2251--2265}.
\newblock


\bibitem[Xian et~al\mbox{.}(2018)]%
        {CVPR_2018_Yongqin_Xian}
\bibfield{author}{\bibinfo{person}{Yongqin Xian}, \bibinfo{person}{Tobias Lorenz}, \bibinfo{person}{Bernt Schiele}, {and} \bibinfo{person}{Zeynep Akata}.} \bibinfo{year}{2018}\natexlab{}.
\newblock \showarticletitle{Feature Generating Networks for Zero-Shot Learning}. In \bibinfo{booktitle}{\emph{2018 {IEEE} Conference on Computer Vision and Pattern Recognition, {CVPR} 2018.}} \bibinfo{pages}{5542--5551}.
\newblock


\bibitem[Xian et~al\mbox{.}(2019b)]%
        {CVPR_2019_f_vaegan_d2}
\bibfield{author}{\bibinfo{person}{Yongqin Xian}, \bibinfo{person}{Saurabh Sharma}, \bibinfo{person}{Bernt Schiele}, {and} \bibinfo{person}{Zeynep Akata}.} \bibinfo{year}{2019}\natexlab{b}.
\newblock \showarticletitle{{F-VAEGAN-D2:} {A} Feature Generating Framework for Any-Shot Learning}. In \bibinfo{booktitle}{\emph{2019 {IEEE} Conference on Computer Vision and Pattern Recognition, {CVPR} 2019.}} \bibinfo{pages}{10275--10284}.
\newblock


\bibitem[Xu et~al\mbox{.}(2020)]%
        {NIPS_2022_Wenjia_Xu}
\bibfield{author}{\bibinfo{person}{Wenjia Xu}, \bibinfo{person}{Yongqin Xian}, \bibinfo{person}{Jiuniu Wang}, \bibinfo{person}{Bernt Schiele}, {and} \bibinfo{person}{Zeynep Akata}.} \bibinfo{year}{2020}\natexlab{}.
\newblock \showarticletitle{Attribute Prototype Network for Zero-Shot Learning}. In \bibinfo{booktitle}{\emph{Advances in Neural Information Processing Systems 33: Annual Conference on Neural Information Processing Systems 2020, NeurIPS 2020.}}
\newblock


\bibitem[Xu et~al\mbox{.}(2022)]%
        {VGSE}
\bibfield{author}{\bibinfo{person}{Wenjia Xu}, \bibinfo{person}{Yongqin Xian}, \bibinfo{person}{Jiuniu Wang}, \bibinfo{person}{Bernt Schiele}, {and} \bibinfo{person}{Zeynep Akata}.} \bibinfo{year}{2022}\natexlab{}.
\newblock \showarticletitle{{VGSE:} Visually-Grounded Semantic Embeddings for Zero-Shot Learning}. In \bibinfo{booktitle}{\emph{2022 {IEEE} Conference on Computer Vision and Pattern Recognition, {CVPR} 2022.}} \bibinfo{pages}{9306--9315}.
\newblock


\bibitem[Yamada et~al\mbox{.}(2020)]%
        {pre_trained_LM_EMNLP_2020_Wikipedia2vec}
\bibfield{author}{\bibinfo{person}{Ikuya Yamada}, \bibinfo{person}{Akari Asai}, \bibinfo{person}{Jin Sakuma}, \bibinfo{person}{Hiroyuki Shindo}, \bibinfo{person}{Hideaki Takeda}, \bibinfo{person}{Yoshiyasu Takefuji}, {and} \bibinfo{person}{Yuji Matsumoto}.} \bibinfo{year}{2020}\natexlab{}.
\newblock \showarticletitle{Wikipedia2Vec: An Efficient Toolkit for Learning and Visualizing the Embeddings of Words and Entities from Wikipedia}. In \bibinfo{booktitle}{\emph{Proceedings of the 2020 Conference on Empirical Methods in Natural Language Processing: System Demonstrations, {EMNLP} 2020.}} \bibinfo{pages}{23--30}.
\newblock


\bibitem[Yu et~al\mbox{.}(2013)]%
        {Attribute_drawback_CVPR_2013}
\bibfield{author}{\bibinfo{person}{Felix~X. Yu}, \bibinfo{person}{Liangliang Cao}, \bibinfo{person}{Rog{\'{e}}rio~Schmidt Feris}, \bibinfo{person}{John~R. Smith}, {and} \bibinfo{person}{Shih{-}Fu Chang}.} \bibinfo{year}{2013}\natexlab{}.
\newblock \showarticletitle{Designing Category-Level Attributes for Discriminative Visual Recognition}. In \bibinfo{booktitle}{\emph{2013 {IEEE} Conference on Computer Vision and Pattern Recognition, {CVPR} 2013.}} \bibinfo{pages}{771--778}.
\newblock


\bibitem[Zhang and Feng(2023)]%
        {MM_23_1}
\bibfield{author}{\bibinfo{person}{Yang Zhang} {and} \bibinfo{person}{Songhe Feng}.} \bibinfo{year}{2023}\natexlab{}.
\newblock \showarticletitle{Enhancing Domain-Invariant Parts for Generalized Zero-Shot Learning}. In \bibinfo{booktitle}{\emph{Proceedings of the 31st {ACM} International Conference on Multimedia, {MM} 2023.}} \bibinfo{pages}{6283--6291}.
\newblock


\bibitem[Zhao et~al\mbox{.}(2023)]%
        {MM_23_2}
\bibfield{author}{\bibinfo{person}{Peng Zhao}, \bibinfo{person}{Qiangchang Wang}, {and} \bibinfo{person}{Yilong Yin}.} \bibinfo{year}{2023}\natexlab{}.
\newblock \showarticletitle{{M3R:} Masked Token Mixup and Cross-Modal Reconstruction for Zero-Shot Learning}. In \bibinfo{booktitle}{\emph{Proceedings of the 31st {ACM} International Conference on Multimedia, {MM} 2023.}} \bibinfo{pages}{3161--3171}.
\newblock


\bibitem[Zhu et~al\mbox{.}(2018)]%
        {Document-based_CVPR_2018_Yizhe_Zhu}
\bibfield{author}{\bibinfo{person}{Yizhe Zhu}, \bibinfo{person}{Mohamed Elhoseiny}, \bibinfo{person}{Bingchen Liu}, \bibinfo{person}{Xi Peng}, {and} \bibinfo{person}{Ahmed Elgammal}.} \bibinfo{year}{2018}\natexlab{}.
\newblock \showarticletitle{A Generative Adversarial Approach for Zero-Shot Learning From Noisy Texts}. In \bibinfo{booktitle}{\emph{2018 {IEEE} Conference on Computer Vision and Pattern Recognition, {CVPR} 2018.}} \bibinfo{pages}{1004--1013}.
\newblock


\bibitem[Zhu et~al\mbox{.}(2019)]%
        {ICCV_2019_Yizhe_zhu}
\bibfield{author}{\bibinfo{person}{Yizhe Zhu}, \bibinfo{person}{Jianwen Xie}, \bibinfo{person}{Bingchen Liu}, {and} \bibinfo{person}{Ahmed Elgammal}.} \bibinfo{year}{2019}\natexlab{}.
\newblock \showarticletitle{Learning Feature-to-Feature Translator by Alternating Back-Propagation for Generative Zero-Shot Learning}. In \bibinfo{booktitle}{\emph{2019 {IEEE/CVF} International Conference on Computer Vision, {ICCV} 2019.}} \bibinfo{pages}{9843--9853}.
\newblock


\end{thebibliography}

\appendix

\clearpage

The supplementary material provides:
\begin{itemize}
    \item Section \ref{sec: ablation_prompts}: Ablation over different prompts.
    \item Section \ref{sec: partial score}: Ablation on $p$ in partial score.
    \item Section \ref{sec: different_LLM}: Comparison with SOTA on different LLMs.
    \item Section \ref{sec: generative_methods}: Comparison with generative methods. 
    \item Section \ref{sec: additional_related_work}: Additional related work.
    \item Section \ref{sec: comparison_CLIP}: Comparison with CLIP without prior on labels.
    \item Section \ref{sec: training_details}: Training details. 
    \item Section \ref{sec: less_described_class}: Details of less-described categories.
    \item Section \ref{sec: category_docs}: Examples of category documents.
\end{itemize}


\section{Extra Ablations on EmDepart}

\subsection{Ablation over Different Prompts}
\label{sec: ablation_prompts}
To find a robust prompt, we consider the following three prompts for enriching less-described documents:

\textbf{Direct Prompt}: \textit{`` What does a \{\textbf{class name}\} look like? Please describe within 500 words. ''}

\textbf{Detailed Prompt}: \textit{`` Now you are a \{\textbf{type}\} expert. I will give you \{\textbf{type}\} name, and you need to give detailed information. I want you to define \{\textbf{class name}\}. Please describe within 500 words. ''}

\textbf{Visual Prompt}: \textit{``Now you are a \{\textbf{type}\} expert. I will give you \{\textbf{type}\} name, and you need to give detailed visual information about its shape, color, appearance, habitat, etc. I want you to define \{\textbf{class name}\}. Please describe within 500 words.''}

We leverage the coarse-grained species as \textit{\{type\}}, \textit{i.e.}, animal for AWA2, bird for CUB, and flower for FLO. 
Besides, \textit{\{class name\}} denotes the name of the labeled class.

In Table \ref{tab: ablation_prompts}, we see that all prompts achieve performance improvements, verifying the effectiveness of enriching less-described documents.
The visual prompt achieves the best result, which enriches less-described classes with more diverse descriptions.

{
    \setlength{\tabcolsep}{3.5pt}
    \renewcommand{\arraystretch}{1.2} 
    
    \begin{table}[hb]
    \setlength{\aboverulesep}{0pt}\setlength{\belowrulesep}{0pt}
    \centering
    \caption{Ablation over different prompts on AWA2 and FLO datasets. We test each prompt with three times. The best results overall are in bold.}
    \vspace{3pt}
    \label{tab: ablation_prompts}
        {
        \begin{tabular}{l ll ll}
        \toprule
        \textbf{Template} &  \multicolumn{2}{c}{\textbf{AWA2}} & \multicolumn{2}{c}{\textbf{FLO}} \\
        \cmidrule(lr){2-3} \cmidrule(lr){4-5} 
        \textbf{Style} & \textbf{T1} & \textbf{H} & \textbf{T1} & \textbf{H} \\
        \midrule
        \texttt{Wiki} & 81.4 & 81.5 & 47.2 & 59.5  \\
        \texttt{Direct Prompt} & $82.2_{\pm0.95}$ & $82.6_{\pm0.81}$ & $48.2_{\pm0.68}$ & $59.9_{\pm0.57}$    \\
        \texttt{Detailed Prompt} & $84.1_{\pm0.53}$ & $83.7_{\pm0.47}$ & $51.2_{\pm0.45}$ & $64.7_{\pm0.87}$   \\
        \texttt{Visual Prompt} & $\textbf{86.1}_{\pm0.16}$ & $\textbf{84.8}_{\pm0.29}$ & $\textbf{53.3}_{\pm0.41}$ & $\textbf{67.3}_{\pm0.82}$  \\
        \bottomrule
        \end{tabular}}
    \end{table}
}

\subsection{Ablation on $p$ in Partial Score}
\label{sec: partial score}

In Table \ref{tab: partial score}, we ablate the value of $p$ in the partial score function to determine the optimal value on three datasets. 
The number of view embeddings $k$ are set to 4, 5, and 4 for AWA2, CUB, and FLO datasets, respectively.

{
    \setlength{\tabcolsep}{3.5pt}
    \renewcommand{\arraystretch}{1.2} 
    
    \begin{table}[htb]
    \setlength{\aboverulesep}{0pt}\setlength{\belowrulesep}{0pt}
    \centering
    \caption{Ablation on $p$ in partial score function. The best results overall are in bold.}
    \label{tab: partial score}
        {
        \begin{tabular}{l c c c }
        \toprule
        \multirow{2}{7em}{\textbf{Model}}  & \textbf{AWA2} & \textbf{CUB} & \textbf{FLO}  \\
        \cmidrule(lr){2-2} \cmidrule(lr){3-3}  \cmidrule(lr){4-4}
        & \textbf{T1} & \textbf{T1} & \textbf{T1} \\
        
        \midrule
        \texttt{w/o Partial Score} & 85.71 & 52.58 & 52.99   \\
        \texttt{$p = 1$} & 85.32 & 52.64 &   \textbf{53.31} \\ 
        \texttt{$p = 2$} & 86.00 & 52.71 & 53.19   \\
        \texttt{$p = 3$} & \textbf{86.13} & \textbf{52.84} &  52.91  \\
        \texttt{$p = 4$} & - & 52.74 & -   \\
        
        \bottomrule
        \end{tabular}}
    \end{table}
}

The results show that the optimal value of $p$ in the partial score function varies with the dataset. 
Specifically, for the AWA2 and CUB datasets, a higher value ($p=3$) leads to better performance. 
In contrast, for the FLO dataset, a lower value ($p=1$) is more effective. It is noteworthy that lower values, such as $p=1$ for AWA2 and CUB, result in the performance decrease. This is due to insufficient engagement with semantic information from view embeddings.

\section{Extra Experiments}

\subsection{Comparison with SOTA on Different LLMs}
\label{sec: different_LLM}

In Table ~\ref{tab: different_LLM}, we compare the SOTA method in document-based ZSL on different LLMs (GPT3 ~\cite{prompt_1}, PaLM ~\cite{PaLM}, ChatGPT~\cite{ChatGPT}). 
We see that our EmDepart with Wiki outperforms I2MVFormer with Wiki+ChatGPT across all metrics. 
It verifies that modeling partial association is helpful for knowledge transfer and significantly improves performance.
With the larger scale of LLMs, the EmDepart achieves better performance with more diverse visual descriptions. 
The model with ChatGPT achieves the best result.

{
    \setlength{\tabcolsep}{3.5pt}
    \renewcommand{\arraystretch}{1.2} 
    
    \begin{table}[hb]
    \setlength{\aboverulesep}{0pt}\setlength{\belowrulesep}{0pt}
    \centering
    \caption{Comparison with SOTA methods on different LLMs. The best results within a method are \underline{underline}. The best results overall are in bold.}
    \vspace{3pt}
    \label{tab: different_LLM}
        {
        \begin{tabular}{l c cc cc}
        \toprule
        \multirow{2}{5em}{\textbf{Model}} & \textbf{Auxiliary} &  \multicolumn{2}{c}{\textbf{AWA2}} & \multicolumn{2}{c}{\textbf{FLO}} \\
        \cmidrule(lr){3-4} \cmidrule(lr){5-6} 
         & \textbf{Information} & \textbf{T1} & \textbf{H} & \textbf{T1} & \textbf{H} \\
        \midrule
        \multirow{4}{*}{I2MVFormer ~\cite{I2MVFormer} }& \texttt{Wiki} & 73.6 & 73.8 & 41.3 & 51.2 \\
         & \texttt{Wiki$+$GPT3}  & 74.2 & 74.2 & 44.2 & 54.5 \\
         & \texttt{Wiki$+$PaLM}  & 79.6 & 76.6 & 46.2 & 57.1 \\
         & \texttt{Wiki$+$ChatGPT} & \underline{79.9} & \underline{80.5} & \underline{47.7} & \underline{59.1} \\
        \midrule
        \multirow{3}{*}{\textbf{EmDepart} (Ours)} & \texttt{Wiki} & 81.4 & 81.5 & 47.2 & 59.5 \\
         & \texttt{Wiki$+$GPT3} & 82.3 & 82.2 & 53.2 & 65.5 \\
         & \texttt{Wiki$+$PaLM} & 84.1 & 82.5 & 50.2 & 64.3 \\
         & \texttt{Wiki$+$ChatGPT} & \underline{\textbf{86.1}} & \underline{\textbf{84.8}}  & \underline{\textbf{53.3}} & \underline{\textbf{67.3}} \\
        
        \bottomrule
        \end{tabular}}
    \end{table}
}

{
    \renewcommand{\arraystretch}{1.2} 
    
    \begin{table*}[!h]
        \caption{
            Comparison with SOTA methods in document-based ZSL on three benchmark datasets. 
            We evaluate methods on documents sourced from Wiki.
            The best results overall are in \textbf{bold}.
            }
        \centering
        \small
    	\setlength{\aboverulesep}{0pt}\setlength{\belowrulesep}{0pt}
    	\resizebox{\linewidth}{!}{ %
    	\begin{tabular}{llc ccc ccc ccc ccc}
    			\toprule
    		&	& & \multicolumn{3}{c}{\textbf{Zero-Shot Learning}} & \multicolumn{9}{c}{\textbf{Generalized Zero-Shot Learning}} \\
    			\cmidrule(lr){4-6} \cmidrule(lr){7-15}
    		 \textbf{Type} &	\textbf{Model}	&  \textbf{ \makecell {Auxiliary \\ Information}} & {\textbf{AWA2}} & {\textbf{CUB}} & {\textbf{FLO}} & \multicolumn{3}{c}{\textbf{AWA2}} & \multicolumn{3}{c}{\textbf{CUB}} & \multicolumn{3}{c}{\textbf{FLO}}  \\
    			\cmidrule(lr){4-4} \cmidrule(lr){5-5} \cmidrule(lr){6-6} \cmidrule(lr){7-9} \cmidrule(lr){10-12} \cmidrule(lr){13-15}
    		& & & \makecell[c]{\textbf{T1}} & \makecell[c]{\textbf{T1}} & \makecell[c]{\textbf{T1}} & \textbf{U} & \textbf{S} & \makecell[c]{\textbf{H}} & \textbf{U} & \textbf{S} & \makecell[c]{\textbf{H}} & \textbf{U} & \textbf{S} & \makecell[c]{\textbf{H}}  \\
    			
                \midrule
                \multirow{2}{*}{Generative}& {\texttt{GAZSL} ~\cite{Document-based_CVPR_2018_Yizhe_Zhu}}
                & \texttt{Wiki} & 
                {83.1} & {42.9} & {34.2} & 
                {56.8} & \textbf{94.7} & {71.0} & 
                {15.9} & {50.4} & {24.1} & 
                {28.8} & {90.1} & {43.7} \\ 
    
                & {\texttt{f-VAEGAN-D2} ~\cite{CVPR_2019_f_vaegan_d2}}
                & \texttt{Wiki} & 
                \textbf{85.1} & {41.9} & {36.9} & 
                {73.2} & {81.7} & {77.2} & 
                {33.4} & {57.3} & {42.2} & 
                {30.0} & {97.3} & {45.8} \\

                \midrule
                \multirow{3}{*}{Discriminative}& {\texttt{I2DFormer}~\cite{I2DFormer}}
                & \texttt{Wiki} & 
                {76.4} & {45.4} & {40.0} & 
                {66.8} & {76.8} & {71.5} & 
                {35.3} & {57.6} & {43.8} & 
                {35.8} & {91.9} & {51.5} \\ 
                
                & {\texttt{I2MVFormer}~\cite{I2MVFormer}}  
                & \texttt{Wiki} & 
                {73.6} & {42.1} & {41.3} & 
                {66.6} & {82.9} & {73.8} &
                {32.4} & \textbf{63.1} &  {42.8} & 
                {34.9} & {96.1} & {51.2} \\
    
                & \textbf{\ModelName} \text{(Ours)}    
                & \texttt{Wiki}  & 
               {81.4}  & \textbf{50.2} & \textbf{47.2} & 
                \textbf{76.0} & 87.8 & \textbf{81.5} & 
                \textbf{42.6} & 56.3 & \textbf{48.5} & 
                \textbf{42.7} & \textbf{97.6} & \textbf{59.5} \\

                \bottomrule
    	\end{tabular}
    }
    	\label{tab: generative_methods}
    \end{table*}
}

\subsection{Comparison with Generative Methods}
\label{sec: generative_methods}
In Table \ref{tab: generative_methods}, we compare our EmDepart with SOTA generative methods, which generate samples for unseen classes and convert the ZSL to a supervised task.

On the AWA2 dataset, generative methods achieve superior performance. 
This is because AWA2 is a coarse-grained dataset, where collected documents provide richer discriminative information than traditional attributes. 
By generating samples for unseen classes, these methods achieve better results.
However, generative methods fail on CUB and FLO datasets compared to discriminative methods. 
This is due to non-visual noisy descriptions (such as sound, diet, and organ) in documents, Which lead to hard knowledge transfer and generate low-quality samples for unseen classes.

In contrast, discriminative methods enhance the fine-grained alignment between image patches and text words, implicitly filtering out irrelevant information. 
Our EmDepart generates embeddings from multiple semantic views, accurately modeling the semantic alignment according to the matching information and achieving better performance.

\subsection{Comparison with CLIP without Prior on Class Labels}
\label{sec: comparison_CLIP}

Recently, some work ~\cite{VDT_2023_ICLR, VDT_2023_ICCV_what_does, VDT_2023_ICCV_Waffle, VDT_2023_ICCV_gpt4, VDT_2023_NIPS_Hierarchical} shows that visual descriptions are helpful for improving the performance of vision-language models like CLIP ~\cite{CLIP} on image classification tasks.
However, these methods heavily rely on prior information on class names. 
We consider the following prompts to evaluate the generalization ability of models under the situation with and without the class name prior.

\textbf{Prompt with document}: ``\textit{A photo of a \{\textbf{class name}\}. \{\textbf{document}\}.}''

\textbf{Prompt with document and without class name prior}: ``\textit{A photo of a \{\textbf{type}\}. \{\textbf{document}\} }''

The \textit{\{type\}} denotes the species of categories, such as animal, bird, and flower. Besides, the \textit{\{class name\}} denotes labels in the dataset. 
We enrich each class with documents sourced from the encyclopedia.

In Table \ref{tab: comparison_CLIP}, we show the results of CLIP and our EmDepart in this situation.
We see that the performance of CLIP increases with the help of documents similar to ~\cite{VDT_2023_ICLR, VDT_2023_ICCV_what_does, VDT_2023_ICCV_Waffle, VDT_2023_ICCV_gpt4, VDT_2023_NIPS_Hierarchical}.
However, the performance decreases a lot when the model is without prior information on the class name.
In this situation, our EmDepart outperforms CLIP and demonstrates superior generation ability.

{
    \setlength{\tabcolsep}{3pt}
    \renewcommand{\arraystretch}{1.2} 

    \begin{table}[t]
    \centering
    \caption{Comparison with CLIP in different settings. The best and worst results are in \textbf{bold} and \red{red}, respectively. We evaluate mean per-class accuracy for CLIP.
    }
    \label{tab: comparison_CLIP}
    \resizebox{\linewidth}{!}
        {
        \begin{tabular}{lccc}
        \toprule
        \textbf{Model} &  \textbf{AWA2} & \textbf{CUB}  &   \textbf{FLO} \\
        \midrule
        \texttt{CLIP} \cite{CLIP} & 92.0 & 54.1 & 66.8  \\
        \texttt{CLIP w/ document}  & \textbf{92.2} & \textbf{57.6} & \textbf{71.0} \\
        \textbf{EmDepart}  & 84.8  & 51.9 & 67.3 \\
        \midrule
        \texttt{CLIP w/ document and w/o class name}  & \red{47.2} & \red{14.1} & \red{18.2} \\
        \textbf{EmDepart} w/o class name & 68.5 & 43.2 & 56.4 \\
        \bottomrule
        \end{tabular}}
    \end{table}
}






\section{Additional Related Work}
\label{sec: additional_related_work}

\subsection{Image Classification with Descriptions from LLMs}

Image classification with descriptions from LLMs aims to leverage descriptions to enhance the generalization ability.
Recent work ~\cite{VDT_2023_ICLR, VDT_2023_ICCV_what_does, VDT_2023_ICCV_Waffle, VDT_2023_ICCV_gpt4, VDT_2023_NIPS_Hierarchical} generates visual descriptions by LLMs to enrich the language-side semantics.
This process improves the performance of large-scale vision-language models like CLIP ~\citep{CLIP} in downstream tasks . 
However, they may fail when recognizing the category without the prior information on the class name.
Other work uses LLMs to provide additional auxiliary information in specific domains.  
I2MVFormer ~\citep{I2MVFormer} employs LLMs to rewrite documents in various styles, aiding knowledge transfer by integrating information among different style descriptions. 
However, it ignores the fact that there are sparse descriptions for some fine-grained categories.  
Although documents are in different styles, they still contain little semantics for these categories.  
In our work, we instruct LLMs to enrich less detailed category descriptions.

\subsection{Attribute Localization}
In attribute-based ZSL, some work enhances local alignment to extract discriminative information for fine-grained classification.
They aim to build an accurate semantic alignment between high-quality visual attributes and image regions.
DAZLE~\cite{DAZLE} utilizes dense attribute-based attention to focus on relevant image regions. 
Transzero~\cite{TransZero} proposes an attribute-guided transformer to improve the transferability of visual features and learn attribute localization.
RGAT~\cite{RGAT} leverages attribute-region graphs to capture relationships between region features. 
In this work, we improve local alignment between image patches and word tokens in noisy documents. 
Unlike annotated attributes, documents contain noisy, non-visual information.
It is suboptimal to align all semantics of noisy words with image regions.
We partially align words with image regions according to their semantic relevance.

\section{Training Details}
\label{sec: training_details}

\subsection{Calibrated Stacking}
We apply calibrated stacking (CS) \cite{calibrated_stack} to trade-off calibration degrees in the GZSL settings. 
This is helpful for reducing the bias towards seen classes. 
We modify Eq. 13 in the main paper:
\begin{equation}
    \hat {\boldsymbol y} = arg\, max_{{\boldsymbol d}' \in \mathcal{D}^s \cup \mathcal{D}^u} \; (S_{p} (\boldsymbol x, {\boldsymbol d}') - \gamma \mathbb{I}_{\mathcal D^s} (\hat {\boldsymbol y})).
\end{equation}
Here, $\mathbb{I}_{\mathcal D^s}$ represents an indicator function, which is 1 when  ${\boldsymbol d}' \in \mathcal{D}^s$ and 0 otherwise. A calibrated factor $\gamma$ is applied to trade off the calibration degree on seen classes.

\subsection{Grid Search}
Hyperparameters are optimized by grid search in the validation split. 
We set the range for $\lambda_{local} \in [0, 0.01, 0.1, 0.25, 0.5, 0.75, 1]$, $\lambda_{var} \in [0, 0.01, 0.1, 0.5, 1, 3, 5]$, $\lambda_{div} \in [0, 0.01, 0.1, 0.5, 1, 3, 5]$, and $k \in [0, 1, 2, 3, 4, 5, 6, 7, 8]$.
Once the hyperparameters are confirmed, we merge the validation with the training split to obtain the performance on the test split.
The effect of hyperparameters in EmDepart is shown in Figure 7 in the main paper.

\subsection{Additional Training Details}
We implement our framework with Pytorch and train on an Nvidia GeForce RTX 3090 GPU.
Similar to \cite{I2DFormer, I2MVFormer}, we use the ViT-B/16 \cite{vit} pre-trained on ImageNet 1K \cite{ImageNet} as the visual backbone, which respects the GUB split \cite{AWA2}.
The model is trained by the Adam optimizer with a cosine-decreased learning rate.
The detailed hyperparameters are shown in Table \ref{tab: train_details_AWA2} for AWA2, Table \ref{tab: train_details_CUB} for CUB and Table \ref{tab: train_details_FLO} for FLO.

{
    \renewcommand{\arraystretch}{1.} 

    \begin{table}[t]
    \centering
    \caption{\textbf{Hyperparameters settings for AWA2 dataset.}}
    \label{tab: train_details_AWA2}
        \resizebox{0.92\linewidth}{!}
        {
        \begin{tabular}{l c}
        \toprule
        \textbf{Config} &  \textbf{AWA2} 
        \\
        \midrule
        \textbf{Regular Training Setting} & \\
        optimizer & Adam\\
        base learning rate & 1.0e-4  \\
        dropout & 0.35  \\
        batch size & 64  \\
        learning rate schedule & cosine decay  \\
        warmup epochs & 0 \\
        epochs & 32  \\
        augmentation & RandomResizedCrop \\ 
        \midrule
        \textbf{Specific Settings in EmDepart} & \\
        $\mathcal{\lambda}_{local}$ & 0.1  \\
        $\mathcal{\lambda}_{var}$ & 1.0  \\
        $\mathcal{\lambda}_{div}$ & 3.0  \\
        number of view embeddings $k$ & 4 \\ 
        $p$ in partial score & 3 \\
        $\tau$ in Eq. 9 & 32.0 \\
        $\epsilon$ in Eq. 4 & 1e-4 \\
        $\gamma$ in Eq.4 & 0.10 \\
        dimension of semantic embedding $r$ & 256 \\
        layers of text encoder & 2 \\
        layers of MLP in image perceiver & 2 \\
        layers of visual SDM & 2 \\
        layers of textual SDM & 2 \\
        \bottomrule
        \end{tabular}
        }
    \end{table}
}

{
    \renewcommand{\arraystretch}{1.} 

    \begin{table}[h]
    \centering
    \caption{\textbf{Hyperparameters settings for CUB dataset.}}
    \label{tab: train_details_CUB}
        \resizebox{0.92\linewidth}{!}
        {
        \begin{tabular}{l c}
        \toprule
        \textbf{Config} &  \textbf{CUB} 
        \\
        \midrule
        \textbf{Regular Training Setting} & \\
        optimizer & Adam\\
        base learning rate & 8.0e-4  \\
        dropout & 0.15  \\
        batch size & 40  \\
        learning rate schedule & cosine decay  \\
        warmup epochs & 2 \\
        epochs & 32  \\
        augmentation & RandomResizedCrop \\ 
        \midrule
        \textbf{Specific Settings in EmDepart} & \\
        $\mathcal{\lambda}_{local}$ & 0.5  \\
        $\mathcal{\lambda}_{var}$ & 1.0  \\
        $\mathcal{\lambda}_{div}$ & 3.0  \\
        number of view embeddings $k$ & 5 \\ 
        $p$ in partial score & 3 \\
        $\tau$ in Eq. 9 & 4.2 \\
        $\epsilon$ in Eq. 4 & 1e-4 \\
        $\gamma$ in Eq.4 & 0.25 \\
        dimension of semantic embedding $r$ & 64 \\
        layers of text encoder & 2 \\
        layers of MLP in image perceiver & 2 \\
        layers of visual SDM & 2 \\
        layers of textual SDM & 2 \\
        \bottomrule
        \end{tabular}
        }
    \end{table}
}

{
    \renewcommand{\arraystretch}{1.} 

    \begin{table}[h]
    \centering
    \caption{\textbf{Hyperparameters settings for FLO dataset.}}
    \label{tab: train_details_FLO}
        \resizebox{0.92\linewidth}{!}
        {
        \begin{tabular}{l c}
        \toprule
        \textbf{Config} &  \textbf{FLO} 
        \\
        \midrule
        \textbf{Regular Training Setting} & \\
        optimizer & Adam\\
        base learning rate & 5.0e-4  \\
        dropout & 0.12  \\
        batch size & 48  \\
        learning rate schedule & cosine decay  \\
        warmup epochs & 0 \\
        epochs & 40  \\
        augmentation & RandomResizedCrop \\ 
        \midrule
        \textbf{Specific Settings in EmDepart} & \\
        $\mathcal{\lambda}_{local}$ & 0.5  \\
        $\mathcal{\lambda}_{var}$ & 1.0  \\
        $\mathcal{\lambda}_{div}$ & 3.0  \\
        number of view embeddings $k$ & 4 \\ 
        $p$ in partial score & 1 \\
        $\tau$ in Eq. 9 & 4.0 \\
        $\epsilon$ in Eq. 4 & 1e-4 \\
        $\gamma$ in Eq.4 & 0.75 \\
        dimension of semantic embedding $r$ & 128 \\
        layers of text encoder & 2 \\
        layers of MLP in image perceiver & 2 \\
        layers of visual SDM & 2 \\
        layers of textual SDM & 2 \\
        \bottomrule
        \end{tabular}
        }
    \end{table}
}

\section{Details of Less-described Categories}
\label{sec: less_described_class}

In our work, we leverage LLMs to supplement less-described category documents.
To save computation costs, we select a set of categories instead of all categories for enriching documents.
The less-described categories for each dataset are shown below.

{
    \setlength{\tabcolsep}{3pt}
    \renewcommand{\arraystretch}{1.2} 

    \begin{table}[h]
    \centering
    \caption{Details of less-described categories.
    }
    \label{tab: less_described_class}
        {
        \begin{tabular}{lccc}
        \toprule
         &  \textbf{AWA2} & \textbf{CUB}  &   \textbf{FLO} \\
        \midrule
        \texttt{number of classes} & 50 & 200 & 102  \\
        \texttt{number of less-described classes}  & 21 & 74 & 59 \\
        
        \bottomrule
        \end{tabular}}
    \end{table}
}

\textbf{AWA2}: dalmatian, persian cat, german shepherd, blue whale, siamese cat, moose, gorilla, ox, fox, rabbit, chihuahua, collie, dolphin, grizzly bear, skunk, hippopotamus, spider monkey, wolf, weasel, zebra, buffalo.

\textbf{CUB}: Laysan Albatross, Parakeet Auklet, Yellow headed Blackbird, Bobolink, Lazuli Bunting, Gray Catbird, Yellow breasted Chat, Eastern Towhee, Chuck will Widow, Red faced Cormorant, Shiny Cowbird, Fish Crow, Mangrove Cuckoo, Least Flycatcher, Scissor tailed Flycatcher, Vermilion Flycatcher, American Goldfinch, Eared Grebe, Pied billed Grebe, Ivory Gull, Anna Hummingbird, Ruby throated Hummingbird, Long tailed Jaeger, Blue Jay, Florida Jay, Green Jay, Tropical Kingbird, Gray Kingbird, Pied Kingfisher, Hooded Merganser, Red breasted Merganser, Clark Nutcracker, White breasted Nuthatch, Orchard Oriole, Ovenbird, Horned Puffin, Common Raven, American Redstart, Baird Sparrow, Clay colored Sparrow, Henslow Sparrow, Vesper Sparrow, Cape Glossy Starling, Summer Tanager, Elegant Tern, Black capped Vireo, Philadelphia Vireo, Bay breasted Warbler, Black throated Blue Warbler, Blue winged Warbler, Canada Warbler, Cerulean Warbler, Hooded Warbler, Kentucky Warbler, Magnolia Warbler, Prairie Warbler, Prothonotary Warbler, Louisiana Waterthrush, Red bellied Woodpecker, Red cockaded Woodpecker, Red headed Woodpecker, Bewick Wren, Rock Wren, Black footed Albatross, Least Auklet, Acadian Flycatcher, Yellow bellied Flycatcher, Pomarine Jaeger, Mockingbird, Black throated Sparrow, Cape May Warbler, Golden winged Warbler, Northern Waterthrush, Bohemian Waxwing.

\textbf{FLO}: hard-leaved pocket orchid, canterbury bells, sweet pea, monkshood, 
colt's foot, 
spear thistle, yellow iris, purple coneflower, peruvian lily, balloon flower, 
giant white arum lily, fritillary, grape hyacinth, prince of wales feathers, artichoke, sweet william, carnation, 
garden phlox, love in the mist, alpine sea holly, ruby-lipped cattleya, cape flower, great masterwort, sword lily, poinsettia, wallflower, marigold, oxeye daisy, wild pansy, primula, sunflower, gaura, black-eyed susan, silverbush, californian poppy, osteospermum, bearded iris, windflower, thorn apple, morning glory, passion flower, lotus, toad lily, anthurium, frangipani, clematis, hibiscus, columbine, tree mallow, magnolia, canna lily, bee balm, ball moss, foxglove, bougainvillea, mexican petunia, blanket flower, trumpet creeper, blackberry lily.

\section{Examples of Category Documents}
\label{sec: category_docs}

We show two examples of category documents for three datasets, which are the auxiliary information in our EmDepart.
We will release all the documents after the review process.

\subsection{AWA2}

\noindent \textbf{Giraffe}: The Giraffe is an animal with an enormously long neck which allows it to exploit the leaves and vegetation that are too high up for other animals to find. Despite their length, the neck of the Giraffe actually contains the same number of bones as numerous other hoofed mammals but they are simply longer in shape. The giraffe's elongated neck leads into a short body, with long and thin, straight legs and a long tail that is tipped with a black tuft that helps to keep flies away. The Giraffe tends to be white in color with brown or reddish markings that cover its body (with the exception of its white lower legs). The markings of each Giraffe are not only unique to that individual but they also vary greatly between the different Giraffe species in size, color, and the amount of white that surrounds them. All giraffes though have large eyes that along with their height give them excellent vision, and small horn-like ossicones on the top of their heads. Giraffes are animals that inhabit open woodlands and savannah where using their height they are able to see for great distances around them to watch out for approaching danger. A giraffe is a tall and slender mammal with a long neck and legs. Their coat is pale yellow with dark brown spots or irregular patches covering their whole body except for their underbelly. Their spots are unique to each individual and help them to blend into their habitat, making them difficult to spot by predators. Giraffes have small horn-like ossicones on top of their head, and their ears are small and tufted with hair. Their eyes are large, dark, and have long eyelashes. Long neck: Giraffes are well-known for their long necks, which are an adaptation for reaching high branches and leaves in trees. 

\noindent \textbf{Horse}: All horses have long necks that hold up their large, long heads. They have big eyes and ears, which are well-adapted for many environments. A mane of long hair grows down along their necks and their short tails are covered in coarse hairs, too. They come in a variety of colors because they have been bred so long for different traits. These animals are famously a hoofed mammal with one large toe at the end of each leg. Their hooves consist of horn material which comes in different colors. Black is the most common hoof color, but horses with white feet often have white hoofs. White hooves are actually more brittle than pigmented ones. Appaloosa horses have a beautiful mixture of multiple colors. These types of painted horses often have striped hoofs that include both pigmented and white hoof material. These animals are well-suited to all kinds of environments and climates. Domestic horses can live almost anywhere as long as they have shelter, food, and space to run. horses are generally known for their distinct physical characteristics such as a long face, large nostrils, muscular build, four legs, hooves, a mane and tail of hair, and varying coat colors and patterns. They also have big, expressive eyes, long necks, and pointed ears. horses are generally found in open fields, meadows, pastures, and sometimes in stables or barns if they are domesticated. Their natural habitat includes grassy plains, hills, and forests with access to water sources. Their surroundings are usually green and have varying degrees of vegetation cover.

\subsection{CUB}

\noindent \textbf{Brown Creeper}: Tiny, lanky songbirds with long, spine-tipped tails, slim bodies, and slender, decurved bills. Length: 4.7-5.5 in (12-14 cm). Weight: 0.2-0.3 oz (5-10 g). Wingspan: 6.7-7.9 in (17-20 cm).The bill is slender and decurved, perfect for probing into crevices in tree bark to find insects and spiders.Brown Creeper have streaked brown and buff upperparts, with a broad, buffy stripe over the eye. The underparts are white, usually hidden against the tree trunk.Their legs and feet are specialized for clinging to tree trunks, supporting their unique foraging behavior.The tail is long and spine-tipped, used for support as Brown Creeper hitch upward in a spiral around tree trunks.Their wings are well-suited for short flights between trees, necessary for their foraging style.Brown Creeper forage by hitching upward in a spiral around tree trunks and limbs, using their stiff tails for support, and fly weakly to the base of another tree to continue foraging.Brown Creeper are found in mature evergreen or mixed evergreen-deciduous forests for breeding. In winter, Brown Creeper can be found in a broader variety of forests, including deciduous woodlands.

\noindent \textbf{Mockingbird}: Medium-sized songbird, more slender than a thrush with a longer tail. Length: 8.3-10.2 in (21-26 cm). Weight: 1.6-2.0 oz (45-58 g). Wingspan: 12.2-13.8 in (31-35 cm).Long, thin bill with a hint of a downward curve.Overall gray-brown, paler on the breast and belly. Two white wingbars on each wing and a white patch in each wing. White outer tail feathers are also flashy in flight.Long legs that are well-adapted for running and hopping on the ground.Long tail that is gray-brown like the body, which appears particularly long in flight and aids in balance and maneuverability.Short, rounded, and broad wings, which are efficient for quick takeoffs and agile flight.Mockingbird are known for their songs and mimicry. Mockingbird sit conspicuously on high vegetation, fences, eaves, or telephone wires, or run and hop along the ground. Mockingbird are territorial and will aggressively chase off intruders.Found in towns, suburbs, backyards, parks, forest edges, and open land at low elevations.

\subsection{FLO}
\noindent \textbf{King Protea}: The king protea, scientifically known as Protea cynaroides, is an extraordinary flowering plant native to the southwestern coastal regions of South Africa. As the largest and most iconic member of the Protea family, this majestic flower possesses an enchanting beauty that captures the essence of its royal title.  The king protea boasts a distinctively regal appearance, with a large, spherical inflorescence that can reach up to 12 inches (30 centimeters) in diameter. This magnificent flower is characterized by its intricate structure, composed of multiple layers of petals surrounding a prominent central cone. The cone is adorned with an array of feathery, needle-like styles that extend outwards, adding a unique texture to the overall appearance.  The petals of the king protea are large and sturdy, each measuring around 4 to 6 inches (10 to 15 centimeters) in length. They can vary in color, ranging from soft hues of creamy whites, blush pinks, and delicate mauves to vibrant shades of deep crimson, burgundy, and coral. The petals exhibit a velvety texture and often feature a slightly waxy coating, enhancing their appeal and adding a touch of lustrous shine.  One of the distinguishing characteristics of the king protea is the presence of a prominent ring of long, stiff bracts that encircle the base of the flower. These bracts, often referred to as \"phylloid\" bracts, serve to protect and support the blooms, giving the inflorescence an impressive crown-like appearance. The bracts themselves can vary in color and are typically seen in shades of pale green, silvery grey, or even a reddish hue.  In its natural habitat, the king protea thrives in a diverse range of environments, predominantly found in the fynbos biome of South Africa. It prefers well-drained soils and can be seen flourishing along sandy coastal areas, mountain slopes, and even in the slightly more arid landscapes of the region. This resilient flower has adapted to withstand harsh conditions, including periods of drought and occasional wildfires, showcasing its remarkable ability to survive and retain its majestic allure.  The king protea is not merely a flower; it represents a symbol of strength, beauty, and resilience. Its captivating presence has made it an iconic emblem and a highly sought-after ornamental bloom. This regal flower is often used as a centerpiece or featured in floral arrangements, adding a sense of grandeur and elegance to any setting, whether it be a sophisticated event or a serene garden. 

\noindent \textbf{Sunflower}: Sunflowers, scientifically known as Helianthus annuus, are iconic and dazzling flowers that invoke thoughts of warm, sunny days and vibrant landscapes. These majestic plants belong to the Asteraceae family and can reach impressive heights, often towering over other plants in gardens and fields. With their distinct appearance and widespread popularity, sunflowers have become a true symbol of joy, vitality, and positivity.  The most striking feature of a sunflower is, undoubtedly, its enormous flower head. These flower heads, also known as inflorescences, are an incredible sight to behold, with an impressive size measuring between 10 to 30 centimeters in diameter or even larger in some cultivated varieties. Sunflowers are aptly named due to their stunning resemblance to the sun, both in shape and color.  When we examine a sunflower's blossoming head, we find a captivating arrangement of intricate details. A circular or semi-circular cluster of florets forms the center of the flower, aptly called the disk florets or the central disc. These disc florets are small, tubular-shaped, and densely packed together, creating a textured surface in stunning shades of dark brown, deep maroon, or even a rich purple-black hue. Upon closer inspection, the disc florets reveal intricate patterns and textures, often showcasing a striking contrast to the vibrant yellow petals surrounding them.  The disk florets are embraced by a ring of larger, elongated florets called ray florets, which contribute to the iconic shape of a sunflower. These ray florets possess a petal-like appearance, featuring vivid yellow or sometimes orange tones. Standing upright and radiating from the center, the ray florets resemble the rays of sunshine, providing an ethereal aura to the flower. While most sunflowers bear yellow petals, cultivated varieties may display a delightful array of shades, including vibrant oranges, warm reds, and even pale creams.  Sunflowers possess a well-defined reproductive structure, positioned at the center of the broad flower head. This structure consists of pistils and stamens, responsible for the plant's pollination and fertilization processes. Bees, butterflies, and other insects are commonly attracted to the sunflower's nectar and vibrant colors, aiding in the transfer of pollen from flower to flower.  In terms of habitat, sunflowers are native to North America and are highly adaptable plants, capable of thriving in diverse environments. They prefer areas with abundant sunlight, often gracing the landscape of fields, meadows, and gardens. Sunflowers have a notable preference for fertile, well-draining soil, but they can also withstand periods of drought.

\end{document}